\documentclass[lettersize,journal]{IEEEtran}
\usepackage{amsmath,amsfonts}
\usepackage{algorithm}
\usepackage{array}
\usepackage[caption=false,font=normalsize,labelfont=sf,textfont=sf]{subfig}
\usepackage{textcomp}
\usepackage{stfloats}
\usepackage{url}
\usepackage{verbatim}
\usepackage{graphicx}
\usepackage{cite}
\usepackage{booktabs}
\usepackage{bbm}
\usepackage{bm}
\usepackage{amssymb}
\usepackage{cite}
\usepackage{multirow}
\usepackage[accsupp]{axessibility}
\usepackage[pagebackref,breaklinks,colorlinks]{hyperref}
\usepackage{orcidlink}
\usepackage{mathrsfs}
\usepackage{epsfig}
\usepackage{enumerate}
\usepackage{cases}
\usepackage{CJK}
\usepackage{algorithmicx}
\usepackage{color}
\usepackage{colortbl}
\usepackage{float}
\usepackage{ragged2e}
\usepackage{csquotes}
\usepackage{threeparttable}
\usepackage{tablefootnote}
\usepackage{pdfpages}
\usepackage{titlesec}
\renewcommand{\thesection}{\arabic{section}}
\renewcommand{\thesubsection}{\thesection.\arabic{subsection}}
\makeatletter
\def\@seccntformat#1{\csname the#1\endcsname \ }
\makeatother
\renewcommand{\thesubsubsection}{\thesubsection.\arabic{subsubsection}}

\makeatletter
\def\@seccntformat#1{\csname the#1\endcsname \ }
\renewcommand{\subsubsection}[1]{%
    \refstepcounter{subsubsection}
    \par\vspace{0.5em}%
    \noindent\itshape \thesubsubsection~#1\normalfont\par%
}

\makeatother

\newcommand{\etal}{~et~al.~}

\begin{document}

	\title{Deep Learning for Camera Calibration and Beyond: A Survey}

	\author{Kang Liao, Lang Nie, Shujuan Huang, Chunyu Lin, Jing Zhang, Yao Zhao,~\IEEEmembership{Fellow,~IEEE}, Moncef Gabbouj,~\IEEEmembership{Fellow,~IEEE}, Dacheng Tao,~\IEEEmembership{Fellow,~IEEE}

		\thanks{Kang Liao, Lang Nie, Shujuan Huang, Chunyu Lin (corresponding author), and Yao Zhao are with the Institute of Information Science, Beijing Jiaotong University, Beijing 100044, China, and also with the Visual Intelligence + X International Joint Laboratory of the Ministry of Education, China (email: kang\_liao@bjtu.edu.cn, nielang@bjtu.edu.cn, shujuanhuang@bjtu.edu.cn, cylin@bjtu.edu.cn, yzhao@bjtu.edu.cn).}
  		\thanks{Moncef Gabbouj is with the Department of Computing Sciences, Tampere University, 33101 Tampere, Finland (e-mail: moncef.gabbouj@tuni.fi).}
          \thanks{Jing Zhang and Dacheng Tao are with the School of Computer Science, Faculty of Engineering, The University of Sydney, Australia (e-mail: jing.zhang1@sydney.edu.au; dacheng.tao@gmail.com).}

	}
    \markboth{Journal of \LaTeX\ Class Files,~Vol.~14, No.~8, August~2021}
    {Shell \MakeLowercase{\textit{et al.}}: A Sample Article Using IEEEtran.cls for IEEE Journals}
	
    \maketitle

		\begin{abstract}
\label{sec:Abstrat}
Camera calibration involves estimating camera parameters to infer geometric features from captured sequences, which is crucial for computer vision and robotics. However, conventional calibration is laborious and requires dedicated collection. Recent efforts show that learning-based solutions have the potential to be used in place of the repeatability works of manual calibrations. Among these solutions, various learning strategies, networks, geometric priors, and datasets have been investigated. In this paper, we provide a comprehensive survey of learning-based camera calibration techniques, by analyzing their strengths and limitations. Our main calibration categories include the standard pinhole camera model, distortion camera model, cross-view model, and cross-sensor model, following the research trend and extended applications. As there is no unified benchmark in this community, we collect a holistic calibration dataset that can serve as a public platform to evaluate the generalization of existing methods. It comprises both synthetic and real-world data, with images and videos captured by different cameras in diverse scenes. Toward the end of this paper, we discuss the challenges and provide further research directions. To our knowledge, this is the first survey for the learning-based camera calibration (spanned 10 years). The summarized methods, datasets, and benchmarks are available and will be regularly updated at \url{https://github.com/KangLiao929/Awesome-Deep-Camera-Calibration}. 
\end{abstract}
		
	
	\section{Introduction}
\label{sec:Introduction}

\begin{figure}[!t]
  \centering
  \includegraphics[width=.45\textwidth]{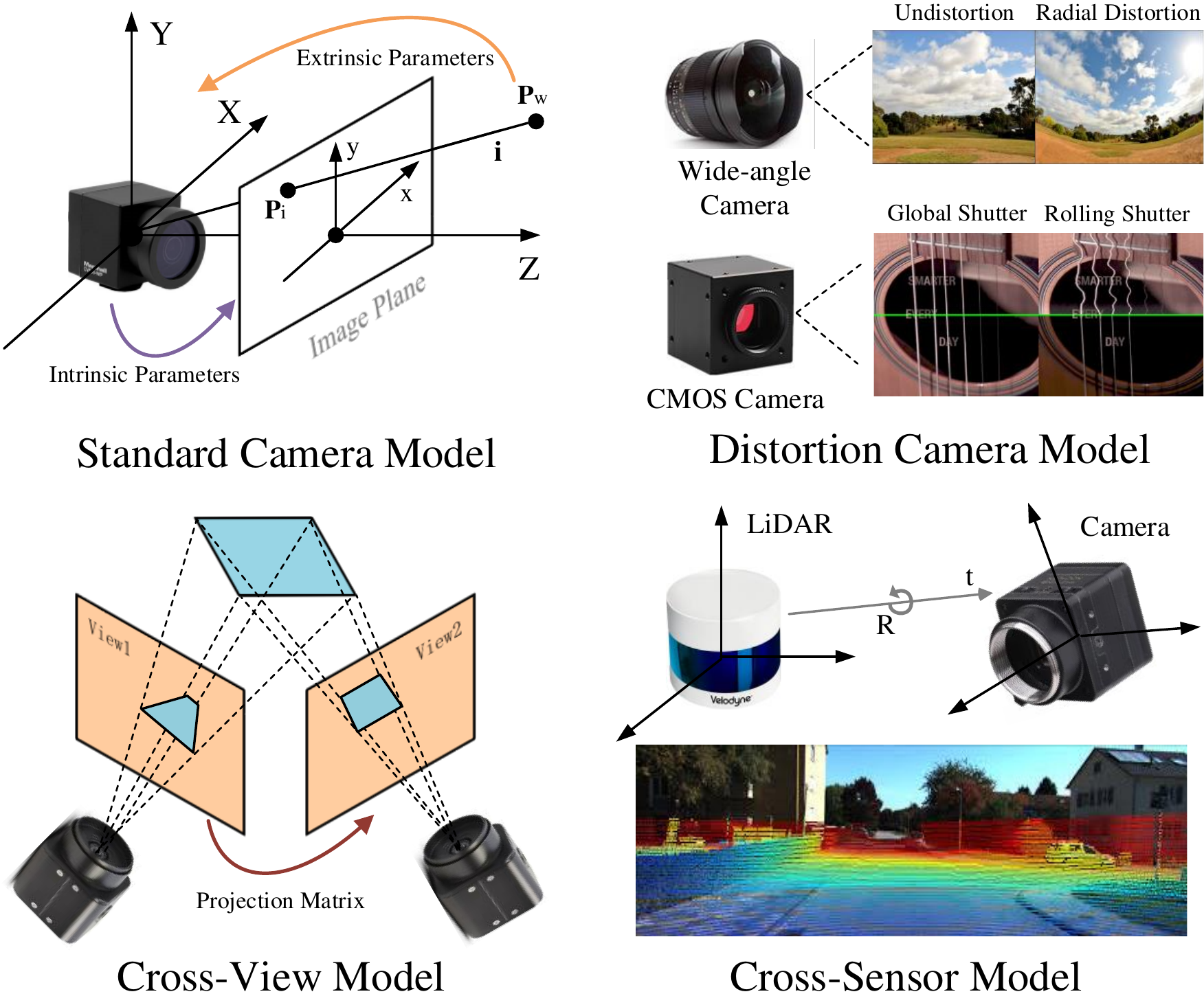}
  \caption{Popular calibration objectives, models, and extended applications in camera calibration.}
  \label{fig:teaser}
  \vspace{-0.4cm}
\end{figure}

\IEEEPARstart{C}{amera} calibration is a fundamental and indispensable field in computer vision and it has a long research history \cite{duane1971close, maybank1992theory, weng1992camera, zhang2000flexible, qi2023minimal}, tracing back to around 60 years ago\cite{brown1966decentering}. The first step for many vision and robotics tasks is to calibrate the intrinsic (image sensor and distortion parameters) and/or extrinsic (rotation and translation) camera parameters, ranging from computational photography, and multi-view geometry, to 3D reconstruction. In terms of the task type, there are different techniques to calibrate the standard pinhole camera, fisheye lens camera, stereo camera, light field camera, event camera, and LiDAR-camera system, etc. Figure~\ref{fig:teaser} shows the popular calibration objectives, models, and extended applications in camera calibration.

Traditional methods for camera calibration generally depend on hand-crafted features and model assumptions. These methods can be broadly divided into three categories. The most prevalent one involves using a known calibration target (\textit{e.g.}, a checkerboard) as it is deliberately moved in the 3D scene \cite{zhang1999flexible, gasparini2009plane, shah1994simple}. Then, the camera captures the target from different viewpoints and the checkerboard corners are detected for calculating the camera parameters. However, such a procedure requires cumbersome manual interactions and it cannot achieve automatic calibration ``in the wild''. To pursue better flexibility, the second category of camera calibration, \textit{i.e.}, the geometric-prior-based calibration has been largely studied \cite{barreto2005geometric, Carroll, Bukhari, Miguel}. To be specific, the geometric structures are leveraged to model the 3D-2D correspondence in the scene, such as lines and vanishing points. However, this type of method heavily relies on structured man-made scenes containing rich geometric priors, leading to poor performance when applied to general environments. The third category is self-calibration\cite{faugeras1992camera, fraser1997digital, hartley1994self}. Such a solution takes a sequence of images as inputs and estimates the camera parameters using multi-view geometry. The accuracy of self-calibration, however, is constrained by the limits of the feature detectors, which can be influenced by diverse lighting conditions and textures. The above parametric models are based on the physical interpretation of camera geometry. While they are user-friendly, they tend to be tailored for specific camera models and may not offer optimal accuracy. Instead, non-parametric models~\cite{camposeco2015non, schops2020having, pan2022camera} link each image pixel to its 3D observation ray, eliminating the constraints of parametric models.

Since there are many standard techniques for calibrating cameras in an industry/laboratory implementation\cite{opencv, matlab}, this process is usually ignored in recent developments. However, calibrating single and wild images remains challenging, especially when images are collected from websites and unknown camera models. This challenge motivates the researchers to investigate a new paradigm.

Recently, deep learning has brought new inspirations to camera calibration and its applications. Learning-based methods achieve state-of-the-art performances on various tasks with higher efficiency. Particularly, diverse deep neural networks have been developed, such as convolutional neural networks (CNNs), generative adversarial networks (GANs), vision transformers (ViTs), and diffusion models, of which the high-level semantic features show more powerful representation capability compared with the hand-crafted features. Moreover, diverse learning strategies have been exploited to boost the geometric perception of networks. Learning-based methods offer a flexible and end-to-end camera calibration solution, without manual interventions or calibration targets, which sets them apart from traditional methods. Furthermore, some of these methods achieve camera model-free and label-free calibration, showing promising and meaningful applications.

With the rapid increase in the number of learning-based camera calibration methods, it has become increasingly challenging to keep up with new advances. Consequently, there is an urgent need to analyze existing works and foster a community dedicated to this field. Previously, some surveys, \textit{e.g.}, \cite{salvi2002comparative, hughes2008review, fan2022wide} only focused on a specific task/camera in camera calibration or one type of approach. For instance, Salvi et al. \cite{salvi2002comparative} reviewed the traditional camera calibration methods in terms of the algorithms. Hughes et al. \cite{hughes2008review} provided a detailed review for calibrating fisheye cameras with traditional solutions. While Fan et al. \cite{fan2022wide} discussed both the traditional methods and deep learning methods, their survey only considers calibrating the wide-angle cameras. In addition, due to the few amount of reviewed learning-based methods (around 10 papers), the readers are difficult to picture the development trend of general camera calibration in Fan et al. \cite{fan2022wide}. 

In this paper, we provide a comprehensive and in-depth overview of recent advances in learning-based camera calibration, covering over \textit{200} related papers. We also discuss potential directions for further improvements and examine various types of cameras and targets. To facilitate future research on different topics, we categorize the current solutions according to calibration objectives and applications. In addition to fundamental parameters such as focal length, rotation, and translation, we also provide detailed reviews for correcting image distortion (radial distortion and rolling shutter distortion), estimating cross-view mapping, calibrating camera-LiDAR systems, and other applications. Such a trend follows the development of cameras and market demands for virtual reality, autonomous driving, neural rendering, etc. 

To our best knowledge, this is the first survey of the learning-based camera calibration and its extended applications, it has the following unique contributions. (1) Our work mainly follows recent advances in deep learning-based camera calibration. In-depth analysis and discussion in various aspects are offered, including publications, network architecture, loss functions, datasets, evaluation metrics, learning strategies, implementation platforms, etc. The detailed information of each literature is listed in Table \ref{table:methods:standard:distortion} and Table \ref{table:methods:cross-view:cross-sensor}. (2) Despite the calibration algorithm, we comprehensively review the classical camera models and their extended models. In particular, we summarize the redesigned calibration objectives in deep learning since some traditional calibration objectives are verified to be hard to learn by neural networks. (3) We collect a dataset containing images and videos captured by different cameras in different environments, which can serve as a platform to evaluate the generalization of existing methods. (4) We discuss the open challenges in learning-based camera calibration and propose future directions to guide further research in this field. (5) An open-source repository is created that provides a taxonomy of all reviewed works and benchmarks. The repository will be updated regularly in \url{https://github.com/KangLiao929/Awesome-Deep-Camera-Calibration}.

In the following sections, we discuss and analyze various aspects of learning-based camera calibration. The remainder of this paper is organized as follows. In Section~\ref{sec2}, we provide the concrete learning paradigms and learning strategies of the learning-based camera calibration. Subsequently, we introduce and discuss the specific methods based on the standard camera model, distortion model, cross-view model, and cross-sensor model in Section~\ref{sec:pure}, Section~\ref{sec:distortion}, Section~\ref{sec:projection}, and Section~\ref{sec:hybrid}, respectively (see Figure~\ref{fig:taxonomy}). The collected benchmark for calibration methods is depicted in Section~\ref{sec:evaluation}. Finally, we conclude the learning-based camera calibration and suggest the future directions of this community in Section~\ref{sec:Future}.
	\section{Preliminaries}
\label{sec2}
Deep learning has brought new inspirations to camera calibration, enabling a fully automatic calibration procedure without manual intervention. Here, we first summarize two prevalent paradigms in learning-based camera calibration: regression-based calibration and reconstruction-based calibration. Then, the widely-used learning strategies are reviewed in this research field. The detailed definitions for classical camera models and their corresponding calibration objectives are exhibited in the supplementary material.

\subsection{Learning Paradigm}
Driven by different architectures of the neural network, the researchers have developed two main paradigms for learning-based camera calibration and its applications.

\noindent \textbf{Regression-based Calibration}
Given an uncalibrated input, the regression-based calibration extracts the high-level semantic features using stacked convolutional layers. Then, the fully connected layers aggregate the semantic features and form a vector of the estimated calibration objective. The regressed parameters are used to conduct subsequent tasks such as distortion rectification, image warping, camera localization, etc. This paradigm is the earliest and has a dominant role in learning-based camera calibration. All the first works in various objectives, \textit{e.g.}, intrinsics: Deepfocal \cite{DeepFocal}, extrinsic: PoseNet \cite{PoseNet}, radial distortion: Rong et al. \cite{Rong}, rolling shutter distortion: URS-CNN \cite{URS-CNN}, homography: DHN \cite{DHN}, hybrid parameters: Hold-Geoffroy et al. \cite{Hold-Geoffroy}, camera-LiDAR parameters: RegNet \cite{schneider2017regnet} have been achieved with this paradigm.

\noindent \textbf{Reconstruction-based Calibration}
On the other hand, the reconstruction-based calibration paradigm discards the parameter regression and directly learns the pixel-level mapping function between the uncalibrated input and target, inspired by the conditional image-to-image translation \cite{pix2pix} and dense visual perception\cite{long2015fully, eigen2014depth}. The reconstructed results are then calculated for the pixel-wise loss with the ground truth. In this regard, most reconstruction-based calibration methods \cite{DR-GAN, DDM, DaRecNet, BlindCor} design their network architecture based on the fully convolutional network such as U-Net\cite{ronneberger2015u}. Specifically, an encoder-decoder network, with skip connections between the encoder and decoder features at the same spatial resolution, progressively extracts the features from low-level to high-level and effectively integrates multi-scale features. At the last convolutional layer, the learned features are aggregated into the target channel, forming the calibrated result or calibration representation at the pixel level. Recent works also explore harnessing the powerful generation ability of the diffusion model to help reconstruct the calibration targets~\cite{CAR, DiffCalib, DM-Calib, RS-Diffusion}.

In contrast to the regression-based paradigm, the reconstruction-based paradigm does not require the label of diverse camera parameters during training. Besides, the imbalance loss problem can be eliminated since it only optimizes the photometric loss of calibrated results or calibration representations. Therefore, the reconstruction-based paradigm enables a blind camera calibration without a strong camera model assumption~\cite{camposeco2015non, schops2020having, pan2022camera}.

\subsection{Learning Strategies}
In the following, we review the learning-based camera calibration literature regarding different learning strategies.

\noindent \textbf{Supervised Learning}
Most learning-based camera calibration methods train their networks with the supervised learning strategy, from the classical methods \cite{DeepFocal, PoseNet, DHN, DeepVP, Rong, DeepCalib} to the state-of-the-art methods \cite{DVPD, EvUnroll, FishFormer, DAMG-Homo, SST-Calib, GeoCalib}. In terms of the learning paradigm, this strategy supervises the network with the ground truth of the camera parameters (regression-based paradigm) or paired data (reconstruction-based paradigm). In general, they synthesize the training dataset from other large-scale datasets, under the random parameter/transformation sampling and camera model simulation. Some recent works \cite{Zhao, Tan, SPEC, DeepUnrollNet} establish their training dataset using a real-world setup and label the captured images with manual annotations, thereby fostering advancements in this research domain.

\noindent \textbf{Semi-Supervised Learning}
Training the network using an annotated dataset under diverse scenarios is an effective learning strategy. However, human annotation can be prone to errors, leading to inconsistent annotation quality or the inclusion of contaminated data. Consequently, increasing the dataset to improve performance can be challenging due to the complexity and construction cost. To address this challenge, SS-WPC\cite{SS-WPC} proposes a semi-supervised method for correcting portraits captured by a wide-angle camera. It employs a surrogate segmentation task and a semi-supervised method that utilizes direction and range consistency and regression consistency to leverage both labeled and unlabeled data.

\noindent \textbf{Weakly-Supervised Learning}
Although significant progress has been made, data labeling for camera calibration is a notorious costly process, and obtaining perfect ground-truth labels is challenging. As a result, it is often preferable to use weak supervision with machine learning methods. Weakly supervised learning refers to the process of building prediction models through learning with inadequate supervision. Zhu et al. \cite{Zhu} present a weakly supervised camera calibration method for single-view metrology in unconstrained environments, where there is only one accessible image of a scene composed of objects of uncertain sizes. This work leverages 2D object annotations from large-scale datasets, where people and buildings are frequently present and serve as useful ``reference objects'' for determining 3D size.

\begin{figure*}[!t]
  \centering
  \includegraphics[width=1\textwidth]{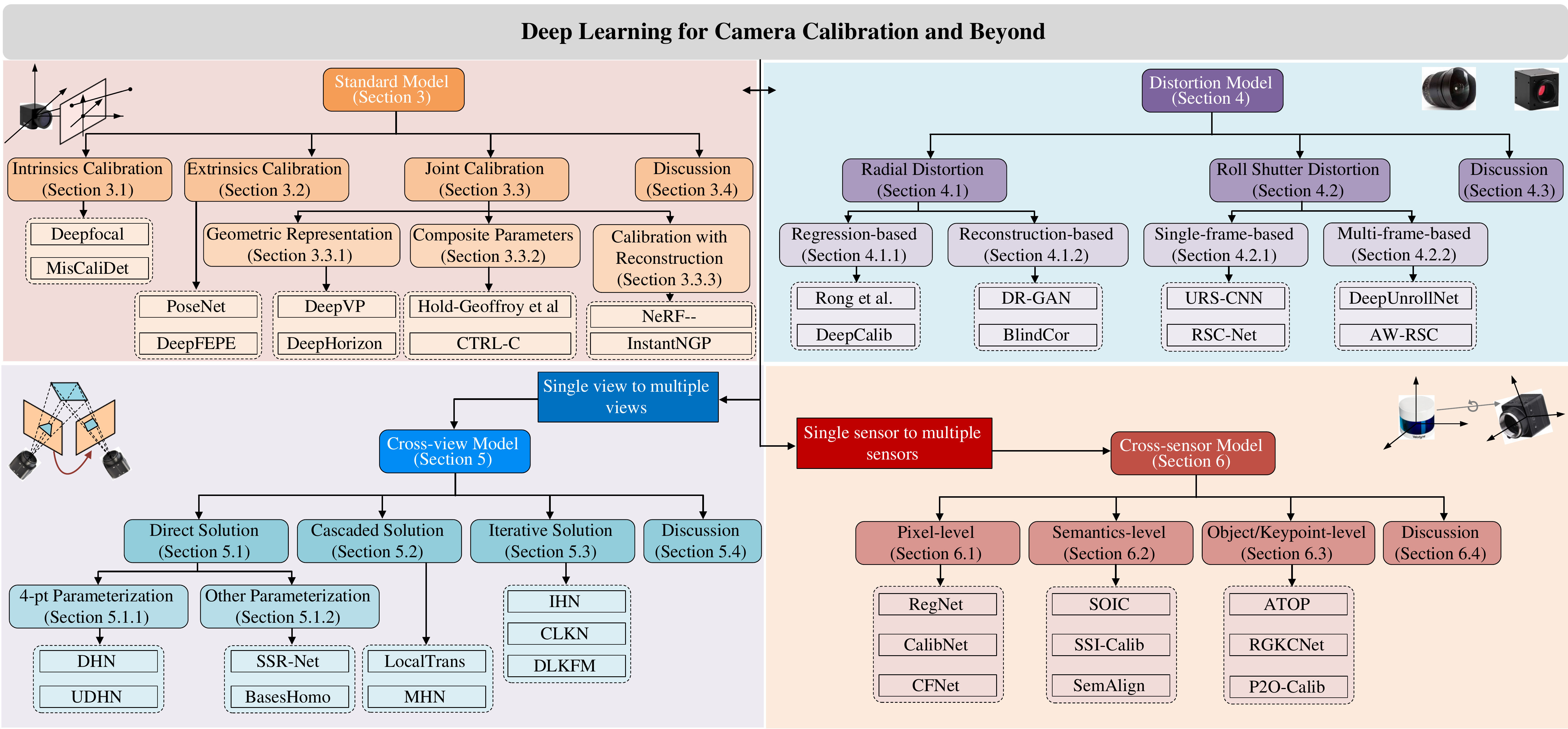}
  \caption{The structural and hierarchical taxonomy of camera calibration with deep learning. Some classical methods are listed under each category.}
  \label{fig:taxonomy}
  \vspace{-0.2cm}
\end{figure*}

\noindent \textbf{Unsupervised Learning}
Unsupervised learning analyzes and groups unlabeled datasets using machine learning algorithms. UDHN \cite{UDHN} is the first work for the cross-view camera model using unsupervised learning, which estimates the homography matrix of a paired image without labels. By reducing a pixel-wise error that does not require ground truth data, UDHN \cite{UDHN} outperforms previous supervised learning techniques. While preserving superior accuracy and robustness to light fluctuations, it can also achieve faster inference time. Inspired by this work, increasing more methods leverage the unsupervised learning strategy to estimate the cross-view mapping such as CA-UDHN \cite{CA-UDHN}, BaseHomo \cite{BasesHomo}, HomoGAN\cite{HomoGAN}, and Liu et al. \cite{Liu}. Besides, UnFishCor \cite{UnFishCor} frees the demands for distortion parameters and designs an unsupervised framework for the wide-angle camera.

\noindent \textbf{Self-supervised Learning}
Robotics is where the phrase ``self-supervised learning'' first appears, as training data is automatically categorized by utilizing relationships between various input sensor signals~\cite{rani2023self}. Compared to supervised learning, self-supervised learning leverages input data itself as the supervision. Many self-supervised techniques are presented to learn visual characteristics from massive amounts of unlabeled photos or videos without the need for time-consuming and expensive human annotations. SSR-Net \cite{SSR-Net} presents a self-supervised deep homography estimation network, which relaxes the need for ground truth annotations and leverages the invertibility constraints of homography. To be specific, SSR-Net \cite{SSR-Net} utilizes the homography matrix representation in place of other approaches' typically-used 4-point parameterization, to apply the invertibility constraints. SIR \cite{SIR} devises a brand-new self-supervised pipeline for wide-angle image rectification, based on the principle that the corrected results of distorted images of the same scene captured by various lenses, should obtain identical rectified results. Using self-supervised depth and pose learning as a proxy task, Fang et al. \cite{Fang} propose a method for self-calibrating a range of generic camera models from raw video, providing the first calibration evaluation of camera model parameters learned entirely through self-supervision.

\noindent \textbf{Reinforcement Learning}
Instead of aiming to minimize at each stage, reinforcement learning can maximize the cumulative benefits of a learning process as a whole. To date, DQN-RecNet~\cite{DQN-RecNet} is the first work in camera calibration using reinforcement learning. It applies a deep reinforcement learning technique to tackle the fisheye image rectification by a single Markov Decision Process, which is a multi-step gradual calibration scheme. In this situation, the current fisheye image represents the state of the environment. The agent, Deep Q-Network \cite{mnih2015human}, generates an action that should be executed to correct the distorted image.

In the following, we will review the specific methods and literature for learning-based camera calibration. The structural and hierarchical taxonomy is shown in Figure~\ref{fig:taxonomy}. 
	
	\begin{table*}
		\rowcolors{1}{gray!20}{white}
		\centering
		\caption{
			{Details of the learning-based camera calibration for \textbf{the standard and distortion camera models} and the extended applications, including the method abbreviation, publication, calibration objective, network architecture, loss function, dataset, evaluation metrics, learning strategy, platform, and simulation or not (training data). For the learning strategies, SL, USL, WSL, Semi-SL, SSL, and RL denote supervised learning, unsupervised learning, weakly-supervised learning, semi-supervised learning, self-supervised learning, and reinforcement learning.}
		}
		\vspace{-6pt}
		\label{table:methods:standard:distortion}
		\begin{threeparttable}
			\resizebox{1.0\textwidth}{!}{
				\setlength\tabcolsep{2pt}
				\renewcommand\arraystretch{1}
				\begin{tabular}{c|r||c|c|c|c|c|c|c|c|c}
					\hline
					&\textbf{Method}~~~~~~~~~&\textbf{Publication} &\textbf{Objective} &\textbf{Network}
					&\textbf{Loss Function} & \textbf{Dataset} &\textbf{Evaluation} & \textbf{Learning} &\textbf{Platform} &\textbf{Simulation}\\
					\hline
					\hline
					\multirow{1}{*}{\rotatebox{0}{\textbf{2015}}}
					&DeepFocal~\cite{DeepFocal} &ICIP &Standard &AlexNet
					&$\mathcal{L}_2$ loss &1DSfM\cite{1DSfM} & Accuracy & SL &Caffe &\\
					&PoseNet~\cite{PoseNet} &ICCV &Standard 
					&GoogLeNet
					&$\mathcal{L}_2$ loss &Cambridge Landmarks\cite{Cambridge_Landmarks} &Accuracy &SL &Caffe& \\
					\hline
					\hline
					\multirow{1}{*}{\rotatebox{0}{\textbf{2016}}}
					&DeepHorizon~\cite{DeepHorizon} &BMVC &Standard &GoogLeNet	&Huber loss &HLW\cite{HLW} & Accuracy & SL &Caffe &\\
					
					&DeepVP~\cite{DeepVP} &CVPR &Standard 
					&AlexNet
					&Logistic loss &YUD\cite{YUD}, ECD\cite{ECD}, HLW\cite{HLW} &Accuracy &SL &Caffe& \\	
					
					&Rong et al.~\cite{Rong} &ACCV &Distortion &AlexNet
					&Softmax loss &ImageNet\cite{ImageNet} &Line length &SL &Caffe&\checkmark\\	
                        \hline
					\hline
					\multirow{1}{*}{\rotatebox{0}{\textbf{2017}}}
					
					&URS-CNN~\cite{URS-CNN} &CVPR &Distortion 
					&CNNs
					&$\mathcal{L}_2$ loss &Sun\cite{xiao2010sun}, Oxford\cite{philbin2007object}, Zubud\cite{shao2003zubud}, LFW\cite{huang2008labeled} &PSNR, RMSE &SL &Torch&\checkmark\\
					
					\hline
					\hline
					\multirow{1}{*}{\rotatebox{0}{\textbf{2018}}}
					&Hold-Geoffroy et al.~\cite{Hold-Geoffroy} &CVPR &Standard &DenseNet	&Entropy loss &SUN360\cite{SUN360} & Human sensitivity & SL &- &\\

                        &Chang et al.\cite{chang2018deepvp} &ICRA &Standard &AlexNet
					&Cross-entropy loss &DeepVP-1M~\cite{chang2018deepvp} &MSE, Accuracy &SL &Matconvnet&\\
					
					&DeepCalib~\cite{DeepCalib} &CVMP &Distortion 
					&Inception-V3
					&Logcosh loss &SUN360\cite{SUN360} &Mean error &SL &TensorFlow&\checkmark \\	
					
                        &FishEyeRecNet~\cite{FishEyeRecNet} &ECCV &Distortion &VGG
					&$\mathcal{L}_2$ loss &ADE20K\cite{ADE20K} &PSNR, SSIM &SL &Caffe&\checkmark\\
					
					&Shi et al.\cite{Shi} &ICPR &Distortion &ResNet
					&$\mathcal{L}_2$ loss &ImageNet\cite{ImageNet} &MSE &SL &PyTorch&\checkmark\\
					
					\hline
					\hline
					\multirow{1}{*}{\rotatebox{0}{\textbf{2019}}}
					
					&UprightNet~\cite{UprightNet} &ICCV &Standard &U-Net	&Geometry loss &InteriorNet\cite{InteriorNet}, ScanNet\cite{ScanNet}, SUN360\cite{SUN360} &Mean error & SL &PyTorch &\\

					&NeurVPS~\cite{zhou2019neurvps} &NeurIPS &Standard &CNNs	&Binary cross entropy, chamfer-$\mathcal{L}_2$ loss &ScanNet~\cite{ScanNet}, SU3~\cite{SU3} &Angle accuracy &SL &PyTorch &\\

					&Deep360Up~\cite{Deep360Up} &VR &Standard &DenseNet	&Log-cosh loss\cite{Log-cosh} & SUN360\cite{SUN360} &Mean error & SL &- &\checkmark\\

					&Lopez et al.~\cite{Lopez} &CVPR &Distortion &DenseNet	&Bearing loss &SUN360\cite{SUN360} &MSE & SL &PyTorch &\\
					
					&Zhuang et al.~\cite{Zhuang} &IROS &Distortion &ResNet	&$\mathcal{L}_1$ loss & KITTI\cite{KITTI} &Mean error, RMSE & SL &PyTorch &\checkmark\\
					
					&DR-GAN~\cite{DR-GAN} &TCSVT &Distortion &GANs	&Perceptual loss & MS-COCO\cite{MS-COCO} &PSNR, SSIM & SL &TensorFlow &\checkmark\\
					
					&STD~\cite{STD} &TCSVT &Distortion &GANs+CNNs	&Perceptual loss & MS-COCO\cite{MS-COCO} &PSNR, SSIM & SL &TensorFlow &\checkmark\\
					
					&UnFishCor~\cite{UnFishCor} &JVCIR &Distortion &VGG	&$\mathcal{L}_1$ loss & Places2\cite{Places2} &PSNR, SSIM & USL &TensorFlow &\checkmark\\
					
					&BlindCor~\cite{BlindCor} &CVPR &Distortion &U-Net	&$\mathcal{L}_2$ loss & Places2\cite{Places2} &MSE & SL &PyTorch &\checkmark\\
					
					&RSC-Net~\cite{RSC-Net} &CVPR &Distortion &ResNet	&$\mathcal{L}_1$ loss & KITTI\cite{KITTI} &Mean error & SL &PyTorch &\checkmark\\
					
					&Xue et al.~\cite{Xue} &CVPR &Distortion &ResNet	&$\mathcal{L}_2$ loss & Wireframes\cite{Wireframes}, SUNCG\cite{SUNCG} &PSNR, SSIM, RPE & SL &PyTorch &\checkmark\\
					
					&Zhao et al.~\cite{Zhao} &ICCV &Distortion &VGG+U-Net	&$\mathcal{L}_1$ loss & Self-constructed+BU-4DFE\cite{BU-4DFE} &Mean error &SL &- &\checkmark\\

					\hline
					\hline
					\multirow{1}{*}{\rotatebox{0}{\textbf{2020}}}
				    
				    &Lee et al.~\cite{Lee} &ECCV &Standard &PointNet+CNNs	& Cross-entropy loss &Google Street View\cite{googleStreet}, HLW\cite{HLW} &Mean error, AUC\cite{AUC} & SL &- &\\

				    &Baradad et al.~\cite{Baradad} &CVPR &Standard &CNNs	&$\mathcal{L}_2$ loss &ScanNet\cite{ScanNet}, NYU\cite{NYU}, SUN360\cite{SUN360} &Mean error, RMS &SL &PyTorch &\\
				    
				    &Zheng et al.~\cite{Zheng} &CVPR &Standard &CNNs	&$\mathcal{L}_1$ loss &FocaLens\cite{FocaLens} &Mean error, PSNR, SSIM &SL &- &\checkmark\\
				    
				    &Zhu et al.~\cite{Zhu} &ECCV &Standard &CNNs+PointNet	&$\mathcal{L}_1$ loss &SUN360\cite{SUN360}, MS-COCO\cite{MS-COCO} &Mean error, Accuracy &WSL &PyTorch &\checkmark\\

				    &Davidson et al.~\cite{Davidson} &ECCV &Standard &FCN	&Dice loss &SUN360\cite{SUN360} &Accuracy &SL &- &\checkmark\\
				    
				    &DeepFEPE~\cite{DeepFEPE} &IROS &Standard &VGG+PointNet	&$\mathcal{L}_2$ loss &KITTI\cite{KITTI}, ApolloScape\cite{Apolloscape} &Mean error &SL &PyTorch &\\
				    
				    &MisCaliDet~\cite{MisCaliDet} &ICRA &Distortion &CNNs	& $\mathcal{L}_2$ loss &KITTI\cite{KITTI} &MSE & SL &TensorFlow &\checkmark\\
				    
				    &DeepPTZ~\cite{DeepPTZ} &WACV &Distortion &Inception-V3	& $\mathcal{L}_1$ loss &SUN360\cite{SUN360} &Mean error & SL &PyTorch &\checkmark\\
				    
				    &DDM~\cite{DDM} &TIP &Distortion &GANs	&$\mathcal{L}_1$ loss &MS-COCO\cite{MS-COCO} &PSNR, SSIM &SL &TensorFlow &\checkmark\\
				    
				    &Li et al.~\cite{Li} &TIP &Distortion &CNNs	&Cross-entropy, $\mathcal{L}_1$ loss &CelebA\cite{CelebA} &Cosine distance &SL &- &\checkmark\\
				    
				    &PSE-GAN~\cite{PSE-GAN} &ICPR &Distortion &GANs	&$\mathcal{L}_1$, WGAN loss &Place2\cite{Places2} &MSE &SL &- &\checkmark\\
				    
				    &RDC-Net~\cite{RDC-Net} &ICIP &Distortion &ResNet	&$\mathcal{L}_1$, $\mathcal{L}_2$ loss &ImageNet\cite{ImageNet} &PSNR, SSIM &SL &PyTorch &\checkmark\\
				    
				    &FE-GAN~\cite{FE-GAN} &ICASSP &Distortion &GANs	&$\mathcal{L}_1$, GAN loss &Wireframe\cite{Wireframes}, LSUN\cite{LSUN} &PSNR, SSIM, RMSE &SSL &PyTorch &\checkmark\\
				    
				    &RDCFace~\cite{RDCFace} &CVPR &Distortion &ResNet	&Cross-entropy, $\mathcal{L}_2$ loss &IMDB-Face\cite{IMDB-Face} &Accuracy &SL &- &\checkmark\\
				    
				    &LaRecNet~\cite{LaRecNet} &arXiv &Distortion &ResNet	&$\mathcal{L}_2$ loss &Wireframes\cite{Wireframes}, SUNCG\cite{SUNCG} &PSNR, SSIM, RPE &SL &PyTorch &\checkmark\\
				    
				    &DeepUnrollNet~\cite{DeepUnrollNet} &CVPR &Distortion &FCN	&$\mathcal{L}_1$, perceptual, total variation loss &Carla-RS\cite{DeepUnrollNet}, Fastec-RS\cite{DeepUnrollNet}  &PSNR, SSIM &SL &PyTorch &\checkmark\\
                        
					\hline
					\hline
					\multirow{1}{*}{\rotatebox{0}{\textbf{2021}}}
					&StereoCaliNet~\cite{StereoCaliNet} &TCI &Standard &U-Net	&$\mathcal{L}_1$ loss &TAUAgent\cite{TAUAgent}, KITTI\cite{KITTI} &Mean error & SL &PyTorch &\checkmark\\
					
					&CTRL-C~\cite{CTRL-C} &ICCV &Standard &Transformer	&Cross-entropy, $\mathcal{L}_1$ loss &Google Street View\cite{googleStreet}, SUN360\cite{SUN360} &Mean error, AUC\cite{AUC} & SL &PyTorch &\checkmark\\

				   &SA-MobileNet~\cite{SA-MobileNet} &BMVC &Standard &MobileNet &Cross-entropy loss& SUN360\cite{SUN360}, ADE20K\cite{ADE20K}, NYU\cite{NYU} &MAE, Accuracy & SL &TensorFlow &\checkmark \\
				   
				   &SPEC~\cite{SPEC} &ICCV &Standard &ResNet &Softargmax-$\mathcal{L}_2$ loss&Self-constructed &W-MPJPE, PA-MPJPE & SL &PyTorch &\checkmark \\
				   
				   &DirectionNet~\cite{DirectionNet} &CVPR &Standard &U-Net &Cosine similarity loss &InteriorNet\cite{InteriorNet}, Matterport3D\cite{Matterport3D}&Mean and median error  & SL &TensorFlow &\checkmark \\
					
				   &Wakai et al.~\cite{Wakai} &ICCVW &Distortion &DenseNet	&Smooth $\mathcal{L}_1$ loss &StreetLearn\cite{StreetLearn} &Mean error, PSNR, SSIM & SL &- &\checkmark\\
				   &OrdianlDistortion~\cite{OrdianlDistortion} &TIP &Distortion &CNNs	&Smooth $\mathcal{L}_1$ loss & MS-COCO\cite{MS-COCO} &PSNR, SSIM, MDLD & SL &TensorFlow &\checkmark\\
				   
				   &PolarRecNet~\cite{PolarRecNet} &TCSVT &Distortion &VGG+U-Net	&$\mathcal{L}_1$, $\mathcal{L}_2$ loss & MS-COCO\cite{MS-COCO}, LMS\cite{LMS} &PSNR, SSIM, MSE & SL &PyTorch &\checkmark\\
				   
				   &DQN-RecNet~\cite{DQN-RecNet} &PRL &Distortion &VGG	&$\mathcal{L}_2$ loss & Wireframes\cite{Wireframes} &PSNR, SSIM, MSE & RL &PyTorch &\checkmark\\
				   
				   &Tan et al.~\cite{Tan} &CVPR &Distortion &U-Net &$\mathcal{L}_2$ loss & Self-constructed &Accuracy & SL &PyTorch & \\
				   
				   &PCN~\cite{PCN} &CVPR &Distortion &U-Net &$\mathcal{L}_1$, $\mathcal{L}_2$, GAN loss & Place2\cite{Places2} &PSNR, SSIM, FID, CW-SSIM & SL &PyTorch &\checkmark \\
				   
				   &DaRecNet~\cite{DaRecNet} &ICCV &Distortion &U-Net &Smooth $\mathcal{L}_1$, $\mathcal{L}_2$ loss & ADE20K\cite{ADE20K} &PSNR, SSIM & SL &PyTorch &\checkmark \\
				   
				   &JCD~\cite{JCD} &CVPR &Distortion &FCN &Charbonnier\cite{Charbonnier}, perceptual loss &BS-RSCD \cite{JCD}, Fastec-RS
                   \cite{DeepUnrollNet}&PSNR, SSIM, LPIPS  & SL &PyTorch & \\

                   &Fan\etal~\cite{fan2021inverting} &ICCV &Distortion &U-Net &$\mathcal{L}_1$, perceptual loss &Carla-RS~\cite{DeepUnrollNet}, Fastec-RS~\cite{DeepUnrollNet} &PSNR, SSIM, LPIPS  & SL &PyTorch & \\

                   &SUNet~\cite{SUNet} &ICCV &Distortion &DenseNet+ResNet &$\mathcal{L}_1$, perceptual loss &Carla-RS~\cite{DeepUnrollNet}, Fastec-RS~\cite{DeepUnrollNet} &PSNR, SSIM  & SL &PyTorch & \\
       
				   \hline
				   \hline
				   \multirow{1}{*}{\rotatebox{0}{\textbf{2022}}}
				   &DVPD~\cite{DVPD} &CVPR &Standard &CNNs	&Cross-entropy loss &SU3\cite{SU3}, ScanNet\cite{ScanNet}, YUD\cite{YUD}, NYU\cite{NYU} &Accuracy, AUC\cite{AUC} & SL &PyTorch &\checkmark\\
				   
				   &Fang et al.~\cite{Fang} &ICRA &Standard &CNNs	&$\mathcal{L}_2$ loss &KITTI\cite{KITTI}, EuRoC\cite{EuRoC}, OmniCam\cite{OmniCam} &MRE, RMSE & SSL &PyTorch &\\
				   
				   &CPL~\cite{CPL} &ICASSP &Standard &Inception-V3	&$\mathcal{L}_1$ loss &CARLA\cite{CARLA}, CyclistDetection\cite{CyclistDetection} &MAE & SL &TensorFlow &\checkmark\\
				   
				   &Do et al.~\cite{Do} &CVPR &Standard  &ResNet&$\mathcal{L}_2$, Robust angular \cite{RobustAngular} loss &Self-constructed, 7-SCENES\cite{7-SCENES} &Median error, Recall &SL &PyTorch &\\
				   
				   &DiffPoseNet~\cite{DiffPoseNet} &CVPR &Standard  &CNNs+LSTMs&$\mathcal{L}_2$ loss &TartanAir\cite{TartanAir}, KITTI\cite{KITTI}, TUM RGB-D\cite{TUM_RGB-D} &PEE, AEE\cite{AEE} &SSL &PyTorch &\\
				   
				   &SceneSqueezer~\cite{SceneSqueezer} &CVPR &Standard  &Transformer&$\mathcal{L}_1$ loss &RobotCar Seasons\cite{RobotCar}, Cambridge Landmarks\cite{Cambridge_Landmarks}  &Mean error, Recall\cite{AEE} &SL &PyTorch &\\
				   
				   &FocalPose~\cite{FocalPose} &CVPR &Standard  &CNNs&$\mathcal{L}_1$, Huber loss &Pix3D\cite{Pix3D}, CompCars\cite{StanfordCars}, StanfordCars\cite{StanfordCars}  &Median error, Accuracy &SL &PyTorch &\\

   				   &SS-WPC~\cite{SS-WPC} &CVPR &Distortion  &Transformer	&Cross-entropy, $\mathcal{L}_1$ loss &Tan et al.\cite{Tan} &Accuracy & Semi-SL &PyTorch &\\
				   
				   &AW-RSC~\cite{AW-RSC} &CVPR &Distortion  &CNNs	&Charbonnier\cite{Charbonnier}, perceptual loss &Self-constructed, FastecRS\cite{DeepUnrollNet} &PSNR, SSIM &SL &PyTorch &\\
				   
				   &EvUnroll~\cite{EvUnroll} &CVPR &Distortion  &U-Net	&Charbonnier, perceptual, TV loss &Self-constructed, FastecRS\cite{DeepUnrollNet} &PSNR, SSIM, LPIPS &SL &PyTorch &\\
				   
				   &CCS-Net~\cite{zhang2022learning} &RAL &Distortion  &U-Net&$\mathcal{L}_1$ loss &TUM RGB-D\cite{TUM_RGB-D} &MAE, RPE &SL &PyTorch &\checkmark\\
				   
				   &FishFormer~\cite{FishFormer} &arXiv &Distortion  &Transformer&$\mathcal{L}_2$ loss &Place2\cite{Places2}, CelebA\cite{CelebA}  &PSNR, SSIM, FID &SL &PyTorch &\checkmark\\
				   
                  &SIR~\cite{SIR} &TIP &Distortion &ResNet &$\mathcal{L}_1$ loss & ADE20K\cite{ADE20K}, WireFrames\cite{Wireframes}, MS-COCO\cite{MS-COCO} &PSNR, SSIM & SSL &PyTorch &\checkmark \\

                    &GenCaliNet~\cite{GenCaliNet} &ECCV &Distortion &DenseNet	&$\mathcal{L}_2$ loss &StreetLearn\cite{StreetLearn}, SP360\cite{SP360} &MAE, PSNR, SSIM & SL &- &\checkmark\\

				   &IFED~\cite{IFED} &ECCV &Distortion  &CNNs	&Charbonnier, perceptual, TV loss &RS-GOPRO~\cite{IFED}, FastecRS\cite{DeepUnrollNet} &PSNR, SSIM, LPIPS &SL &PyTorch &\checkmark\\

                      \hline
				   \hline
				   \multirow{1}{*}{\rotatebox{0}{\textbf{2023}}}

					&PM-Calib~\cite{PM-Calib} &TPAMI &Standard &CNNs	&Kullback-Leibler loss &360Cities~\cite{360Cities} &Mean error, Human sensitivity & SL &- &\checkmark\\

					&PerspectiveField~\cite{PerspectiveField} &CVPR &Standard &CNNs+Transformers	&Cross-entropy loss &360Cities~\cite{360Cities}, Stanford2D3D~\cite{Stanford2D3D}, TartanAir~\cite{TartanAir} &Mean error & SL &PyTorch &\checkmark\\

					&Orienternet~\cite{Orienternet} &CVPR &Standard &CNNs	&Log-likelihood loss &MGL~\cite{Orienternet} &Positions and rotation errors & SL &PyTorch &\\

					&Neumap~\cite{Neumap} &CVPR &Standard &CNNs+Transformers	&$\mathcal{L}_2$, cross-entropy loss &Cambridge Landmarks~\cite{Cambridge_Landmarks}, ScanNet~\cite{ScanNet}, 7scenes~\cite{7-SCENES} &Median error & SL &PyTorch &\\

					&SESC~\cite{SESC} &IROS &Standard &CNNs	&$\mathcal{L}_1$ loss &KITTI~\cite{KITTI}, DDAD~\cite{DDAD} &Median error & SSL &PyTorch &\\

					&CROSSFIRE~\cite{CROSSFIRE} &ICCV &Standard &CNNs	&$\mathcal{L}_2$, TV loss &Cambridge Landmarks~\cite{Cambridge_Landmarks}, 7scenes~\cite{7-SCENES} &Median error & SSL &PyTorch &\\

					&WildCamera~\cite{WildCamera} &NeurIPS &Standard &CNNs	&Cosine similarity loss &ScanNet~\cite{ScanNet}, MegaDepth~\cite{MegaDepth}, KITTI~\cite{KITTI} &Median error & SL &PyTorch &\checkmark\\

					&DroidCalib~\cite{DroidCalib} &ICCV &Standard &CNNs	&$\mathcal{L}_1$ loss &TartanAir~\cite{TartanAir}, EuRoC~\cite{EuRoC}, TUM RGB-D~\cite{TUM_RGB-D} &Median error & SL &PyTorch &\checkmark\\

					&NeuralLens~\cite{NeuralLens} &CVPR &Distortion &CNNs	&$\mathcal{L}_2$ loss &SynLens~\cite{NeuralLens} &RMS & SL &PyTorch &\checkmark\\
                    
				   &DDA~\cite{DDA} &ICCV &Distortion &Diffusion	&Diffusion loss &Places2~\cite{Places2}, Woodscape~\cite{yogamani2019woodscape} &PSNR, SSIM, MS-SSIM, FID, LPIPS & USL &PyTorch &\checkmark\\

				   &CACM-Net~\cite{CACM-Net} &CVPR &Distortion &CNNs	&$\mathcal{L}_2$ loss &Wireframes\cite{Wireframes} &StraightAcc, ShapeAcc, ConformalAcc & SL &- &\checkmark\\

				   &RDTR~\cite{RDTR} &TIP &Distortion &CNNs+Transformers	&Cross-entropy, $\mathcal{L}_1$ loss &Places2~\cite{Places2} &EPE, PSNR, SSIM & SL &PyTorch &\checkmark\\

				   &DaFIR~\cite{DaFIR} &TCSVT &Distortion &Transformers	&$\mathcal{L}_1$, $\mathcal{L}_2$ loss &Places2~\cite{Places2} &PSNR, SSIM & SL &PyTorch &\checkmark\\
				   
				   &SimFIR~\cite{SimFIR} &ICCV &Distortion &Transformers	&Cross-entropy, $\mathcal{L}_1$ loss &Places2~\cite{Places2} &PSNR, SSIM, FID & SL &PyTorch &\checkmark\\

				   &RecRecNet~\cite{Recrecnet} &ICCV &Distortion &CNNs	&$\mathcal{L}_1$, $\mathcal{L}_2$ loss &MS-COCO\cite{MS-COCO} &PSNR, SSIM, FID, LPIPS & SL &PyTorch &\checkmark\\

				   &SDP-Net~\cite{SDP-Net} &AAAI &Distortion &CNNs	&$\mathcal{L}_1$, $\mathcal{L}_2$ loss &DAVIS~\cite{perazzi2016benchmark}, Youtube-VOS~\cite{xu2018youtube} &PSNR, SSIM, FID, EPE & SL &PyTorch &\checkmark\\

				   &Darswin~\cite{Darswin} &ICCV &Distortion &Transformers	&Cross-entropy loss &ImageNet~\cite{ImageNet} &Accuracy & SL &PyTorch &\checkmark\\

				   &REG-Net~\cite{REG-Net} &ACM MM &Distortion &CNNs	&$\mathcal{L}_2$ loss &REG-HDR~\cite{REG-Net} &PSNR, SSIM, LPIPS & SL &PyTorch &\\

				   &SelfDRSC~\cite{SelfDRSC} &ICCV &Distortion &CNNs	&$\mathcal{L}_2$ loss &RS-GOPRO~\cite{IFED} &PSNR, SSIM, LPIPS, NIQE, NRQM, PI & SSL &PyTorch &\checkmark\\

				   &SSL-RSC~\cite{SSL-RSC} &arXiv &Distortion &CNNs	&$\mathcal{L}_1$ loss &Gev-RS~\cite{EvUnroll}, Fastec-RS~\cite{DeepUnrollNet}, ERS-VFI~\cite{SSL-RSC} &PSNR, SSIM, LPIPS & SSL &- &\checkmark\\

				   &EvShutter~\cite{EvShutter} &CVPR &Distortion &CNNs	&$\mathcal{L}_1$ loss &RS-ERGB~\cite{EvShutter}, Fastec-RS~\cite{DeepUnrollNet} &PSNR, SSIM, LPIPS & SL &- &\\

				   &SelfUnroll~\cite{SelfUnroll} &arXiv &Distortion &CNNs	&$\mathcal{L}_1$, TV loss &RS-ERGB~\cite{EvShutter}, Fastec-RS~\cite{DeepUnrollNet}, Gev-RS~\cite{EvUnroll}, DRE~\cite{SelfUnroll} &PSNR, SSIM, LPIPS & SSL &PyTorch &\checkmark\\

				   &PatchNet~\cite{PatchNet} &WACV &Distortion &CNNs	&$\mathcal{L}_1$, TV loss &BS-RSCD~\cite{JCD}, Fastec-RS~\cite{DeepUnrollNet} &PSNR, SSIM, LPIPS & SL &PyTorch &\checkmark\\

				   &JAMNet~\cite{JAMNet} &CVPR &Distortion &CNNs+Transformers	&$\mathcal{L}_1$, TV loss &Carla-RS~\cite{DeepUnrollNet}, Fastec-RS~\cite{DeepUnrollNet}, BS-RSCD~\cite{JCD} &PSNR, SSIM, LPIPS & SL &PyTorch &\checkmark\\

				   &QRSC~\cite{QRSC} &ICCV &Distortion &Transformers	&$\mathcal{L}_1$, $\mathcal{L}_2$ loss &Carla-RS~\cite{DeepUnrollNet}, Fastec-RS~\cite{DeepUnrollNet}, BS-RSCD~\cite{JCD} &PSNR, SSIM, LPIPS & SL &PyTorch &\checkmark\\

				   &Deep\_HM~\cite{Deep_HM} &ICCV &Distortion &CNNs	&$\mathcal{L}_2$ loss &Carla-RS~\cite{DeepUnrollNet}, RS-Homo~\cite{Deep_HM} &PSNR, SSIM & SL &PyTorch &\checkmark\\

                      \hline
				   \hline
				   \multirow{1}{*}{\rotatebox{0}{\textbf{2024}}}

					&DM-Calib~\cite{DM-Calib} &arXiv &Standard &Diffusion	&Diffusion loss & NuScenes~\cite{caesar2020nuscenes}, KITTI~\cite{KITTI}, SUN3D~\cite{SUN3D}, CityScapes~\cite{CityScapes}, etc &Relative error& SL &PyTorch &\checkmark\\

					&GeoCalib~\cite{GeoCalib} &ECCV &Standard &CNNs	&$\mathcal{L}_1$ loss & OpenPano~\cite{GeoCalib} &Median error, AUC& SL &PyTorch &\checkmark\\

					&ExtremeRotation~\cite{ExtremeRotation} &arXiv &Standard &Transformers	&Cross-entropy loss & ExtremeLandmarkPairs~\cite{ExtremeRotation} &Geodesic error& SL &PyTorch &\\

					&GAT-Calib~\cite{GAT-Calib} &WACV &Standard &GNNs	&Cross-entropy, $\mathcal{L}_2$ loss & World Cup 2014~\cite{homayounfar2017sports} &IoU& SL &- &\checkmark\\

					&SOFI~\cite{SOFI} &BMVC &Standard &CNNs+Transformers	&Focal loss & Google Street View~\cite{googleStreet}, SUN360~\cite{SUN360}, HLW\cite{HLW} &Mean error, AUC& SL &PyTorch &\\

					&PWT-Calib~\cite{PWT-Calib} &WACV &Standard &CNNs	&$\mathcal{L}_2$ loss & CUTC~\cite{PWT-Calib} &Mean error, RMSE& SL &PyTorch &\\

					&NeFeS~\cite{NeFeS} &CVPR &Standard &CNNs+MLPs	&$\mathcal{L}_1$, cosine similarity loss & Cambridge Landmarks~\cite{Cambridge_Landmarks}, 7-Scenes~\cite{7-SCENES} &Median error& SL &PyTorch &\\

					&U-ARE-ME~\cite{U-ARE-ME} &arXiv &Standard &CNNs	&$\mathcal{L}_2$ loss & ICL-NUIM~\cite{ICL-NUIM}, TUM RGB-D~\cite{TUM_RGB-D}, ScanNet~\cite{ScanNet} &ARE& SL &PyTorch &\checkmark\\

					&FlowMap~\cite{FlowMap} &arXiv &Standard &CNNs	&$\mathcal{L}_1$ loss & MipNeRF-360~\cite{Barron_2021_ICCV}, Tanks \& Temples~\cite{TAT}, LLFF~\cite{LLFF}, CO3D~\cite{CO3D} &ATE& SL &PyTorch &\\

					&MSCC~\cite{MSCC} &WACV &Standard &CNNs+Transformers	&Cross-entropy, angular distance loss & Google Street View~\cite{googleStreet}, SYN-Citypark~\cite{SPEC}, Flickr~\cite{SPEC}, HLW~\cite{HLW} &Mean, median errors& SL &PyTorch &\checkmark\\

                  &HC-Net~\cite{HC-Net} &NeurIPS &Standard &CNNs	&infoNCE loss  &VIGOR~\cite{VIGOR}, KITTI~\cite{KITTI} &Mean, median errors & SL &Pytorch &\checkmark\\

					&DiffCalib~\cite{DiffCalib} &arXiv &Standard &Diffusion	&Diffusion loss & Hypersim~\cite{Hypersim}, NuScenes~\cite{caesar2020nuscenes}, KITTI~\cite{KITTI}, CitySpace~\cite{CityScapes}, NYUv2~\cite{NYU} &Relative error& SL &PyTorch &\checkmark\\

					&CAR~\cite{CAR} &ICLR &Standard &Diffusion	&Diffusion loss & CO3Dv2~\cite{CO3D} &Relative error& SL &PyTorch &\\
                    
				   &ADPs~\cite{ADPs} &CVPR &Distortion &CNNs	&$\mathcal{L}_2$ loss &StreetLearn\cite{StreetLearn}, SP360\cite{SP360} &Mean absolute error, REPE & SL &PyTorch &\checkmark\\
                   
				   &CDM~\cite{CDM} &TCSVT &Distortion &CNNs	&$\mathcal{L}_1$, $\mathcal{L}_2$ loss &MS-COCO~\cite{MS-COCO}, ADE20K~\cite{ADE20K}, Wireframes~\cite{Wireframes} &PSNR, SSIM & SL &PyTorch &\checkmark\\

				   &QueryCDR~\cite{QueryCDR} &ECCV &Distortion &CNNs+Transformers	&$\mathcal{L}_1$ loss &MS-COCO~\cite{MS-COCO}, Places2~\cite{Places2} &PSNR, SSIM & SL &PyTorch &\checkmark\\

				   &VACR~\cite{VACR} &ECCV &Distortion &CNNs	&$\mathcal{L}_2$ loss &KITTI-360~\cite{liao2022kitti}, StreetLearn\cite{StreetLearn}, Woodscape~\cite{yogamani2019woodscape}   &PSNR, SSIM, FID & SL &PyTorch &\checkmark\\

				   &DualPriorsCorrection~\cite{DualPriorsCorrection} &ECCV &Distortion &CNNs+GANs	&$\mathcal{L}_2$ loss &Tan~\cite{Tan}   &LineAcc, ShapeAcc & SL &PyTorch &\checkmark\\

				   &Disco~\cite{Disco} &IJCV &Distortion &GANs	&$\mathcal{L}_2$ loss &CMDP~\cite{fried2016perspective}, USCPP~\cite{Zhao}  &LMK-E, PSNR, SSIM, LIPIPS, ID & SL &- &\\

				   &MOWA~\cite{MOWA} &arXiv &Distortion &Transformers	&Cross-entroy, $\mathcal{L}_1$, $\mathcal{L}_2$ loss & StitchRect~\cite{StitchRect}, Place2~\cite{Places2}, MS-COCO~\cite{MS-COCO}, RotationCorr~\cite{RotationCorr}, Tan~\cite{Tan} &PSNR, SSIM, ShapeAcc & SL &PyTorch &\checkmark\\

				   &LBCNet~\cite{LBCNet} &TPAMI &Distortion &CNNs	&$\mathcal{L}_1$ loss &Carla-RS~\cite{DeepUnrollNet}, Fastec-RS~\cite{DeepUnrollNet} &PSNR, SSIM & SL &PyTorch &\checkmark\\

				   &TACA-Net~\cite{TACA-Net} &ACM MM &Distortion &Transformers	&$\mathcal{L}_1$ loss &Gev-RS~\cite{EvUnroll}, Fastec-RS~\cite{DeepUnrollNet} &PSNR, SSIM, LPIPS & SL &- &\checkmark\\

				   &UniINR~\cite{UniINR} &ECCV &Distortion &CNNs	&$\mathcal{L}_1$ loss &Gev-RS~\cite{EvUnroll}, Fastec-RS~\cite{DeepUnrollNet} &PSNR, SSIM & SL &PyTorch &\checkmark\\

				   &DFRSC~\cite{DFRSC} &CVPR &Distortion &CNNs	&$\mathcal{L}_1$ loss &Carla-RS~\cite{DeepUnrollNet}, Fastec-RS~\cite{DeepUnrollNet}, BS-RSCD~\cite{JCD} &PSNR, SSIM, LPIPS & SL &PyTorch &\checkmark\\

				   &RS-Diffusion~\cite{RS-Diffusion} &arXiv &Distortion &Diffusion	&Diffusion loss &RS-Homo~\cite{Deep_HM}, RS-Real~\cite{RS-Diffusion} &PSNR, SSIM, EPE & SL &PyTorch &\checkmark\\
                   
				\hline
				\end{tabular}
			}
		\end{threeparttable}
	\end{table*}

	\begin{table*}
		\rowcolors{1}{gray!20}{white}
		
		\caption{
			{Details of the learning-based camera calibration for \textbf{the cross-view and cross-sensor camera models} and the extended applications, including the method abbreviation, publication, calibration objective, network architecture, loss function, dataset, evaluation metrics, learning strategy, platform, and simulation or not (training data). For the learning strategies, SL, USL, WSL, Semi-SL, SSL, and RL denote supervised learning, unsupervised learning, weakly-supervised learning, semi-supervised learning, self-supervised learning, and reinforcement learning. }
		}
            \centering
		\vspace{-6pt}
		\label{table:methods:cross-view:cross-sensor}
		\begin{threeparttable}
			\resizebox{1\textwidth}{!}{
				\setlength\tabcolsep{2pt}
				\renewcommand\arraystretch{0.98}
				\begin{tabular}{c|r||c|c|c|c|c|c|c|c|c}
					\hline
					&\textbf{Method}~~~~~~~~~&\textbf{Publication} &\textbf{Objective} &\textbf{Network}
					&\textbf{Loss Function} & \textbf{Dataset} &\textbf{Evaluation} & \textbf{Learning} &\textbf{Platform} &\textbf{Simulation}\\
					\hline
					\hline
					\multirow{1}{*}{\rotatebox{0}{\textbf{2016}}}
					&DHN\cite{DHN} &RSSW &Cross-View &VGG
					&$\mathcal{L}_2$ loss &MS-COCO\cite{MS-COCO} &MSE &SL &Caffe&\checkmark\\		
                        \hline
					\hline
					\multirow{1}{*}{\rotatebox{0}{\textbf{2017}}}
					&CLKN~\cite{CLKN} &CVPR &Cross-View  &CNNs	&Hinge loss &MS-COCO\cite{MS-COCO} & MSE & SL &Torch &\checkmark\\
					
		            &HierarchicalNet~\cite{HierarchicalNet} &ICCVW &Cross-View 
					&VGG
					&$\mathcal{L}_2$ loss &MS-COCO\cite{MS-COCO} &MSE &SL &TensorFlow&\checkmark \\
					
					&RegNet~\cite{schneider2017regnet} &IV &Cross-Sensor 
					&CNNs
					&$\mathcal{L}_2$ loss &KITTI\cite{KITTI} &MAE &SL &Caffe&\checkmark\\
					
					\hline
					\hline
					\multirow{1}{*}{\rotatebox{0}{\textbf{2018}}}
					&DeepFM\cite{DeepFM} &ECCV &Cross-View &ResNet
					&$\mathcal{L}_2$ loss &T\&T\cite{TT}, KITTI\cite{KITTI}, 1DSfM\cite{1DSfM} &F-score, Mean &SL &PyTorch&\checkmark\\
					
					&Poursaeed et al.\cite{Poursaeed} &ECCVW &Cross-View &CNNs
					&$\mathcal{L}_1$, $\mathcal{L}_2$ loss &KITTI\cite{KITTI} &EPI-ABS, EPI-SQR &SL &-& \\
					
					&UDHN\cite{UDHN} &RAL &Cross-View &VGG
					&$\mathcal{L}_1$ loss &MS-COCO\cite{MS-COCO} &RMSE &USL &TensorFlow&\checkmark\\
					
					&PFNet\cite{PFNet} &ACCV &Cross-View &FCN
					&Smooth $\mathcal{L}_1$ loss &MS-COCO\cite{MS-COCO} &MAE &SL &TensorFlow&\checkmark\\
					
					&CalibNet\cite{iyer2018calibnet} &IROS &Cross-Sensor &ResNet
					&Point cloud distance, $\mathcal{L}_2$ loss &KITTI\cite{KITTI} &Geodesic distance, MAE &SL &TensorFlow&\checkmark\\
					
					\hline
					\hline
					\multirow{1}{*}{\rotatebox{0}{\textbf{2019}}}
					&SSR-Net~\cite{SSR-Net} &PRL &Cross-View &ResNet	&$\mathcal{L}_2$ loss & MS-COCO\cite{MS-COCO} &MAE & SSL &PyTorch &\checkmark\\
					
					&Abbas et al.~\cite{Abbas} &ICCVW &Cross-View &CNNs	&Softmax loss & CARLA\cite{CARLA} &AUC\cite{AUC}, Mean error & SL &TensorFlow &\checkmark\\
					
					\hline
					\hline
					\multirow{1}{*}{\rotatebox{0}{\textbf{2020}}}
					&Sha et al.~\cite{Sha} &CVPR &Cross-View &U-Net	& Cross-entropy loss &World Cup 2014\cite{homayounfar2017sports} &IoU & SL &TensorFlow &\\
				    
				    &MHN~\cite{MHN} &CVPR &Cross-View &VGG	&Cross-entropy loss &MS-COCO\cite{MS-COCO}, Self-constructed &MAE & SL &TensorFlow &\checkmark\\
				    
				    &CA-UDHN~\cite{CA-UDHN} &ECCV &Cross-View &FCN+ResNet	&Triplet loss &Self-constructed &MSE &USL &PyTorch &\\

				    &SRHEN~\cite{SRHEN} &ACM MM &Cross-View &CNNs	&$\mathcal{L}_2$ loss &MS-COCO~\cite{MS-COCO}, SUN397~\cite{SUN360}  &MACE &SL &- &\checkmark\\
				    
				    &RGGNet~\cite{yuan2020rggnet} &RAL &Cross-Sensor &ResNet	&Geodesic distance loss &KITTI\cite{KITTI}  &MSE, MSEE, MRR &SL &TensorFlow &\checkmark\\
				    
				    &CalibRCNN~\cite{shi2020calibrcnn} &IROS &Cross-Sensor &RNNs	&$\mathcal{L}_2$, Epipolar geometry loss &KITTI~\cite{KITTI}  &MAE &SL &TensorFlow &\checkmark\\

				    &SSI-Calib~\cite{zhu2020online} &ICRA &Cross-Sensor &CNNs	&$\mathcal{L}_2$ loss &Pascal VOC 2012~\cite{pascal-voc-2012}  &Mean/standard deviation &SL &TensorFlow &\checkmark\\

				    &SOIC~\cite{wang2020soic} &arXiv &Cross-Sensor &ResNet+PointRCNN	& Cost function &KITTI~\cite{KITTI}  &Mean error &SL &- &\\        

				    &NetCalib~\cite{wu2021netcalib} &ICPR &Cross-Sensor &CNNs	&$\mathcal{L}_1$ loss &KITTI~\cite{KITTI}  &MAE &SL &PyTorch &\checkmark\\
                        				   
					\hline
					\hline
					\multirow{1}{*}{\rotatebox{0}{\textbf{2021}}}
				   
				   &DLKFM~\cite{DLKFM} &CVPR &Cross-View &Siamese-Net &$\mathcal{L}_2$ loss & MS-COCO\cite{MS-COCO}, Google Earth, Google Map &MSE & SL &TensorFlow &\checkmark \\
				   
				   &LocalTrans~\cite{LocalTrans} &ICCV &Cross-View &Transformer &$\mathcal{L}_1$ loss & MS-COCO\cite{MS-COCO} &MSE, PSNR, SSIM & SL &PyTorch &\checkmark \\
				   
				   &BasesHomo~\cite{BasesHomo} &ICCV &Cross-View &ResNet &Triplet loss & CA-UDHN\cite{CA-UDHN} &MSE & USL &PyTorch & \\
				   &ShuffleHomoNet~\cite{ShuffleHomoNet} &ICIP &Cross-View &ShuffleNet &$\mathcal{L}_2$ loss & MS-COCO\cite{MS-COCO} &RMSE & SL &TensorFlow &\checkmark \\
				   
				   &DAMG-Homo~\cite{DAMG-Homo} &TCSVT &Cross-View &CNNs &$\mathcal{L}_1$ loss & MS-COCO\cite{MS-COCO}, UDIS\cite{UDIS} &RMSE, PSNR, SSIM & SL &TensorFlow &\checkmark \\
                   
                   &LCCNet~\cite{lv2021lccnet} &CVPRW &Cross-Sensor &CNNs &Smooth $\mathcal{L}_1$, $\mathcal{L}_2$ loss &KITTI\cite{KITTI} &MSE  & SL &PyTorch &\checkmark \\
                   
                   &CFNet~\cite{lv2021cfnet} &Sensors &Cross-Sensor &FCN &$\mathcal{L}_1$, Charbonnier\cite{Charbonnier} loss &KITTI\cite{KITTI}, KITTI-360\cite{liao2022kitti} &MAE, MSEE, MRR  & SL &PyTorch &\checkmark \\

                   &SemAlign~\cite{liu2021semalign} &IROS &Cross-Sensor &CNNs & Semantic alignment loss &KITTI~\cite{KITTI} &Mean/median rotation errors & SL &PyTorch &\checkmark\\
       
				   \hline
				   \hline
				   \multirow{1}{*}{\rotatebox{0}{\textbf{2022}}}
				   &IHN~\cite{IHN} &CVPR &Cross-View  &Siamese-Net	&$\mathcal{L}_1$ loss &MS-COCO\cite{MS-COCO}, Google Earth, Google Map &MACE & SL &PyTorch &\checkmark\\
				   
				   &HomoGAN~\cite{HomoGAN} &CVPR &Cross-View  &GANs	&Cross-entropy, WGAN loss &CA-UDHN\cite{CA-UDHN} &Mean error & USL &PyTorch &\checkmark\\

				   &Liu et al.~\cite{Liu} &TPAMI &Cross-View &ResNet&Triplet loss &Self-constructed  &MSE, Accuracy &USL &PyTorch &\\
				   
				   &DXQ-Net~\cite{jing2022dxq} &IROS &Cross-Sensor  &CNNs+RNNs&$\mathcal{L}_1$, geodesic loss &KITTI\cite{KITTI}, KITTI-360\cite{liao2022kitti}  &MSE &SL &PyTorch &\checkmark\\
				   
				   &SST-Calib~\cite{SST-Calib} &ITSC &Cross-Sensor  &CNNs &$\mathcal{L}_2$ loss &KITTI\cite{KITTI}  &QAD, AEAD &SL &PyTorch &\checkmark\\

				   &ATOP~\cite{ATOP} &TIV &Cross-Sensor  &CNNs &Cross entropy loss &Self-constructed + KITTI\cite{KITTI}  &RRE, RTE &SL &- &\\

				   &FusionNet~\cite{wang2022fusionnet} &ICRA &Cross-Sensor  &CNNs+PointNet &$\mathcal{L}_2$ loss &KITTI\cite{KITTI}  &MAE &SL &PyTorch &\checkmark\\

				   &RGKCNet~\cite{RGKCNet} &TIM &Cross-Sensor  &CNNs+PointNet &$\mathcal{L}_1$ loss &KITTI\cite{KITTI}  &MSE &SL &PyTorch &\checkmark\\

                      \hline
				   \hline
				   \multirow{1}{*}{\rotatebox{0}{\textbf{2023}}}

					&BinoStereo~\cite{BinoStereo} &TIV &Cross-View &CNNs	&Quaternion distance loss &KITTI~\cite{KITTI} &Mean error, SSIM & SL &- &\\

					&DPO-Net~\cite{DPO-Net} &ICCV &Cross-View &GNNs	&Negative log-likelihood, cosine similarity loss &ScanNet~\cite{ScanNet}, MegaDepth~\cite{MegaDepth} &AUC & SL &PyTorch &\\

					&RHWF~\cite{RHWF} &CVPR &Cross-View &CNNs+Transformers	&$\mathcal{L}_1$ loss &Google Earth, Google Map, MS-COCO~\cite{MS-COCO} &MACE &SL &PyTorch &\checkmark\\                    
					&EAC-Homo~\cite{EAC-Homo} &TCSVT &Cross-View &CNNs	&$\mathcal{L}_1$ loss &UDIS-D~\cite{UDIS}, MS-COCO~\cite{MS-COCO} &MACE, PSNR, SSIM & USL &PyTorch &\\

					&PLS-Homo~\cite{PLS-Homo} &TCSVT &Cross-View &CNNs	&Triple, $\mathcal{L}_1$ loss &CA-UDHN~\cite{CA-UDHN} &Point matching error & USL &PyTorch &\\

					&LBHomo~\cite{LBHomo} &AAAI &Cross-View &CNNs	&$\mathcal{L}_1$ loss &Self-constructed &Point matching error & Semi-SL &PyTorch &\\

					&RealSH~\cite{RealSH} &ICCV &Cross-View &CNNs	&$\mathcal{L}_1$ loss &MS-COCO~\cite{MS-COCO}, CA-UDHN~\cite{CA-UDHN}, GHOF~\cite{Gyroflow+} &Point matching error & SL &PyTorch &\\

					&SE-Calib~\cite{SE-Calib} &TGRS &Cross-Sensor &CNNs	&SSRCM score &KITTI~\cite{KITTI} &Reprojection error, mean error & SL &PyTorch &\checkmark\\

					&CM-GNN~\cite{CM-GNN} &TIM &Cross-Sensor &CNNs+GNNs	&Focal, smooth $\mathcal{L}_1$ loss &KITTI~\cite{KITTI} &Geodesic distance & SL &PyTorch &\checkmark\\

					&Calibdepth~\cite{Calibdepth} &ICRA &Cross-Sensor &CNNs+LSTMs	&Chamfer distance, smooth $\mathcal{L}_1$, berHu, geodesic distance loss &KITTI~\cite{KITTI} &Absolute error & SL &PyTorch &\checkmark\\

					&DEdgeNet~\cite{DEdgeNet} &ICRA &Cross-Sensor &CNNs	&$\mathcal{L}_2$ loss, quaternion distance loss &KITTI~\cite{KITTI} &Mean error & SL &- &\checkmark\\

					&MOISST~\cite{MOISST} &IROS &Cross-Sensor &MLPs	&$\mathcal{L}_2$ loss &KITTI-360~\cite{liao2022kitti} &Mean error & SL &- &\checkmark\\

					&ELR-Calib~\cite{ELR-Calib} &ICASSP &Cross-Sensor &CNNs	&Contrastive loss &RELLIS-3D~\cite{RELLIS-3D} &Averaged translation error and rotation error & SL &- &\checkmark\\

					&P2O-Calib~\cite{P2O-Calib} &IROS &Cross-Sensor &CNNs	&Cross-entropy loss &KITTI~\cite{KITTI} &Averaged translation error and rotation error & SL &Pytorch &\checkmark\\

					&SCNet~\cite{SCNet} &RAL &Cross-Sensor &CNNs+Transformers	&Smooth $\mathcal{L}_1$, $\mathcal{L}_2$ loss &KITTI~\cite{KITTI}, nuScenes~\cite{caesar2020nuscenes} &Averaged translation error and rotation error & SL &Pytorch &\checkmark\\

					&RobustCalib~\cite{RobustCalib} &arXiv &Cross-Sensor &CNNs	&$\mathcal{L}_1$, cross-entropy loss &KITTI~\cite{KITTI}, nuScenes~\cite{caesar2020nuscenes} &Averaged translation error and rotation error & SL &- &\checkmark\\

					&PseudoCal~\cite{PseudoCal} &BMVC &Cross-Sensor &CNNs+Transformers	&$\mathcal{L}_2$ loss &KITTI~\cite{KITTI} &Averaged translation error and rotation error & SL &- &\checkmark\\

					&BatchCalib~\cite{BatchCalib} &CoRL &Cross-Sensor &CNNs	&Reprojection error &KITTI~\cite{KITTI} &Mean, median error  & SL &- &\checkmark\\
                    
                      \hline
				   \hline
				   \multirow{1}{*}{\rotatebox{0}{\textbf{2024}}}
					&NeuralRecalibration~\cite{NeuralRecalibration} &arXiv &Cross-View &Transformers+PointNet	&Geodesic, $\mathcal{L}_2$ loss & Self-constructed &RMSE& SL &PyTorch &\checkmark\\
				  
                  &ArcGeo~\cite{ArcGeo} &WACV &Cross-View &Transformers	&Cross-entropy loss & CVUSA~\cite{CVUSA}, CVACT~\cite{CVACT} &Top-K recall, mAR& SL &- &\\

                  &CalibRBEV~\cite{CalibRBEV} &ACM MM &Cross-View &CNNs+Transformers	&Focal, $\mathcal{L}_1$ loss & nuScenes~\cite{caesar2020nuscenes}, Waymo~\cite{Waymo} &mAP, NDS, mATE, mASE, etc& SL &- &\\

                  &FG-Rect~\cite{FG-Rect} &CVPR &Cross-View &CNNs+Transformers	&$\mathcal{L}_1$ loss & Semi-Truck Highway~\cite{FG-Rect}, Carla~\cite{FG-Rect} &MAE& SL &Pytorch &\\

                  &Mask-Homo~\cite{Mask-Homo} &AAAI &Cross-View &Transformers	&Triplet loss &CA-UDHN~\cite{CA-UDHN} &MSE, PSNR& USL &Pytorch &\\

                  &DSM-DHN~\cite{DSM-DHN} &RAL &Cross-View &CNNs	&$\mathcal{L}_1$ loss &UDIS-D~\cite{UDIS}, MS-COCO~\cite{MS-COCO} &RMSE, PSNR, SSIM& USL &- &\\

                  &DMHomo~\cite{DMHomo} &ToG &Cross-View &Diffusion	&Diffusion loss &CA-UDHN~\cite{CA-UDHN} &PME& USL &Pytorch &\\

                  &DPP-Homo~\cite{DPP-Homo} &ICASSP &Cross-View &CNNs	&$\mathcal{L}_1$ loss &UDIS-D~\cite{UDIS} &PSNR, SSIM& USL &Pytorch &\\

                  &DHE-VPR~\cite{DHE-VPR} &AAAI &Cross-View &Transformer	&Re-projection, triplet loss &Pitts30k~\cite{Pitts30k} &Recall& USL &Pytorch &\\

                  &SRMatcher~\cite{SRMatcher} &ACM MM &Cross-View &CNNs+Transformer	&Focal binary cross-entropy Loss &Oxford5K~\cite{Oxford5K}, Paris6K~\cite{Paris6K} &AUC& SSL &Pytorch &\\

                  &AbHE~\cite{AbHE} &TIM &Cross-View &CNNs+Transformer	&$\mathcal{L}_1$ loss &UDIS-D~\cite{UDIS}, MS-COCO~\cite{MS-COCO} &PSNR, SSIM& USL &TensorFlow &\\

                  &AGNet~\cite{AGNet} &TCSVT &Cross-View &CNNs+Transformer	&$\mathcal{L}_1$ loss &Google Earth, Google Map, MS-COCO~\cite{MS-COCO} &MACE& SL &Pytorch &\checkmark\\

                  &CrossHomo~\cite{CrossHomo} &TPAMI &Cross-View &CNNs	&$\mathcal{L}_1$, $\mathcal{L}_2$ loss &MS-COCO~\cite{MS-COCO}, DPDN~\cite{DPDN} &RMSE, PSNR, SSIM& SL &Pytorch &\checkmark\\

                  &AltO~\cite{AltO} &NeurIPS &Cross-View &CNNs	&Barlow Twins loss &Google Earth, Google Map, DeepNIR~\cite{DeepNIR} &MACE & USL &Pytorch &\checkmark\\

                  &Gyroflow+~\cite{Gyroflow+} &IJCV &Cross-View &CNNs	&Triplet loss  &GOF~\cite{Gyroflow} &MACE & USL &Pytorch &\\

                  &HEN~\cite{HEN} &WACV &Cross-View &CNNs	&$\mathcal{L}_1$ loss  &MS-COCO~\cite{MS-COCO} &MAE & SL &TensorFlow &\checkmark\\

                  &JEDL-Homo~\cite{JEDL-Homo} &ICME &Cross-View &CNNs	&$\mathcal{L}_1$ loss  &MS-COCO~\cite{MS-COCO} &MACE & SL &Pytorch &\checkmark\\

                  &InterNet~\cite{InterNet} &arXiv &Cross-View &CNNs	&$\mathcal{L}_1$, $\mathcal{L}_2$ loss  &Google Map, DPDN~\cite{DPDN}, RGB/NIR~\cite{RGB/NIR} &MACE & SL &Pytorch &\checkmark\\

                  &STHN~\cite{STHN} &RAL &Cross-View &CNNs	&$\mathcal{L}_1$ loss  &Boson-nighttime~\cite{Boson-nighttime} &MACE, CE & SL &Pytorch &\\

                  &SCPNet~\cite{SCPNet} &ECCV &Cross-View &CNNs	&$\mathcal{L}_1$ loss &Google Map, RGB/NIR~\cite{RGB/NIR} &MACE &USL &PyTorch &\checkmark\\ 

                  &MCNet~\cite{MCNet} &CVPR &Cross-View &CNNs	&$\mathcal{L}_1$, FGO loss &Google Earth, Google Map, MS-COCO~\cite{MS-COCO} &MACE &SL &PyTorch &\checkmark\\ 

                  &CodingHomo~\cite{CodingHomo} &TCSVT &Cross-View &CNNs	&Triplet, binary cross-entropy, negative log likelihood loss &CA-UDHN~\cite{CA-UDHN}, GOF~\cite{Gyroflow} &Point matching error &USL &PyTorch &\\ 

                &SOAC~\cite{SOAC} &CVPR &Cross-Sensor &MLPs	& $\mathcal{L}_1$, $\mathcal{L}_2$ loss &KITTI-360~\cite{liao2022kitti}, nuScenes~\cite{caesar2020nuscenes}, PandaSet~\cite{PandaSet} &Mean error & SL &- &\checkmark\\

                &UniCal~\cite{UniCal} &ECCV &Cross-Sensor &MLPs	& $\mathcal{L}_1$, $\mathcal{L}_2$, surface alignment loss &MS-Cal~\cite{UniCal}, PandaSet~\cite{PandaSet} &Re-projection error, Point-to-Plane distance, PSNR, SSIM, LPIPS & SL &- &\\

                &L2C-Calib~\cite{L2C-Calib} &TIM &Cross-Sensor &Transformers	& $\mathcal{L}_1$, $\mathcal{L}_2$ loss &KITTI~\cite{KITTI} &Mean error, median error & SL &- &\checkmark\\

                &CalibFormer~\cite{CalibFormer} &ICRA &Cross-Sensor &Transformers	& Smooth $\mathcal{L}_1$, angular distance loss &KITTI~\cite{KITTI} &Mean error & SL &- &\checkmark\\

                &LCCRAFT~\cite{LCCRAFT} &ICRA &Cross-Sensor &CNNs	&Smooth $\mathcal{L}_1$, $\mathcal{L}_2$ loss &KITTI~\cite{KITTI} &Mean error & SL &PyTorch &\checkmark\\

                &SGCalib~\cite{SGCalib} &ICRA &Cross-Sensor &CNNs	&$\mathcal{L}_2$ loss &KITTI~\cite{KITTI} &Mean error & SL &- &\checkmark\\

                &LCANet~\cite{LCANet} &TITS &Cross-Sensor &CNNs+Transformers	&Smooth$\mathcal{L}_1$ loss &KITTI~\cite{KITTI} &Mean error & SL &PyTorch &\checkmark\\

                &SAM-Calib~\cite{SAM-Calib} &ICRA &Cross-Sensor &Transformers	&Cost function &KITTI~\cite{KITTI}, nuScenes~\cite{caesar2020nuscenes} &Mean error & SL &PyTorch &\checkmark\\

                &HIFMNet~\cite{HIFMNet} &ICRA &Cross-Sensor &CNNs	&Quaternion distance, smooth $\mathcal{L}_1$ loss  &KITTI~\cite{KITTI} &Mean error & SL &- &\checkmark\\

                &SensorX2Vehicle~\cite{SensorX2Vehicle} &RAL &Cross-Sensor &CNNs+Transformers	&Cross-entropy, cosine similarity, $\mathcal{L}_1$ loss  &KITTI~\cite{KITTI}, nuScenes~\cite{caesar2020nuscenes} &Mean error & SL &PyTorch &\checkmark\\

                &Edgecalib~\cite{Edgecalib} &RAL &Cross-Sensor &Transformers	&Projection function  &KITTI~\cite{KITTI} &Mean error & SL &- &\checkmark\\
                   
				\hline
				\end{tabular}
			}
		\end{threeparttable}
	\end{table*}

	\section{Standard Model}
\label{sec:pure}
Generally, for learning-based calibration works, the objectives of the intrinsics calibration contain focal length and optical center, and the objectives of the extrinsic calibration contain the rotation matrix and translation vector.

\subsection{Intrinsics Calibration}
Deepfocal \cite{DeepFocal} is a pioneer work in learning-based camera calibration, it aims to estimate the focal length of any image ``in the wild''. In detail, Deepfocal considered a simple pinhole camera model and regressed the horizontal field of view using a deep convolutional neural network. Given the width $w$ of an image, the relationship between the horizontal field of view $H_\theta$ and focal length $f$ can be described by:
\begin{equation}
H_\theta = 2\arctan(\frac{w}{2f}).
\label{eq-focal-length}
\end{equation}

Due to component wear, temperature fluctuations, or outside disturbances like collisions, the calibrated parameters of a camera are susceptible to change over time. To this end, MisCaliDet \cite{MisCaliDet} proposed to identify if a camera needs to be recalibrated intrinsically. Compared to the conventional intrinsic parameters such as the focal length and image center, MisCaliDet presented a new scalar metric, \textit{i.e.}, the average pixel position difference (APPD) to measure the degree of camera miscalibration, which describes the mean value of the pixel position differences over the entire image.

\subsection{Extrinsics Calibration}
In contrast to intrinsic calibration, extrinsic calibration infers the spatial correspondence of the camera and its located 3D scene. PoseNet\cite{PoseNet} first proposed deep convolutional neural networks to regress 6-DoF camera pose in real-time. A pose vector $\textbf{p}$ was predicted by PoseNet, given by the 3D position $\textbf{x}$ and orientation represented by quaternion $\textbf{q}$ of a camera, namely, $\textbf{p} = [\textbf{x}, \textbf{q}]$. For constructing the training dataset, the labels are automatically calculated from a video of the scenario using a structure from motion method \cite{wu2013towards}.

Inspired by PoseNet\cite{PoseNet}, the following works improved the extrinsic calibration in terms of the intermediate representation, interpretability, data format, learning objective, etc. For example, to optimize the geometric pose objective, DeepFEPE \cite{DeepFEPE} designed an end-to-end keypoint-based framework with learnable modules for detection, feature extraction, matching, and outlier rejection. Such a pipeline imitated the traditional baseline, in which the final performance can be analyzed and improved by the intermediate differentiable module. To bridge the domain gap between the extrinsic objective and image features, recent works proposed to first learn an intermediate representation from the input, such as surface geometry \cite{UprightNet}, depth map \cite{StereoCaliNet}, directional probability distribution \cite{DirectionNet}, camera rays~\cite{CAR}, normal flow \cite{DiffPoseNet}, and optical flow~\cite{DroidCalib}, etc. Then, the extrinsic are reasoned by geometric constraints and learned representation. Therefore, the neural networks are gradually guided to perceive the geometry-related features, which are crucial for extrinsic estimation. Considering the privacy concerns and limited storage problem, some recent works~\cite{Do, SceneSqueezer, Neumap} compressed the scene and exploited the point-like feature to estimate the extrinsic. For example, Do et al. \cite{Do} trained a network to recognize sparse but significant 3D points, dubbed scene landmarks, by encoding their appearance as implicit features. The camera pose can be calculated using a robust minimal solver followed by a Levenberg-Marquardt-based nonlinear refinement. SceneSqueezer \cite{SceneSqueezer} compressed the scene information from three levels: the database frames are clustered using pairwise co-visibility information, a point selection module prunes each cluster based on estimation performance, and learned quantization further compresses the selected points.

Learning-based SLAM (Simultaneous Localization and Mapping) and SfM (Structure from Motion) are also closely related to extrinsic calibration tasks. Both techniques involve pose estimation and tracking to localize cameras in 3D scenes. Direct attempts focused on modifying classical modules, such as feature extraction\cite{detone2018superpoint,sun2021loftr}, feature matching\cite{sarlin2020superglue,lindenberger2023lightglue} and pose estimation\cite{brachmann2017dsac,sarlin2021back}. These modifications have been proven to be efficient and robust but still need to be integrated into conventional pipelines. There is also a series of methods that solve problems in an end-to-end manner which aim to simplify the traditional pipeline by integrating neural networks and differentiable operations. DeepVO\cite{wang2017deepvo} is one of the pioneering supervised solutions that utilized CNNs and RNNs to estimate camera poses. UnDeepVO\cite{li2018undeepvo} further introduced an unsupervised training strategy that combines depth map prediction with pose estimation. Subsequent end-to-end solutions generally adopted similar formulations but primarily focused on incorporating more constraints from depth \cite{bian2019unsupervised,yang2020d3vo,zhu2024revisit}, optical flow\cite{min2020voldor,teed2021droid,wang2021tartanvo}, and semantics \cite{yu2018ds,wu2020eao,chen2022accurate}.

With the advancement of neural rendering, NeRF-based methods have gained significant attention for their promising results. iMAP\cite{sucar2021imap} pioneered the first online SLAM framework using NeRF, enabling joint optimization of camera poses and scene representations. Subsequent studies expanded scene representations to hierarchical voxel grids\cite{zhu2022nice,rosinol2022nerf}, signed distance fields\cite{ortiz2022isdf,johari2023eslam}, point clouds\cite{sandstrom2023point,hu2024cp}, and 3D Gaussians\cite{yan2024gs}. While most NeRF-based approaches depend on traditional odometry for pose initialization, NeRF remains a potential solution for training pose estimation models with limited accurate labels and enhancing pose accuracy. To reduce reliance on precise pose assumptions, researchers have incorporated geometric priors such as depth estimation~\cite{truong2023sparf,wang2023altnerf}, multi-view correspondence~\cite{jeong2021self,bian2023nope,wang2023altnerf}, and GAN-based constraints~\cite{meng2021gnerf} to stabilize optimization.
 
\subsection{Joint Intrinsic and Extrinsic Calibration}

\subsubsection{Geometric Representations}

\noindent \textbf{Vanishing Points}
The intersection of projections of a set of parallel lines in the world leads to a vanishing point. The detection of vanishing points is a fundamental and crucial challenge in 3D vision. In general, vanishing points reveal the direction of 3D lines, allowing the agent to deduce 3D scene information from a single 2D image.

DeepVP \cite{DeepVP} is the first learning-based work for detecting the vanishing points given a single image. It reversed the conventional process by scoring the horizon line candidates according to the vanishing points they contain. Chang et al. \cite{chang2018deepvp} redesigned this task as a classification problem using an output layer with possible vanishing point locations. To directly leverage the geometric properties of vanishing points, NeurVPS \cite{zhou2019neurvps} proposed a canonical conic space and a conic convolution operator that can be implemented as regular convolutions in this space, where the learning model is capable of calculating the global geometric information of vanishing points locally. To overcome the need for a large amount of training data, DVPD \cite{DVPD} incorporated the neural network with two geometric priors: Hough transformation and Gaussian sphere. First, the convolutional features are transformed into a Hough domain, mapping lines to distinct bins. The projection of the Hough bins is then extended to the Gaussian sphere, where lines are transformed into great circles and vanishing points are located at the intersection of these circles. Geometric priors are data-efficient because they eliminate the necessity for learning this information from data, which enables an interpretable learning framework and generalizes better to domains with slightly different data distributions.

\noindent \textbf{Horizon Lines}
The horizon line is a crucial contextual attribute for various computer vision tasks, especially image metrology, computational photography, and 3D scene understanding. The projection of the line at infinity onto any plane that is perpendicular to the local gravity vector determines the location of the horizon line. Given the FoV, pitch, and roll of a camera, it is straightforward to locate the horizon line in its captured image space. DeepHorizon \cite{DeepHorizon} proposed the first learning-based solution for estimating the horizon line from an image. To train the network, a new benchmark dataset, Horizon Lines in the Wild (HLW), was constructed, which consists of real-world images with labeled horizon lines. SA-MobileNet \cite{SA-MobileNet} proposed an image tilt detection and correction with self-attention MobileNet \cite{howard2019searching} for smartphones. A spatial self-attention module was devised to learn long-range dependencies and global context within the input images. To address the regression difficulty, the network is trained to estimate multiple angles within a narrow interval of the ground truth tilt.

\begin{figure}[!t]
  \centering
  \includegraphics[width=.47\textwidth]{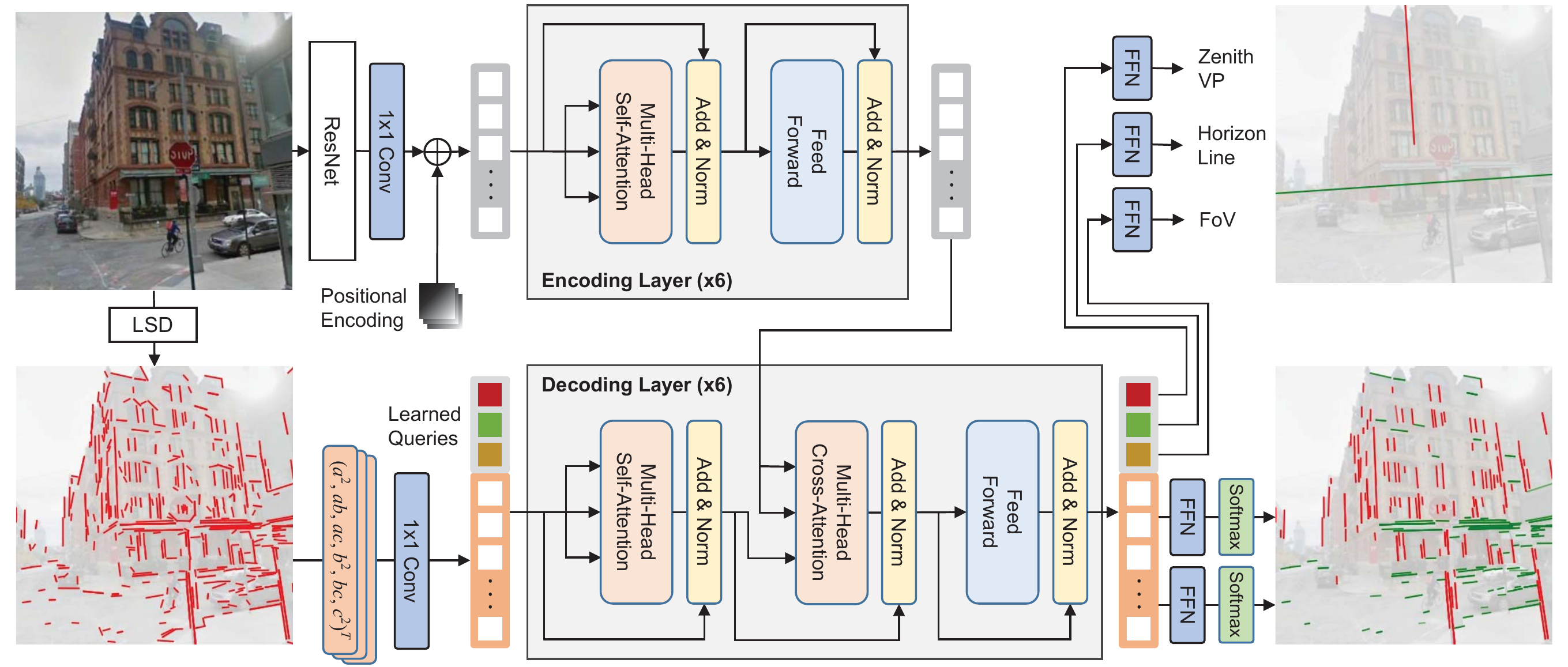}
  \caption{Overview of CTRL-C. The figure is from ~\cite{CTRL-C}. It estimates parameters including the zenith VP, FoV, and horizon line for camera calibration from an input image and a set of line segments. Moreover, two auxiliary outputs (vertical and horizontal convergence line scores) guide the network in learning scene geometry for calibration.}
  \label{fig:CTRL-C}
  \vspace{-0.3cm}
\end{figure}

\noindent \textbf{Geometry Fields}
Recent calibration works tend to design a novel \textit{geometry field} to replace the traditional geometric representations as the new learning target, which is inspired by the prior of camera models or the perspective properties of captured images, such as the distortion distribution map~\cite{DDM, OrdianlDistortion}, incidence field~\cite{WildCamera}, incident map~\cite{DiffCalib}, perspective field~\cite{PerspectiveField, GeoCalib}, camera rays~\cite{CAR}, and camera image~\cite{DM-Calib}, etc. These fields represent a pixel-wise or patch-wise parametrization of the intrinsic and/or extrinsic invariants. They show an explicit relationship to the image details and are easy to learn for neural networks. After predicting the geometry field, the calibrated camera parameters can be easily converted and computed via RANSAC, camera reprojection, or Levenberg-Marquardt optimization, etc.

\subsubsection{Composite Parameters}
Calibrating the composite parameters aims to estimate the intrinsic parameters and extrinsic parameters simultaneously. By jointly estimating composite parameters and training using data from a large-scale panorama dataset \cite{SUN360}, Hold-Geoffroy\etal \cite{Hold-Geoffroy} outperformed previous independent calibration tasks. Moreover, Hold-Geoffroy\etal \cite{Hold-Geoffroy, PM-Calib} performed human perception research in which the participants were asked to evaluate the realism of 3D objects composited with and without accurate calibration. This data was further designed to a new perceptual measure for the calibration errors. In terms of the feature category, some methods~\cite{Lee, CTRL-C, SOFI, MSCC} considered both semantic features and geometric cues for camera calibration. They showed how making use of geometric features, is capable of facilitating the network to comprehend the underlying perspective structure. The pipeline of CTRL-C is illustrated in Figure~\ref{fig:CTRL-C}. In recent literature, more applications are jointly studied with camera calibration, for example, single view metrology \cite{Zhu}, 3D human pose and shape estimation \cite{SPEC}, depth estimation \cite{Baradad, Fang, FlowMap}, object pose estimation \cite{FocalPose}, and image reflection removal \cite{Zheng}, etc. Considering the heterogeneousness and visual implicitness of different camera parameters, CPL \cite{CPL} estimated the parameters using a novel camera projection loss, exploiting the neural network to reconstruct the 3D point cloud. The proposed loss addressed the training imbalance problem by representing different errors of camera parameters using a unified metric.

\subsubsection{Calibration with Reconstruction}

Dense reconstruction tasks often involve complex constraints, enabling joint calibration alongside reconstruction. While earlier methods \cite{sucar2021imap,zhu2022nice,rosinol2022nerf,johari2023eslam,sandstrom2023point} primarily focused on pose estimation and geometry reconstruction under the assumption of known and accurate calibration parameters, recent advances in NeRF have facilitated simultaneous calibration. These approaches jointly optimize intrinsic and extrinsic parameters during NeRF training. Notably, NeRF--~\cite{wang2021nerf} pioneered the simultaneous optimization of camera parameters by introducing a trainable pinhole camera model. This pipeline was further advanced by SCNeRF~\cite{jeong2021self}, which proposed a comprehensive camera model incorporating pinhole design, radial distortion, and pixel-specific noise. Similarly, SiNeRF\cite{xia2022sinerf} introduced a sinusoidal activation function and a Mixed Region Sampling strategy to alleviate systematic sub-optimality in joint optimization. Additionally, CAMP\cite{park2023camp} analyzed the selection of camera parameterization and proposed a preconditioned camera optimization technique.

Popular frameworks like InstantNGP~\cite{muller2022instant} and NeRFStudio~\cite{tancik2023nerfstudio} offer features to fine-tune camera parameters. Typically, NeRF methods use outputs from tools like COLMAP or Polycam, with rendering quality tied closely to initial data quality. Minor errors in intrinsics or poses can lead to noticeable rendering artifacts. Current approaches~\cite{muller2022instant, tancik2023nerfstudio} integrate intrinsic parameters, distortion coefficients, and camera poses into NeRF's training optimization, enhancing both geometric structure learning and parameter optimization. This boosts 3D geometry robustness and refines camera models.

\subsection{Discussion}

\subsubsection{Technique Summary}
The above methods target automatic calibration without manual intervention and scene assumption. Early literature~\cite{DeepFocal, PoseNet} separately studied the intrinsic calibration or extrinsic calibration. Driven by large-scale datasets and powerful networks, subsequent works~\cite{DeepVP, DeepHorizon, Hold-Geoffroy, CTRL-C} considered a comprehensive camera calibration, inferring various parameters and geometric representations. To relieve the difficulty of learning the camera parameters, some works~\cite{UprightNet, StereoCaliNet, DirectionNet, DiffPoseNet, PerspectiveField} proposed to learn an intermediate representation. In recent literature, more applications are jointly studied with camera calibration~\cite{Zhu, SPEC, Baradad, Fang, Zheng}. This suggests solving the downstream vision tasks, especially in 3D tasks may require prior knowledge of the image formation model. Moreover, some geometric priors~\cite{DVPD} can alleviate the data-starved requirement of deep learning, showing the potential to bridge the gap between the calibration target and semantic features.  

It is interesting to find that increasing more extrinsic calibration methods~\cite{DeepFEPE, Do, SceneSqueezer} revisited the traditional feature point-based solutions. The extrinsics that describe the camera motion contain limited degrees of freedom, and thus some local features can well represent the spatial correspondence. Besides, the network designed for point learning improves the efficiency of calibration models, such as PointNet~\cite{qi2017pointnet} and PointCNN~\cite{li2018pointcnn}. Such a pipeline also enables clear interpretability of learning-based camera calibration, which promotes understanding of how the network calibrates and magnifies the influences of intermediate modules.

\subsubsection{Future Effort}

(1) Explore more model priors. Most learning-based methods study the parametric camera models but their generalization abilities are limited. In contrast, non-parametric models directly model the relationship between the 3D imaging ray and its resulting pixel in the image, encoding valuable priors in the learned semantic features to reason the camera parameters. A non-parametric model based on implicit neural representations, as explored in \cite{lin2023learning}, has shown remarkable advantages in accurately capturing the camera within a higher-dimensional space. This approach holds potential for future extensions, such as modeling physical factors like aberration. Recent works~\cite{NeuralLens, WildCamera, PerspectiveField} incorporate the perspective of modeling pixel-wise information for camera calibration, making minimal assumptions on the camera model and showing more interpretable and in line with how humans perceive.

(2) Decouple different stages in an end-to-end calibration learning model. Most learning-based methods include a feature extraction stage and an objective estimation stage. However, how the networks learn the features related to calibration is ambiguous. Therefore, decoupling the learning process by different traditional calibration stages can guide the way of feature extraction. It would be meaningful to extend the idea from the extrinsic calibration~\cite{DeepFEPE, Do, SceneSqueezer} to more general calibration problems.

(3) Transfer the measurement space from the parameter error to the geometric difference. Jointly calibrating multiple camera parameters poses an optimization imbalance due to their varying sample distributions. Simple normalization fails to unify their error spaces. A potential solution is to establish a direct measurement space based on the geometric properties of different camera parameters.

(4) Training NeRF without precise camera parameters remains challenging, particularly in scenarios with sparse views, significant motion, low-texture regions, and suboptimal initial values. While modern NeRF-based methods optimize camera parameters with notable results, they are computationally demanding and lack the generalization of deep-learning calibration techniques. We argue that in current NeRF-based approaches, camera parameters often play a secondary role. Therefore, developing effective calibration algorithms that leverage NeRF remains a difficult yet promising pursuit.
	\section{Distortion Model}
\label{sec:distortion}
In the learning-based camera calibration, calibrating the radial distortion and roll shutter distortion gains increasing attention due to their widely used applications for the wide-angle lens and CMOS sensor. In this part, we mainly review the calibration/rectification of these two distortions.

\subsection{Radial Distortion}
The literature on learning-based radial distortion calibration can be classified into two main categories: regression-based solutions and reconstruction-based solutions.

\begin{figure}[!t]
  \centering
  \includegraphics[width=.45\textwidth]{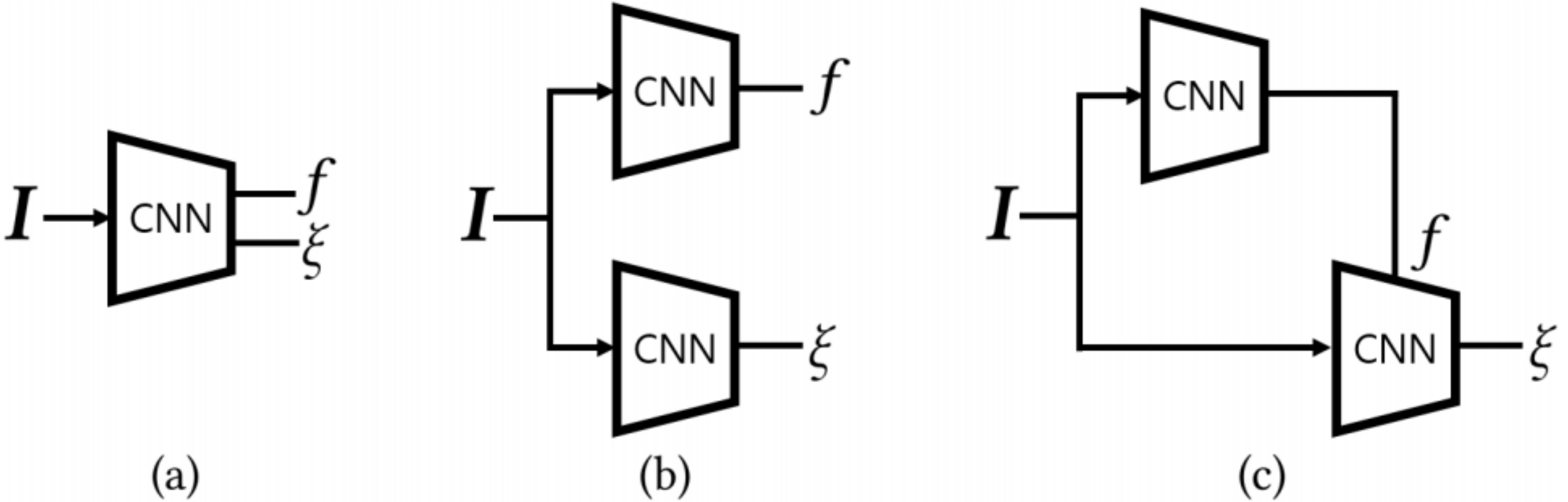}
  \caption{Three common learning solutions of the regression-based wide-angle camera calibration: (a) SingleNet, (b) DualNet, (c) SeqNet, where $\mathbf{I}$ is the distortion image and $f$ and $\xi$ denote the focal length and distortion parameters, respectively. The figure is from ~\cite{DeepCalib}.}
  \label{fig:DeepCalib}
  \vspace{-0.3cm}
\end{figure}

\subsubsection{Regression-based Solution}
Rong \etal~\cite{Rong} and DeepCalib~\cite{DeepCalib} are pioneer works for the learning-based wide-angle camera calibration. They treated the camera calibration as a supervised classification~\cite{Rong} or regression~\cite{DeepCalib} problem, and then the networks with the convolutional layers and fully connected layers were used to learn the distortion features of inputs and predict the camera parameters. In particular, DeepCalib~\cite{DeepCalib} explored three learning solutions for wide-angle camera calibration as illustrated in Figure~\ref{fig:DeepCalib}. Their experiments showed the simplest architecture SingleNet achieves the best performance on both accuracy and efficiency. To enhance the distortion perception of networks, the following works investigated introducing more diverse features such as the semantic features~\cite{FishEyeRecNet} and geometry features~\cite{Xue, LaRecNet, RDCFace}. Additionally, some works improved the generalization by designing learning strategies such as unsupervised learning~\cite{UnFishCor}, self-supervised learning~\cite{SIR}, and reinforcement learning~\cite{Zhao}. By randomly chosen coefficients throughout each mini-batch of the training process, RDC-Net~\cite{RDC-Net} was able to dynamically generate distortion images on-the-fly. It enhanced the rectification performance and prevents the learning model from overfitting. Instead of contributing to the techniques of deep learning, other works leaned to explore the vision prior to interpretable calibration. For example, having observed the radial distortion image owns the center symmetry characteristics, in which the texture far from the image center has stronger distortion, Shi\etal~\cite{Shi} and PSE-GAN~\cite{PSE-GAN} developed a position-aware weight layer (fixed~\cite{Shi} and learnable~\cite{PSE-GAN}) of this property and enabled the network to explicitly perceive the distortion. Lopez\etal~\cite{Lopez} proposed a novel parameterization for radial distortion that is better suited for networks than directly learning the distortion parameters. Furthermore, OrdinalDistortion~\cite{OrdianlDistortion} presented a learning-friendly representation, \textit{i.e.}, ordinal distortion. Compared to the implicit and heterogeneous camera parameters, such a representation can facilitate the distortion perception of the neural network due to its clear relation to the image features.

\subsubsection{Reconstruction-based Solution}
\begin{figure}[!t]
  \centering
  \includegraphics[width=.47\textwidth]{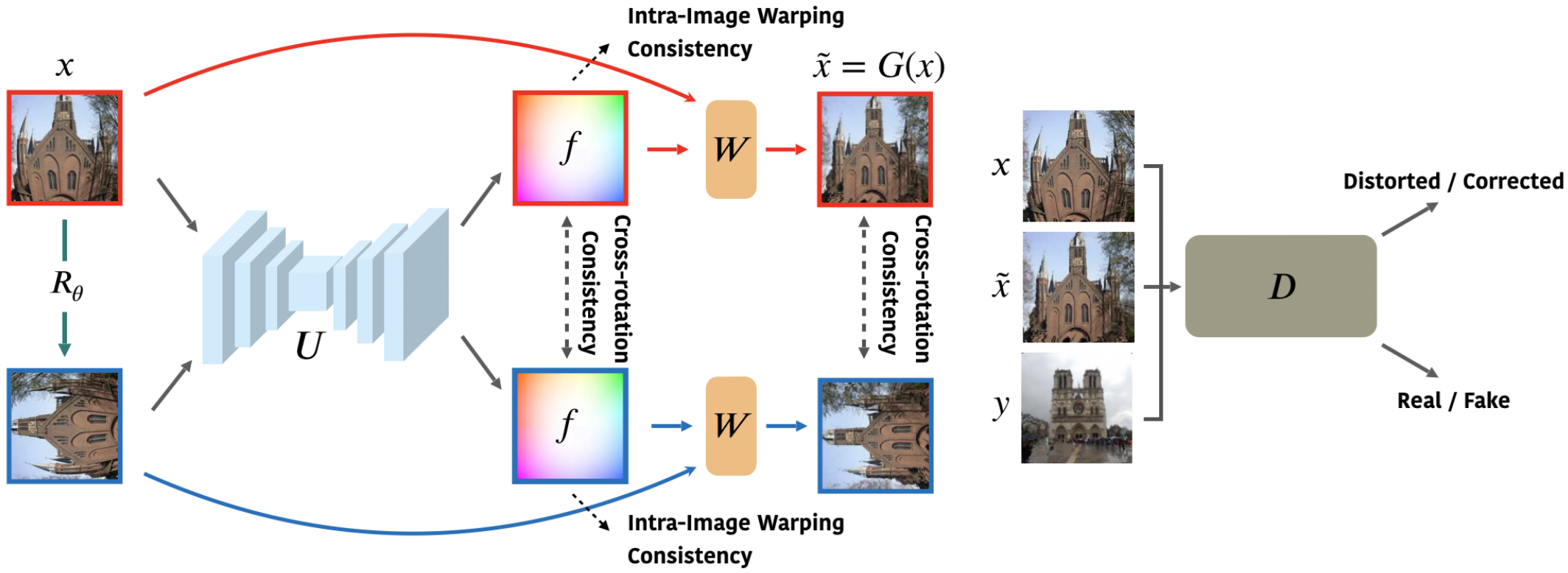}
  \caption{Architecture of FE-GAN. The figure is from ~\cite{FE-GAN}. It consists of two components: a generator $G = (U, W)$ that rectifies the distortion image $x$, and a discriminator $D = (D_{adv}, D_{cls})$. The module $U$ in $G$ predicts the distortion flow $f = U(x)$, while $W$ rectifies the distortion image using $f$.}
  \label{fig:FE-GAN}
  \vspace{-0.3cm}
\end{figure}

Inspired by the image-to-image translation and dense visual perception, the reconstruction-based solution starts to evolve from the conventional regression-based paradigm. DR-GAN~\cite{DR-GAN} is the first reconstruction-based solution for calibrating the radial distortion, which directly models the pixel-wise mapping between the distorted image and the rectified image. It achieved the camera parameter-free training and one-stage rectification. Thanks to the liberation of the assumption of camera models, the reconstruction-based solution showed the potential to calibrate various types of cameras in one learning network. For example, DDM~\cite{DDM} unified different camera models into a domain by presenting the distortion distribution map, which explicitly describes the distortion level of each pixel in an image. Then, the network learned to reconstruct the rectified image using this geometric prior map. To make the mapping function interpretable, the subsequent works \cite{STD, BlindCor, Zhao, FE-GAN, PCN, Tan, SS-WPC, PolarRecNet, SimFIR, MOWA} developed the displacement field between the distorted image and rectified image. Such a manner eliminates the generated artifacts in the pixel-level reconstruction. In particular, FE-GAN~\cite{FE-GAN} integrated the geometry prior like Shi\etal~\cite{Shi} and PSE-GAN~\cite{PSE-GAN} into their reconstruction-based solution and presented a self-supervised strategy to learn the distortion flow for wide-angle camera calibration in Figure~\ref{fig:FE-GAN}. Most reconstruction-based solutions exploit a U-Net-like architecture to learn pixel-level mapping. However, the distortion feature can be transferred from encoder to decoder by the skip-connection operation, leading to a blurring appearance and incomplete correction in reconstruction results. To address this issue, Li\etal~\cite{Li} abandoned the skip-connection in their rectification network. To keep the feature fusion and restrain the geometric difference simultaneously, PCN~\cite{PCN} designed a correction layer in skip-connection and applied the appearance flows to revise the features in different encoder layers. Having noticed that the previous sampling strategy of the convolution kernel neglected the radial symmetry of distortion, some works~\cite{PolarRecNet, DarSwin-Unet} transformed the image from the Cartesian coordinates domain into the polar coordinates domain.

\subsection{Roll Shutter Distortion}
The existing deep learning calibration works on roll shutter (RS) distortion can be classified into two categories: single-frame-based~\cite{URS-CNN, RSC-Net, EvUnroll} and multi-frame-based~\cite{DeepUnrollNet, JCD, SUNet, fan2021inverting, AW-RSC}. The single-frame-based solution studies the case of a single roll shutter image as input and directly learns to correct the distortion using neural networks. The ideal corrected result can be regarded as the global shutter (GS) image. It is an ill-posed problem and requires some additional prior assumptions to be defined. On the contrary, the multi-frame-based solution considers the consecutive frames (two or more) of a video taken by a roll shutter camera, in which the strong temporal correlation can be investigated for more reasonable correction. 

\subsubsection{Single-frame-based Solution}
\begin{figure}[!t]
  \centering
  \includegraphics[width=.47\textwidth]{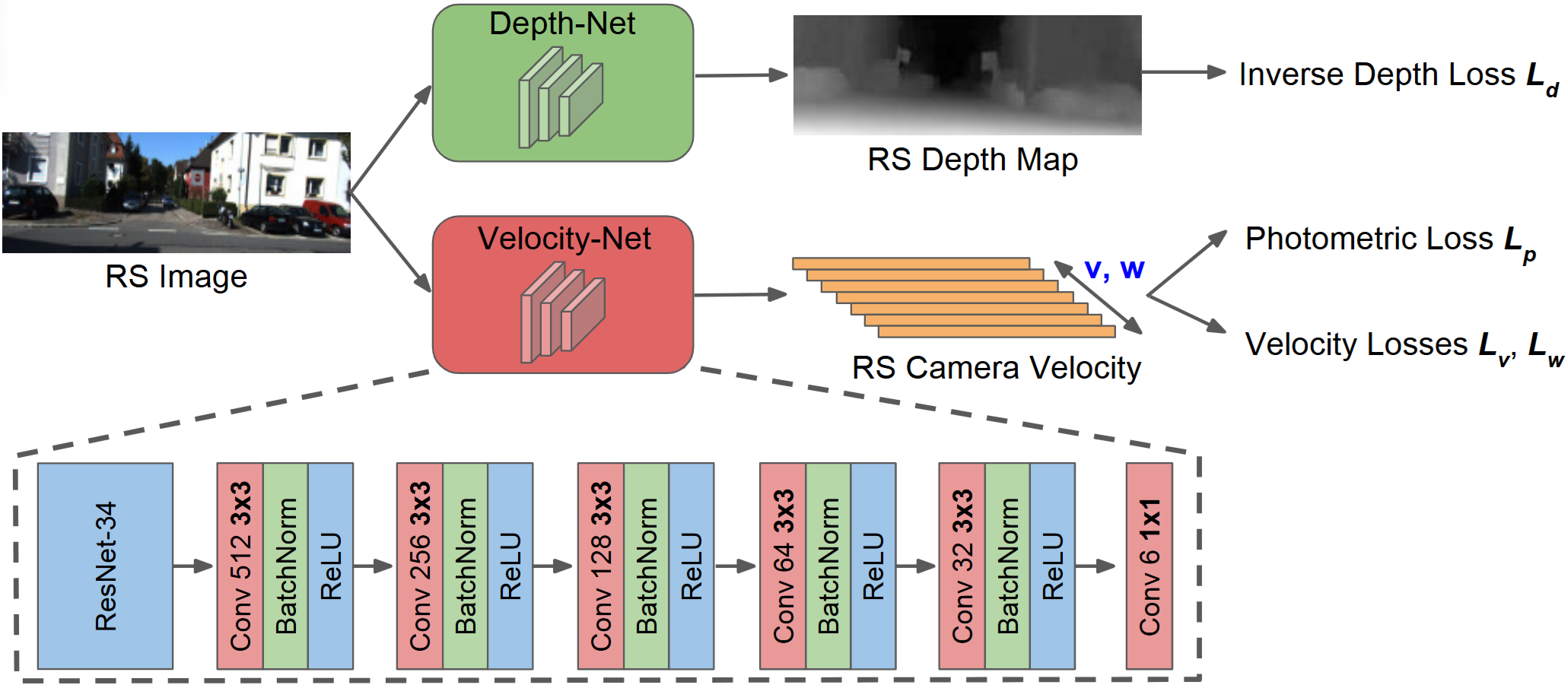}
  \caption{Architecture of RSC-Net. The figure is from ~\cite{RSC-Net}. It consists of two sub-networks, namely DepthNet and Velocity-Net, for learning an RS depth map and RS camera motion from an input image, respectively. Among them, a 6-DOF camera velocity is regressed, including a 3D translational velocity vector $v$ and 3D angular velocity vector $w$.}
  \label{fig:RSC-Net}
  \vspace{-0.06cm}
\end{figure}

URS-CNN~\cite{URS-CNN} is the first learning work for calibrating the rolling shutter camera. In this work, a neural network with long kernel characteristics was used to understand how the scene structure and row-wise camera motion interact. To specifically address the nature of the RS effect produced by the row-wise exposure, the row-kernel and column-kernel convolutions were leveraged to extract attributes along horizontal and vertical axes. RSC-Net~\cite{RSC-Net} improved URS-CNN~\cite{URS-CNN} from 2 degrees of freedom (DoF) to 6-DoF and presents a structure-and-motion-aware model, where the camera scanline velocity and depth were estimated. Compared to URS-CNN~\cite{URS-CNN}, RSC-Net~\cite{RSC-Net} reasoned about the concealed motion between the scanlines as well as the scene structure as shown in Figure~\ref{fig:RSC-Net}. To bridge the spatiotemporal connection between RS and GS, some methods~\cite{EvUnroll, TACA-Net, REG-Net, UniINR} exploited the neuromorphic events to correct the RS effect. Event cameras can overcome many drawbacks of conventional frame-based activities for dynamic situations with quick motion due to their high temporal resolution property with microsecond-level sensitivity.

\begin{figure}[!t]
  \centering
  \includegraphics[width=.47\textwidth]{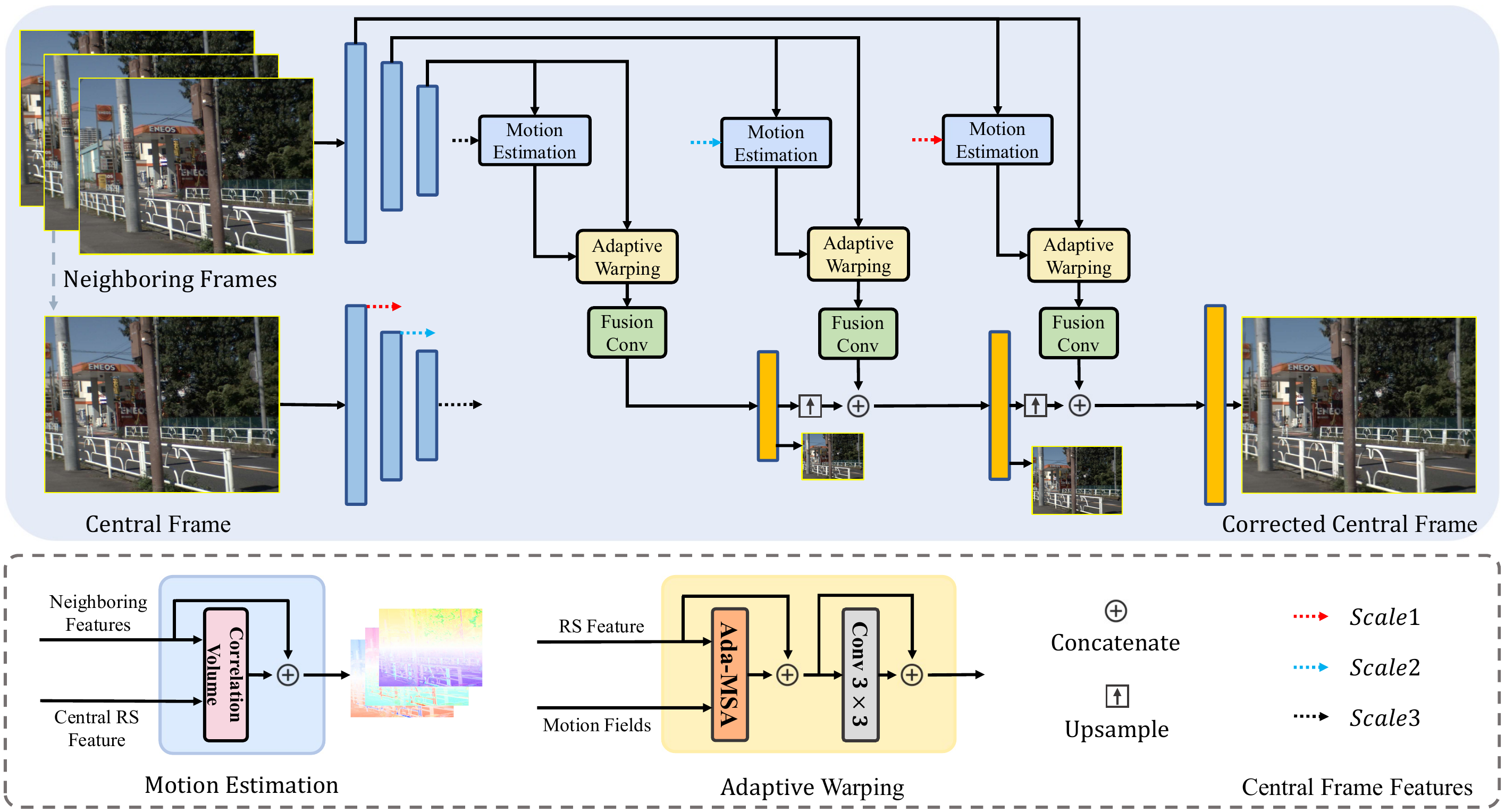}
  \caption{Architecture of AW-RSC. The figure is from ~\cite{AW-RSC}. To address current imprecise motion estimation, it attempts to predict multiple displacement fields instead of only one. Additionally, AW-RSC suggests an adaptive warping module that uses the bundle of fields to guide the adaptive warping of the RS features into the GS one.}
  \label{fig:AW-RSC}
  \vspace{-0.3cm}
\end{figure}

\subsubsection{Multi-frame-based Solution}
Most multi-frame-based solutions are based on the reconstruction paradigm, they mainly devote to contributing how to represent the dense displacement field between RS and global GS images and accurately warp the RS domain to the GS domain. For the first time, DeepUnrollNet~\cite{DeepUnrollNet} proposed an end-to-end network for two consecutive rolling shutter images using a differentiable forward warping module. In this method, a motion estimation network is used to estimate the dense displacement field from a rolling shutter image to its matching global shutter image. The second contribution of DeepUnrollNet~\cite{DeepUnrollNet} is to construct two novel datasets: the Fastec-RS dataset and the Carla-RS dataset. Furthermore, JCD~\cite{JCD} jointly considered the rolling shutter correction and deblurring (RSCD) techniques, which largely exist in the medium and long exposure cases of rolling shutter cameras. It applied bi-directional warping streams to compensate for the displacement while keeping the non-warped deblurring stream to restore details. The authors also contributed a real-world dataset using a well-designed beam-splitter acquisition system, BS-RSCD, which includes both ego-motion and object motion in dynamic scenes. SUNet~\cite{SUNet} extended DeepUnrollNet~\cite{DeepUnrollNet} from the middle time of the second frame ($\frac{3\tau}{2}$) into the intermediate time of two frames ($\tau$). By using PWC-Net~\cite{Sun_2018_CVPR}, SUNet~\cite{SUNet} estimated the symmetric undistortion fields and reconstructed the potential GS frames by a time-centered GS image decoder network. To effectively reduce the misalignment between the contexts warped from two consecutive RS images, the context-aware undistortion flow estimator and the symmetric consistency enforcement were designed. To achieve a higher frame rate, Fan\etal~\cite{fan2021inverting} generated a GS video from two consecutive RS images based on the scanline-dependent nature of the RS camera. In particular, they first analyzed the inherent connection between bidirectional RS undistortion flow and optical flow, demonstrating the RS undistortion flow map has a more pronounced scanline dependency than the isotropically smooth optical flow map. Then, they developed the bidirectional undistortion flows to describe the pixel-wise RS-aware displacement, and further devised a computation technique for the mutual conversion between different RS undistortion flows corresponding to various scanlines. To eliminate the inaccurate displacement field estimation and error-prone warping problems in previous methods, AW-RSC ~\cite{AW-RSC} proposed to predict multiple fields and adaptively warped the learned RS features into global shutter counterparts. Using a coarse-to-fine approach, these warped features were combined and generated to precise global shutter frames as shown in Figure~\ref{fig:AW-RSC}. Compared to previous works~\cite{DeepUnrollNet, JCD, SUNet, fan2021inverting}, the warping operation consisting of adaptive multi-head attention and a convolutional block in AW-RSC~\cite{AW-RSC} is learnable and effective. The following works further incorporate more tractable designs for accurate RS corrections, such as directly estimating the Distortion Flow from consecutive RS frames~\cite{DFRSC}, jointly performing the GS appearance refinement and correction motion estimation~\cite{JAMNet}, jointly learning rolling shutter correction and super-resolution~\cite{PatchNet}, and introducing a self-supervised learning framework with bidirectional distortion warping modules~\cite{SelfDRSC}, etc.

\subsection{Discussion}

\subsubsection{Technique Summary}
The deep learning works on wide-angle camera and roll shutter calibration share a similar technique pipeline. Along this research trend, most early literature begins with the regression-based solution~\cite{Rong, DeepCalib, URS-CNN}. The subsequent works innovated the traditional calibration with a reconstruction perspective~\cite{DR-GAN, DDM, FE-GAN, DeepUnrollNet}, which directly learns the displacement field to rectify the uncalibrated input. For higher accuracy of calibration, a more intuitive displacement field, and more effective warping strategy have been developed~\cite{PCN, AW-RSC, JCD, fan2021inverting, Deep_HM}. To fit the distribution of different distortions, some works designed different shapes of the convolutional kernel~\cite{URS-CNN}, transformed the convolved coordinates~\cite{PolarRecNet}, and equipped with the distortion-aware modules~\cite{FishFormer, Darswin, DarSwin-Unet}.

Existing works devoted themselves to designing more powerful networks and introducing more diverse features to facilitate calibration performance. Recent methods focused on the geometry priors of the distortion~\cite{FE-GAN, PSE-GAN, Shi, VACR, DualPriorsCorrection, NeuralLens}. These priors can be directly weighted into the convolutional layers or used to supervise network training, promoting the learning model to converge faster.

\subsubsection{Future Effort}

(1) The development of wide-angle camera calibration and roll shutter camera calibration could promote each other. For instance, the well-studied multi-frame-based solution in roll shutter calibration is able to inspire wide-angle calibration. The same object located at different sequences could provide useful priors regarding to radial distortion. Additionally, the elaborate studies of the displacement field and warping layer~\cite{AW-RSC, JCD, fan2021inverting} have the potential to motivate the development of wide-angle camera calibration and other fields. Furthermore, the investigation of geometric priors in wide-angle calibration could also improve the interpretability of the network in roll shutter calibration.

(2) Most methods synthesize their training dataset based on random samples from all camera parameters. However, for the images captured by real lenses, the distribution of camera parameters is probably located at a potential manifold \cite{Lopez}. Learning on a label-redundant calibration dataset makes the training process inefficient. Thus, exploring a practical sampling strategy for the synthesized dataset could be a meaningful task in the future direction.

(3) To overcome the ill-posed problem of single-frame calibration, introducing other high-precision sensors can compensate for the current calibration performance, such as event cameras~\cite {EvUnroll, EvShutter, SelfUnroll, REG-Net, UniINR}. With the rapid development of vision sensors, joint calibration using multiple sensors is highly valuable. Consequently, more cross-modal and multi-modal fusion techniques will be investigated along with this research path.

	\section{Cross-View Model}
\label{sec:projection}

Existing methods can estimate specific camera parameters for a single camera. However, in multi-camera scenarios, parameter representations can be more complex. For example, in the multi-view model, the fundamental matrix and essential matrix describe the epipolar geometry and they are intricately tangled with intrinsics and extrinsics~\cite{DPO-Net, CalibRBEV, FG-Rect}. The homography depicts the pixel-level correspondences between different views. In addition to intrinsics and extrinsics, it is also intertwined with depth. Among these complex parameter representations, homography is the most widely leveraged and its related learning-based methods are the most investigated. To this end, we mainly focus on the review of homography estimation solutions for the cross-view model. They can be divided into three categories: direct, cascaded, and iterative solutions.

\begin{figure}[!t]
  \centering
  \includegraphics[width=.49\textwidth]{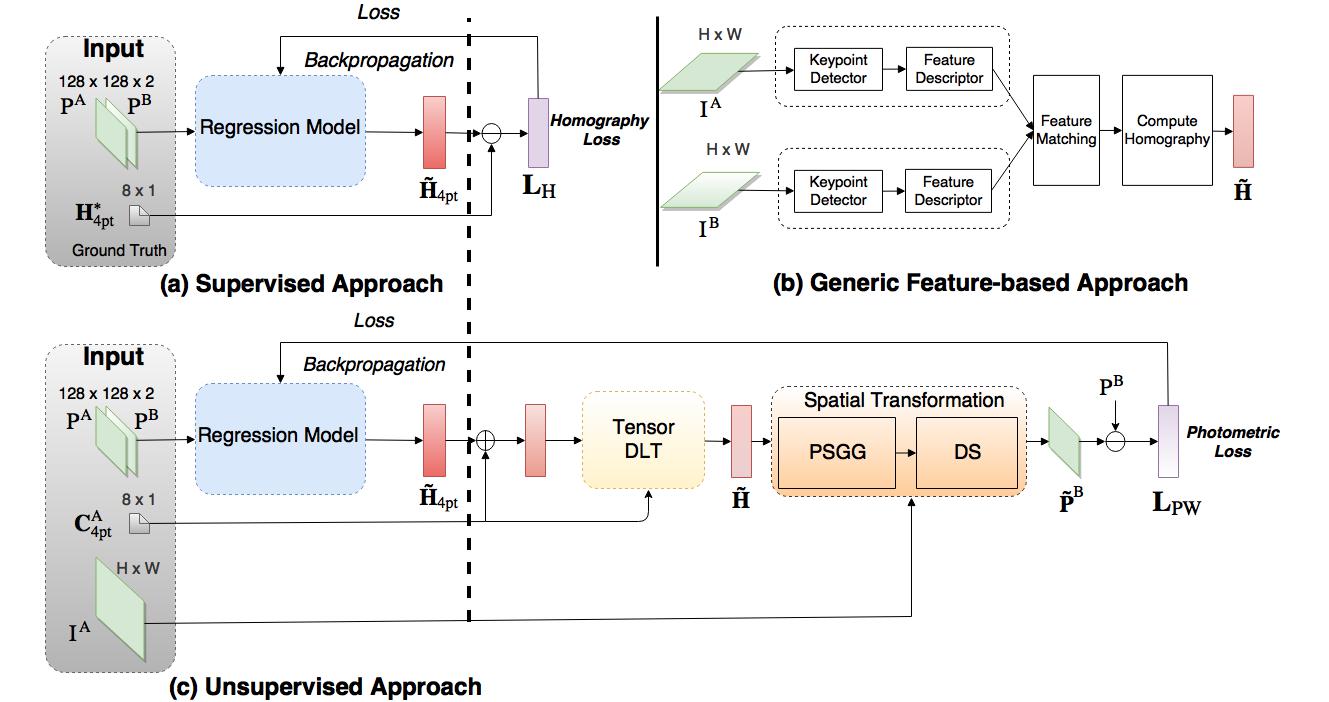}
  \caption{Architectures of DHN~\cite{DHN} and UDHN~\cite{UDHN}. The figure is from ~\cite{UDHN}. The supervised approach~\cite{DHN} learns to regress a 4 point parameterization of homography $\mathbf{\tilde{H}}_{4pt}$ using $\mathcal{L}_2$ loss. The unsupervised approach~\cite{UDHN}  outputs $\mathbf{\tilde{H}}_{4pt}$ that minimizes the $\mathcal{L}_1$ pixel-wise photometric loss of paired inputs (DLT: direct linear transform; PSGG: parameterized sampling grid generator; DS: differentiable sampling).}
  \label{fig:UDHN}
  \vspace{-0.3cm}
\end{figure}

\subsection{Direct Solution}
\label{subsec:Direct}
We review the direct deep homography solutions from the perspective of different parameterizations, including the classical 4-pt parameterization and other parameterizations.
\subsubsection{4-pt Parameterization}
\label{subsec:4pt_parameterization}

Deep homography estimation is first proposed in DHN\cite{DHN}, where a VGG-style network is adopted to predict the 4-pt parameterization $H_{4pt}$. To train and evaluate the network, a synthetic dataset named Warped MS-COCO is created to provide ground truth 4-pt parameterization $\hat{H}_{4pt}$. The pipeline is illustrated in Fig. \ref{fig:UDHN}(a), and the objective function is formulated as $L_{H}$:
\begin{equation}
    L_{H} = \frac{1}{2}\parallel H_{4pt}-\hat{H}_{4pt}\parallel_2^2.
    \label{super_homo}
 \end{equation}
The 4-pt parameterization can be solved as a $3\times 3$ homography matrix using normalized DLT\cite{hartley2003multiple}. However, DHN is limited to synthetic datasets where the ground truth can be generated for free or requires costly labeling of real-world datasets. Subsequently, the unsupervised solution named UDHN\cite{UDHN} is proposed to address this problem. As shown in Fig. \ref{fig:UDHN}(c), it used the same network architecture as DHN and defined an unsupervised loss function by minimizing the average photometric error motivated by traditional methods\cite{lucas1981iterative}:

 \begin{equation}
    L_{PW} = \parallel\mathcal{P}(I_A(x))-\mathcal{P}(I_B(\mathcal{W}(x;p)))\parallel_1,
    \label{unsuper_homo}
 \end{equation}
where $\mathcal{W}(\cdot;\cdot)$ and $\mathcal{P}(\cdot)$ denote the operations of warping via homography parameters $p$ and extracting an image patch, respectively. $I_A$ and $I_B$ are the original images with overlapping regions.
The input of UDHN is a pair of image patches, but it warps the original images when calculating the loss. In this manner, it avoids the adverse effects of invalid pixels after warping and lifts the magnitude of pixel supervision. To gain accuracy and speed with a tiny model, Chen et al. proposed ShuffleHomoNet~\cite{ShuffleHomoNet}, which integrates ShuffleNet compressed units\cite{ma2018shufflenet} and location-aware pooling\cite{Poursaeed} into a lightweight model. To further handle large displacement, a multi-scale weight-sharing version is exploited by extracting multi-scale feature representations and adaptively fusing multi-scale predictions. However, the homography cannot perfectly align images with parallax caused by non-planar structures with non-overlapping camera centers. To deal with parallax, CA-UDHN\cite{CA-UDHN} designs learnable attention masks to overlook the parallax regions, contributing to better background plane alignment. Besides, the 4-pt homography can be extended to meshflow\cite{Liu} to realize non-planar accurate alignment. 

\subsubsection{Other Parameterizations}
\label{subsec:other_parameterization}
In addition to 4-pt parameterization, the homography can be parameterized as other formulations. To better utilize homography invertibility, Wang et al. proposed SSR-Net~\cite{SSR-Net}. They established the invertibility constraint through a conventional matrix representation in a cyclic manner.
Zeng et al. \cite{PFNet} argued that the 4-point parameterization regressed by a fully-connected layer can harm the spatial order of the corners and be susceptible to perturbations, since four points are the minimum requirement to solve the homography. To address these issues, they formulated the parameterization as a perspective field (PF) that models pixel-to-pixel bijection and designed a PFNet. This extends the displacements of the four vertices to as many dense pixel points as possible. The homography can then be solved using RANSAC \cite{fischler1981random} with outlier filtering, enabling robust estimation by utilizing dense correspondences. Nevertheless, dense correspondences lead to a significant increase in the computational complexity of RANSAC. Furthermore, Ye et al.\cite{BasesHomo} proposed an 8-DoF flow representation without extra post-processing, which has a size of $H\times W \times 2$ in an 8D subspace constrained by the homography. To represent arbitrary homography flows in this subspace, 8 flow bases are defined, and the proposed BasesHomo is to predict the coefficients for the flow bases. To obtain desirable bases, BasesHomo first generates 8 homography flows by modifying every single entry of an identity homography matrix except for the last entry. Then, these flows are normalized by their largest flow magnitude followed by a QR decomposition, enforcing all the bases normalized and orthogonal.

\begin{figure}[!t]
  \centering
  \includegraphics[width=.48\textwidth]{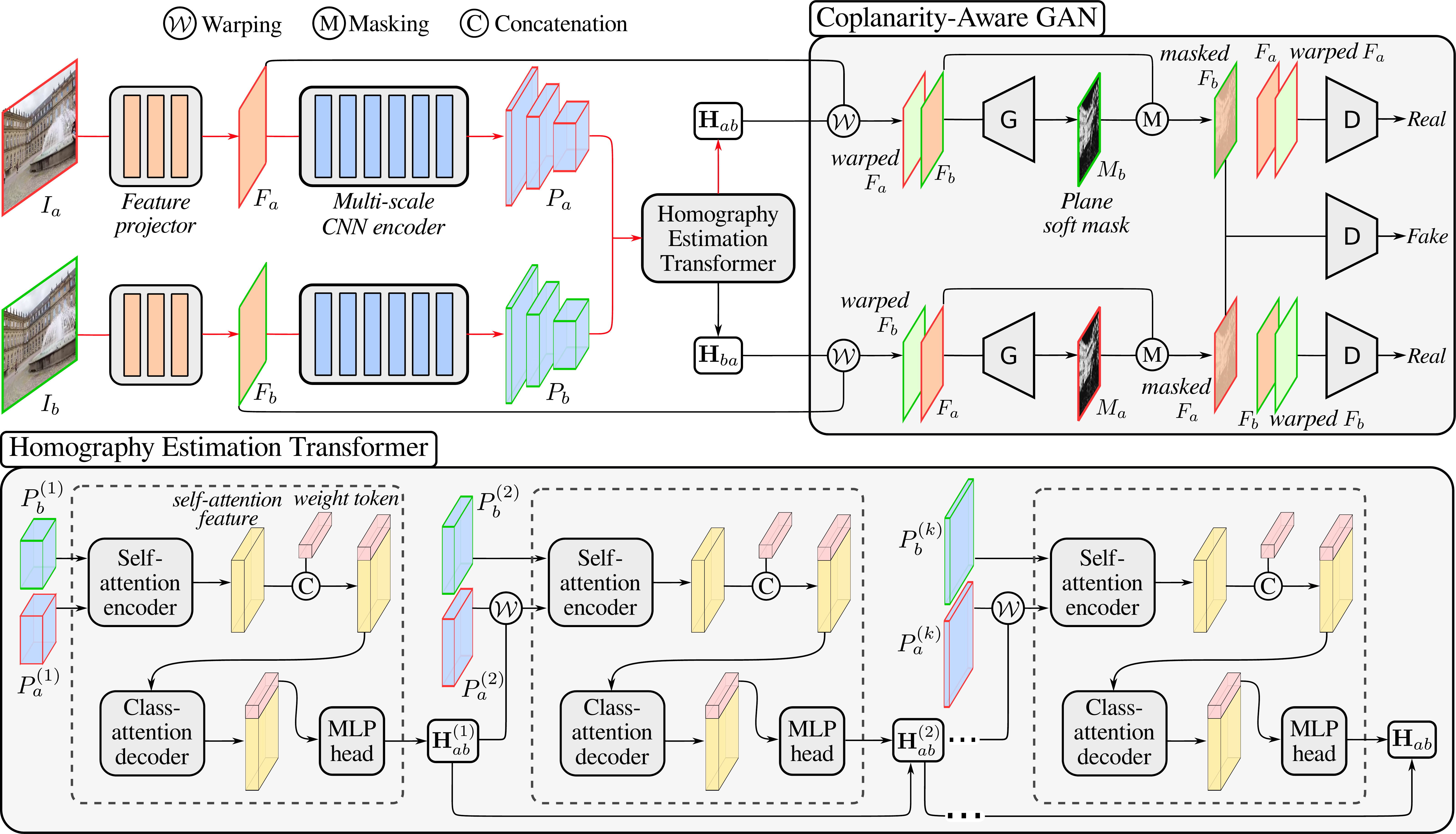}
  \caption{Architecture of HomoGAN. The figure is from ~\cite{HomoGAN}. In particular, the homography estimation transformer with cascaded encoder-decoder blocks takes a feature pyramid of each image as inputs, and predicts the homography from coarse to fine. Coplanarity-aware GAN imposes coplanarity constraints on the model by predicting soft masks of the dominant plane.}
  \label{fig:HomoGAN}
  \vspace{-0.3cm}
\end{figure}

\subsection{Cascaded Solution}
\label{subsec:Cascaded}
Direct solutions explore various homography parameterizations with simple network structures, while the cascaded ones focus on complex designs of network architectures. HierarchicalNet\cite{HierarchicalNet} held that the warped images can be regarded as the input of another network. Therefore it stacked the networks sequentially to reduce the error bounds of the estimate. Based on HierarchicalNet, SRHEN~\cite{SRHEN} introduced the cost volume\cite{Sun_2018_CVPR} to the cascaded network, measuring the feature correlation by cosine distance and formulating it as a volume. The stacked networks and cost volume increase the performance, but they cannot handle the dynamic scenes. MHN~\cite{MHN} developed a multi-scale neural network and proposed to learn homography estimation and dynamic content detection simultaneously. Moreover, to tackle the cross-resolution problem, LocalTrans \cite{LocalTrans} formulated it as a multimodal problem and proposed a local transformer network embedded within a multiscale structure to explicitly learn correspondences between the multimodal inputs. These inputs include images with different resolutions, and LocalTrans achieved superior performance on cross-resolution cases with a resolution gap of up to 10$\times$. All the solutions mentioned above leverage image pyramids to progressively enhance the ability to address large displacements. However, every image pair at each level requires a unique feature extraction network, resulting in the redundancy of feature maps. To alleviate this problem, some researchers\cite{nie2022learning, UDIS, DAMG-Homo, HomoGAN} replaced image pyramids with feature pyramids. Specifically, they warped the feature maps directly instead of images to avoid excessive feature extraction networks. To address the low-overlap homography estimation problem in real-world images\cite{UDIS}, Nie et al.\cite{UDIS} modified the unsupervised constraint (Eq. \ref{unsuper_homo}) to adapt to low-overlap scenes:
\begin{equation}
    L'_{PW} = \parallel I_A(x)\cdot\mathbbm{1}(\mathcal{W}(x;p))-I_B(\mathcal{W}(x;p))\parallel_1,
 \end{equation}
where $\mathbbm{1}$ is an all-one matrix with the same size as $I_A$ or $I_B$. It solved the low-overlap problem by taking the original images as network input and ablating the corresponding pixels of $I_A$ to the invalid pixels of warped $I_B$. 
To solve the non-planar homography estimation problem, DAMG-Homo\cite{DAMG-Homo} proposed backward multi-gird deformation with contextual correlation to align parallax images. Compared with traditional cost volume, the proposed contextual correlation helped to reach better accuracy with lower computational complexity. Another way to address the non-planar problem is to focus on the dominant plane. In HomoGAN \cite{HomoGAN}, an unsupervised GAN is proposed to impose a coplanarity constraint on the predicted homography, as shown in Figure \ref{fig:HomoGAN}. Specifically, a generator is used to predict masks of aligned regions, while a discriminator is used to determine whether two masked feature maps were produced by a single homography. To eliminate the lack of adequate labeled training data, recent methods explored generating realistic and accurate labels using diffusion models~\cite{DMHomo} and dominant plane detection network~\cite{RealSH}, etc.

\subsection{Iterative Solution}
\label{subsec:Iterative}
Compared with cascaded methods, iterative solutions achieve higher accuracy by iteratively optimizing the last estimation. Lucas-Kanade (LK) algorithm\cite{lucas1981iterative} is usually used in image registration to estimate the parameterized warps iteratively, such as affine transformation, optical flow, etc. It aims at the incremental update of warp parameters $\varDelta p$ every iteration by minimizing the sum of squared error between a template image $T$ and an input image $I$:
\begin{equation}
    E(\varDelta p) = \parallel T(x)-I(\mathcal{W}(x;p+\varDelta p))\parallel_2^2.
    \label{lk}
 \end{equation}
However, when optimizing Eq. \ref{lk} using first-order Taylor expansion, $\partial I(\mathcal{W}(x;p))/\partial p$ should be recomputed every iteration because $I(\mathcal{W}(x;p))$ varies with $p$. To avoid this problem, the inverse compositional (IC) LK algorithm\cite{baker2004lucas}, an equivalence to LK algorithm, can be used to reformulate the optimization goal as follows:
\begin{equation}
    E'(\varDelta p) = \parallel T(\mathcal{W}(x;\varDelta p))-I(\mathcal{W}(x;p))\parallel_2^2.
    \label{ic-lk}
 \end{equation}
After linearizing Eq. \ref{ic-lk} with first-order Taylor expansion, we compute $\partial T(\mathcal{W}(x;0))/\partial p$ instead of $\partial I(\mathcal{W}(x;p))/\partial p$, which would not vary every iteration.

To combine the advantages of deep learning with IC-LK iterator, CLKN~\cite{CLKN} conducted LK iterative optimization on semantic feature maps extracted by CNNs as follows:
 \begin{equation}
    E^{f}(\varDelta p) = \parallel F_T(\mathcal{W}(x;\varDelta p))-F_I(\mathcal{W}(x;p))\parallel_2^2,
    \label{feature-ic-lk}
 \end{equation}
where $F_T$ and $F_I$ are the feature maps of the template and input images. Then, they enforced the network to run a single iteration with a hinge loss, while the network runs multiple iterations until the stopping condition is met in the testing stage. Besides, CLKN stacked three similar LK networks to boost the performance by treating the output of the last LK network as the initial warp parameters of the next LK network. From Eq. \ref{feature-ic-lk}, the IC-LK algorithm heavily relied on feature maps, which tend to fail in multimodal images. Instead, DLKFM \cite{DLKFM} constructed a single-channel feature map using the eigenvalues of the local covariance matrix on the output tensor. It designed two special constraint terms to align multimodal feature maps and contribute to convergence.

However, LK-based algorithms can fail if the Jacobian matrix is rank-deficient \cite{nocedal1999numerical}. Additionally, the IC-LK iterator is untrainable, which means this drawback is theoretically unavoidable. To address this issue, a completely trainable iterative homography network (IHN) \cite{IHN} was proposed. Inspired by RAFT \cite{teed2020raft}, IHN updates the cost volume to refine the estimated homography using the same estimator repeatedly every iteration. Furthermore, IHN can handle dynamic scenes by producing an inlier mask in the estimator without requiring extra supervision. The following works further explore more efficient and holistic frameworks for the cross-modal homography estimation~\cite{SCPNet, MCNet, InterNet, CrossHomo}.

\subsection{Discussion}
\label{subsec:discussion}
\subsubsection{Technique Summary}

The above works are devoted to exploring different homography parameterizations such as 4-pt parameterization\cite{DHN}, perspective field\cite{PFNet}, and motion bases representation\cite{BasesHomo}, which contributes to better convergence and performance. Others tend to design various network architectures. In particular, cascaded and iterative solutions are proposed to refine the performance progressively, which can be combined jointly to reach higher accuracy. To make the methods more practical, various challenging problems are preliminarily addressed, such as cross resolutions\cite{LocalTrans}, multiple modalities\cite{DLKFM, IHN, CrossHomo, AltO}, dynamic objects\cite{MHN, IHN}, and non-planar scenes\cite{CA-UDHN, HomoGAN, DAMG-Homo, Mask-Homo}, etc.

\subsubsection{Challenge and Future Effort}
We summarize the existing challenges as follows:

(1) Many homography estimation solutions are designed for fixed resolutions, while real-world applications often involve much more flexible resolutions. When pre-trained models are applied to images with different resolutions, performance can dramatically drop due to the need for input resizing to satisfy the regulated resolution.

(2) Unlike optical flow estimation, which assumes small motions between images, homography estimation often deals with images that have significantly low-overlap rates. In such cases, existing methods may exhibit inferior performance due to limited receptive fields.

(3) Existing methods address the parallax or dynamic objects by learning to reject outliers in the feature extractor\cite{CA-UDHN}, cost volume\cite{li2022ssorn}, or estimator\cite{IHN}. However, it is still unclear which stage is more appropriate for outlier rejection.

Based on the challenges we have discussed, some potential research directions for future efforts can be identified:

(1) To overcome the first challenge, we can design various strategies to enhance resolution robustness, such as resolution-related data augmentation, and continual learning on multiple datasets with different resolutions. Besides, we can also formulate a resolution-free parameterization form. The perspective field \cite{PFNet} is a typical case, which represents the homography as dense correspondences with the same resolution as input images. But it requires RANSAC as the post-processing approach, introducing extra computational complexity, especially in extensive correspondences. Therefore, a resolution-free and efficient parameterization form could be explored.

(2) To enhance the performance in low-overlap rate, the main insight is to increase the receptive fields of a network. To this end, the cross-attention module of the Transformer explicitly leverages the long-range correlation to eliminate short-range inductive bias\cite{vaswani2017attention}. On the other hand, we can exploit beneficial varieties of cost volume to integrate feature correlation \cite{DAMG-Homo, IHN}.

(3) As there is no interaction between different image features in the feature extractor, it is reasonable to assume that outlier rejection should occur after feature extraction. Identifying outliers within a single image is impossible as the depth alone cannot be used as an outlier cue. For example, images captured by purely rotated cameras do not contain parallax outliers. Additionally, it seems intuitive to learn the capability of outlier rejection by combining global and local correlation, similar to the insight of RANSAC.
	\section{Cross-Sensor Model}
\label{sec:hybrid}

Multi-sensor calibration estimates intrinsic and extrinsic parameters of multiple sensors like cameras, LiDARs, and IMUs. This ensures that data from different sensors are synchronized and registered in a common coordinate system, allowing them to be fused together for a more accurate representation of the environment. Accurate multi-sensor calibration is crucial for applications like autonomous driving and robotics, where reliable sensor fusion is necessary for safe and efficient operation.

In this part, we mainly review the literature on learning-based camera-LiDAR calibration, \textit{i.e.}, predicting the 6-DoF rigid body transformation between a camera and a 3D LiDAR, without requiring any presence of specific features or landmarks in the implementation. Like the works on other types of cameras/systems, this research field can also be classified into regression-based solutions and flow/reconstruction-based solutions. But we are prone to follow the special \textit{matching} principle in camera-LiDAR calibration and divide the existing learning-based literature into three categories: pixel-level, semantics-level, and object/keypoint-level solutions.

\subsection{Pixel-level Solution}
The first deep learning technique in camera-LiDAR calibration, RegNet~\cite{schneider2017regnet}, used CNNs to combine feature extraction, feature matching, and global regression to infer the 6-DoF extrinsic parameters. It processed the RGB and LiDAR depth maps separately and branched two parallel data network streams. Then, a specific correlation layer was proposed to convolve the stacked LiDAR and RGB features as a joint representation. Subsequently, the global information fusion and parameter regression were achieved by two fully connected layers with an Euclidean loss function. Motivated by this work, the following works made a further step into more accurate calibration in terms of the geometric constraint~\cite{iyer2018calibnet,shi2020calibrcnn}, temporal correlation~\cite{shi2020calibrcnn}, loss design~\cite{yuan2020rggnet}, feature extraction~\cite{wang2022fusionnet}, feature matching~\cite{lv2021lccnet, wu2021netcalib}, feature fusion~\cite{wang2022fusionnet}, and calibration representation~\cite{lv2021cfnet, jing2022dxq}, etc.

\begin{figure}[!t]
  \centering
  \includegraphics[width=.47\textwidth]{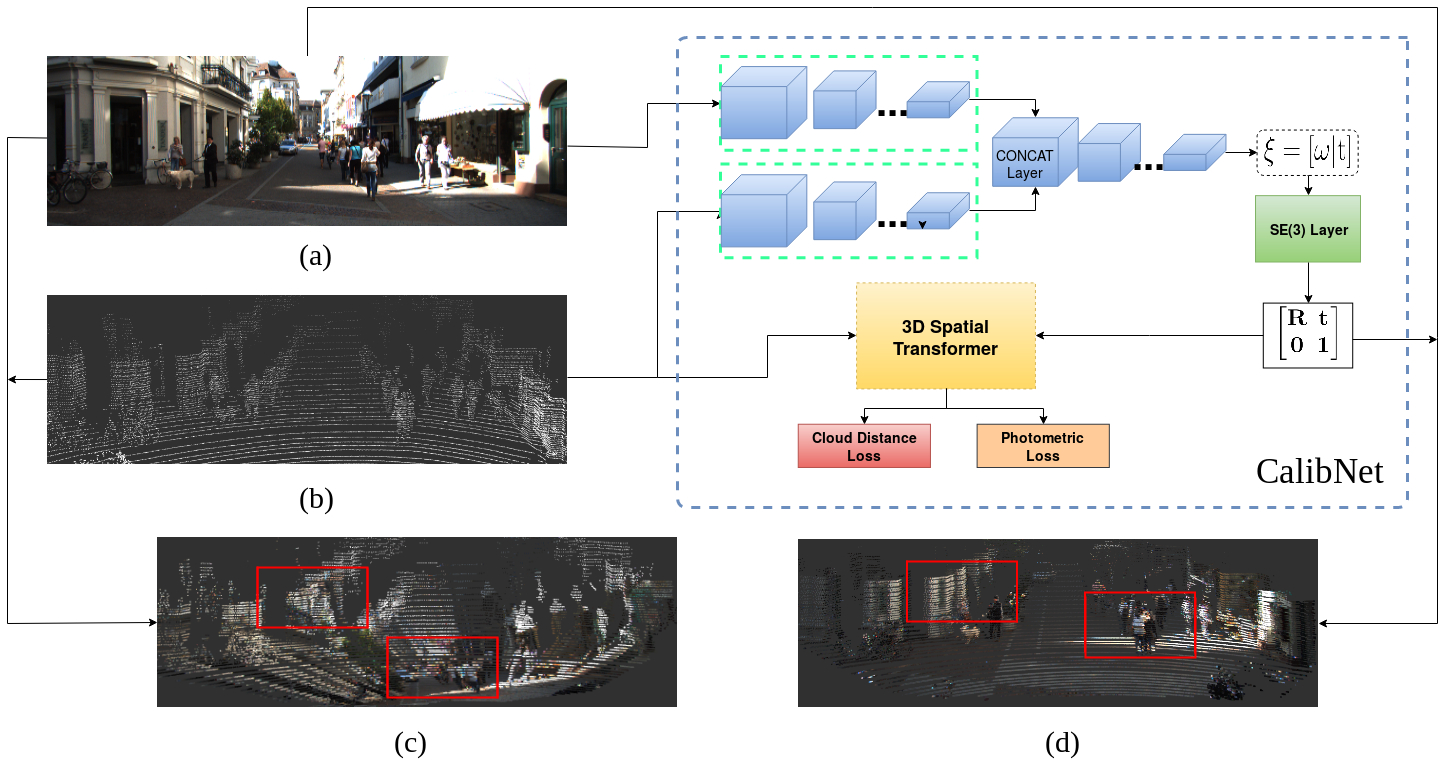}
  \caption{Network architecture of CalibNet. The figure is from ~\cite{iyer2018calibnet}. It takes an RGB image from a calibrated camera and a raw LiDAR point cloud as inputs, and regresses a 6-DoF transformation by an SE(3) layer.}
  \label{fig:CalibNet}
  \vspace{-0.3cm}
\end{figure}

For example, in Figure~\ref{fig:CalibNet}, CalibNet~\cite{iyer2018calibnet} designed a network to predict calibration parameters that maximize the geometric and photometric consistency of images and point clouds, solving the underlying physical problem by 3D Spatial Transformers~\cite{handa2016gvnn}. To refine the calibration model, CalibRCNN~\cite{shi2020calibrcnn} presented a synthetic view and an epipolar geometry constraint to measure the photometric and geometric inaccuracies between consecutive frames, of which the temporal information learned by the LSTM network has been investigated in the learning-based camera-LiDAR calibration. Since the output space of the LiDAR-camera calibration is on the 3D Special Euclidean Group ($SE(3)$) rather than the normal Euclidean space, RGGNet~\cite{yuan2020rggnet} considered Riemannian geometry constraints in the loss function, namely, used an $SE(3)$ geodesic distance equipped with left-invariant Riemannian metrics to optimize the calibration network. LCCNet~\cite{lv2021lccnet} exploited the cost volume layer to learn the correlation between the image and the depth transformed by the point cloud. Because the depth map ignores the 3D geometric structure of the point cloud, FusionNet~\cite{wang2022fusionnet} leveraged PointNet++~\cite{qi2017pointnet++} to directly learn the features from the 3D point cloud. Subsequently, a feature fusion with Ball Query~\cite{qi2017pointnet++} and attention strategy was proposed to effectively fuse the features of images and point clouds. CFNet~\cite{lv2021cfnet} first proposed the calibration flow for camera-LiDAR calibration, which represents the deviation between the positions of initial projected 2D points and ground truth. Compared to directly predicting extrinsic parameters, learning the calibration flow helps the network understand the underlying geometric constraint. To build precise 2D-3D correspondences, CFNet~\cite{lv2021cfnet} corrected the originally projected points using the estimated calibration flow. Then the efficient Perspective-n-Point (EPnP) algorithm was applied to calculate the extrinsic parameters by RANSAC. Because RANSAC is nondifferentiable, DXQ-Net~\cite{jing2022dxq} presented a probabilistic model to estimate the uncertainty to measure the quality of LiDAR-camera data association. Then, the differentiable pose estimation module was designed for solving extrinsic parameters, back-propagating the extrinsic error to the flow prediction network. Recent efforts further improve the calibration performance by presenting/introducing cross-modal graph neural network~\cite{RobustCalib}, multi-head correlation module~\cite{CalibFormer}, vanishing point and horizon line estimation~\cite{SensorX2Vehicle}, monocular depth estimation~\cite{Calibdepth}, homogeneous local-global aware representation~\cite{LCANet}, and hierarchical and iterative feature matching~\cite{HIFMNet}, etc.

\subsection{Semantics-level Solution}
Semantic features can be well learned and represented by neural networks. A perfect calibration enables to accurately align the same instance in different sensors. To this end, some works~\cite{wang2020soic, zhu2020online, liu2021semalign, SST-Calib, SE-Calib, Edgecalib} explored to guide the calibration with the semantic information. SOIC~\cite{wang2020soic} calibrated and transformed the initialization issue into the semantic centroids' PnP problem. Since the 3D semantic centroids of the point cloud and the 2D semantic centroids of the image cannot match precisely, a matching constraint cost function based on the semantic components was presented. SSI-Calib~\cite{zhu2020online} reformulated the calibration as an optimization problem with a novel calibration quality metric based on semantic features. Then, a non-monotonic subgradient ascent was proposed to calculate the calibration parameters. Other works used the off-the-shelf segmentation networks~\cite{SAM-Calib} and optimized the calibration parameters by minimizing semantic alignment loss in single-direction~\cite{liu2021semalign} and bi-direction~\cite{SST-Calib}.

\subsection{Object/Keypoint-level Solution}
ATOP~\cite{ATOP} designed an attention-based object-level matching network, \textit{i.e.}, Cross-Modal Matching Network to explore the overlapped FoV between camera and LiDAR, which facilitated generating the 2D-3D object-level correspondences. 2D and 3D object proposals were
detected by YOLOv4~\cite{bochkovskiy2020yolov4} and PointPillar~\cite{lang2019pointpillars}. Then, two cascaded PSO-based algorithms~\cite{poli2007particle} were devised to estimate the calibration extrinsic parameters in the optimization stage. Using the deep declarative network (DDN)~\cite{gould2021deep}, RGKCNet~\cite{RGKCNet} combined the standard neural layer and a PnP solver in the same network, formulating the 2D–3D data association and pose estimation as a bilevel optimization problem. Thus, both the feature extraction capability of the convolutional layer and the conventional geometric solver can be employed. Additionally, RGKCNet~\cite{RGKCNet} presented a learnable weight layer that determines the keypoints involved in the solver, enabling the whole pipeline to be trained end-to-end. P2O-Calib~\cite{P2O-Calib} proposed a target-less calibration approach using the 2D-3D edge point extraction, which effectively investigates the occlusion relationship in 3D space.

\subsection{Discussion}
\subsubsection{Technique Summary}
The current method can be briefly classified based on the principle of building 2D and 3D matching, namely, the calibration target. In summary, most pixel-level solutions utilized the end-to-end framework to address this task. While these solutions delivered satisfactory performances on specific datasets, their generalization abilities are limited.
Semantics-level and object/keypoint-level methods derived from traditional calibration offered both acceptable performances and generalization abilities. However, they heavily relied on the quality of fore-end feature extraction~\cite{SAM-Calib}. Recent efforts~\cite{MOISST, SOAC, UniCal} jointly learn the cross-modal calibration and the implicit scene representation~\cite{mildenhall2020nerf} using differentiable volume rendering, showing more scalable applications than existing calibration solutions.

\subsubsection{Research Trend}

(1) Network is becoming more complex with the use of different structures for feature extraction, matching, and fusion. Current methods employ strategies like multi-scale feature extraction, cross-modal interaction, cost-volume establishment, and confidence-guided fusion.

(2) Directly regressing 6-DoF parameters yields weak generalization ability. To overcome this, intermediate representations like calibration flow have been introduced. Additionally, calibration flow can handle non-rigid transformations that are common in real-world applications.

(3) Traditional methods require specific environments but have well-designed strategies. To balance accuracy and generalization, a combination of geometric solving algorithms and learning methods has been investigated.

\subsubsection{Future Effort}

(1) Camera-LiDAR calibration methods typically rely on datasets like KITTI, which provide only initial extrinsic parameters. To create a decalibration dataset, researchers add noise transformations to the initial extrinsics, but this approach assumes a fixed position camera-LiDAR system with miscalibration. In real-world applications, the camera-LiDAR relative pose varies, making it challenging to collect large-scale real data with ground truth extrinsics. To address this challenge, generating synthetic camera-LiDAR data using simulation systems could be a valuable solution.

(2) To optimize the combination of networks and traditional solutions, a more compact approach is needed. Current methods mainly use networks as feature extractors, resulting in non-end-to-end pipelines with inadequate feature extraction adjustments for calibration. A deep declarative network (DDN) is a promising framework for making the entire pipeline differentiable. The aggregation of learning and traditional methods can be optimized using DDN.

(3) The most important aspect of camera-LiDAR calibration is 2D-3D matching. To achieve this, the point cloud is commonly transformed into a depth image. However, large deviations in extrinsic simulation can result in detail loss. With the great development of Transformer and cross-modal techniques, we believe leveraging Transformer to directly learn the features of image and point cloud in the same pipeline could facilitate better 2D-3D matching. 
	\section{Benchmark}
\label{sec:evaluation}

\begin{figure*}[!t]
  \centering
  \includegraphics[width=1\textwidth]{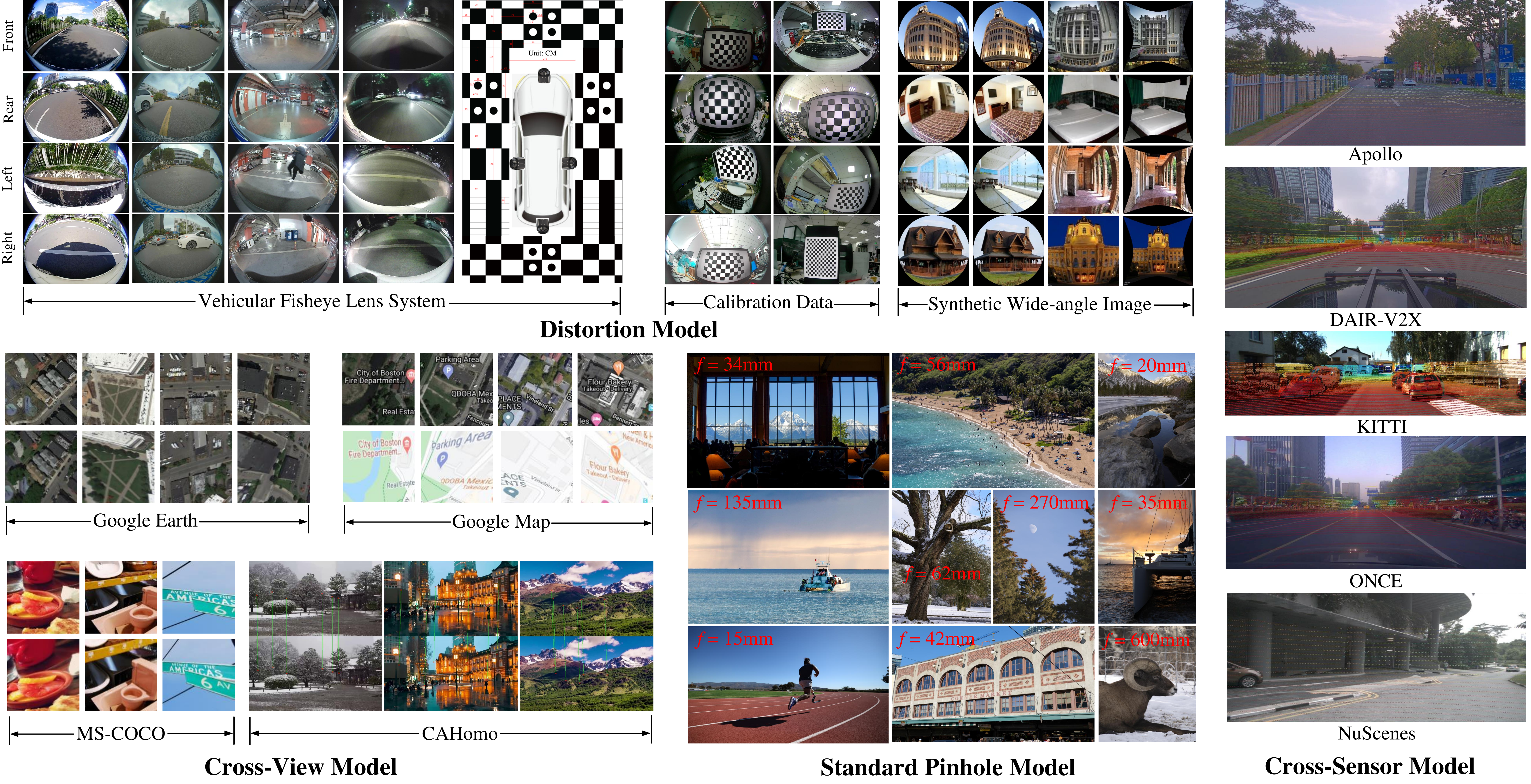}
  \caption{Overview of our collected benchmark, which covers all models reviewed in this paper. In this dataset, the image and video derive from diverse cameras under different environments. The accurate ground truth and label are provided for each sample.}
  \label{fig:benchmark}
  \vspace{-0.3cm}
\end{figure*}

As there is no public and unified benchmark in learning-based camera calibration, we contribute a dataset that can serve as a platform for generalization evaluations. In this dataset, the images and videos are captured by different cameras under diverse scenes, including simulation environments and real-world settings. Besides, we provide the calibration ground truth, parameter label, and visual clues based on different conditions. Figure~\ref{fig:benchmark} shows some samples of our collected dataset. Please refer to the evaluation of representative calibration methods on this benchmark in supplementary material.

\textbf{Standard Model}. We collected 300 high-resolution images on the Internet, captured by popular digital cameras such as Canon, Fujifilm, Nikon, Olympus, Sigma, Sony, etc. For each image, we provide the specific focal length of its lens. We have included a diverse range of subjects, including landscapes, portraits, wildlife, architecture, etc. The range of focal length is from 4.5mm to 600mm.
  
\textbf{Distortion Model}. We created a comprehensive dataset for the distortion camera model, with a focus on wide-angle cameras. The dataset is comprised of three subcategories. The first is a synthetic dataset, which was generated using the widely-used 4$^{th}$ order polynomial model. It contains both circular and rectangular structures, with 1,000 distortion-rectification image pairs. The second consists of data captured under real-world settings, derived from the raw calibration data for around 40 types of wide-angle cameras. For each calibration data, the intrinsics, extrinsics, and distortion coefficients are provided. Finally, we exploited a car equipped with different cameras to capture video sequences. The scenes cover both indoor and outdoor, including daytime and nighttime footage.
  
\textbf{Cross-View Model}. We selected 500 testing samples at random from each of four representative datasets (MS-COCO~\cite{DHN}, GoogleEarch~\cite{DLKFM}, GoogleMap~\cite{DLKFM}, CAHomo~\cite{CA-UDHN}) to create a dataset for the cross-view model. It covers a range of scenarios: MS-COCO provides natural synthetic data, GoogleEarch contains aerial synthetic data, and GoogleMap offers multi-modal synthetic data. Parallax is not a factor in these three datasets, while CAHomo provides real-world data with non-planar scenes. To standardize the dataset, we converted all images to a unified format and recorded the matched points between two views. In MS-COCO, GoogleEarch, and GoogleMap, we used four vertices of the images as the matched points. In CAHomo, we identified six matched key points within the same plane.

\textbf{Cross-Sensor Model}. We collected RGB and point cloud data from Apollo~\cite{huang2019apolloscape}, DAIR-V2X~\cite{yu2022dair}, KITTI~\cite{KITTI}, KUCL~\cite{kang-2020-jfr}, NuScenes~\cite{caesar2020nuscenes}, and ONCE~\cite{mao2021one}. Around 300 data pairs with calibration parameters are included in each category. The datasets are captured in different countries to provide enough variety. Each dataset has a different sensor setup, obtaining camera-LiDAR data with varying image resolution, LiDAR scan pattern, and camera-LiDAR relative location. The image resolution ranges from 2448$\times$2048 to 1242$\times$375, while the LiDAR sensors are from Velodyne and Hesai, with 16, 32, 40, 64, and 128 beams. They include not only normal surrounding multi-view images but also small baseline multi-view data. Additionally, we also added random disturbance of around 20 degrees rotation and 1.5 meters translation based on classical settings~\cite{schneider2017regnet} to simulate vibration and collision.

	\section{Future Research Directions}
\label{sec:Future}
Camera calibration is a fundamental and challenging research topic. From the above technical reviews and limitation analysis, we can conclude there is still room for improvement with deep learning. From Section~\ref{sec:pure} to Section~\ref{sec:hybrid}, specific future efforts are discussed for each model. In this section, we suggest more general future research directions. 
\subsection{Sequences}
Bundle adjustment is a well-established technique central to Multi-View Stereo (MVS) and Simultaneous Localization and Mapping (SLAM) using multi-view constraints. Traditional bundle adjustment focuses on pose estimation, often under the assumption of pre-calibrated cameras, thus sidelining the nuances of camera parameter fine-tuning. While learning-based camera calibration has made significant strides, most methods are tailored for a single image. We have highlighted intrinsics calibration to underscore how sequence constraints bolster prediction accuracy. Notably, there is a burgeoning interest in integrating bundle adjustment into end-to-end deep learning pipelines. By transitioning from conventional keypoint extraction and matching to learning-based methods, recent works~\cite{tang2018ba, teed2018deepv2d,wei2020deepsfm,gu2023dro, DroidCalib, teed2021droid, teed2021tangent} propose differentiable bundle adjustment layers to refine pose, depth, and camera parameters together. Consequently, there is immense potential in further harnessing sequence constraints for accurate calibration. Current methods combine front-end matching with a back-end solver, which can be inefficient and unreliable in cases like fast motion. We suggest separating front-end and back-end refinements, using large models for features and introducing more trainable parameters in optimization.
\subsection{Learning Target}
Due to the implicit relationship to image features, conventional calibration objectives can be challenging for neural networks to learn. To this end, some works have developed novel learning targets that replace conventional calibration objectives, providing learning-friendly representations for neural networks. Additionally, intermediate geometric representations have been presented to bridge the gap between image features and calibration objectives, such as reflective amplitude coefficient maps~\cite{Zheng}, rectification flow~\cite{BlindCor}, surface geometry~\cite{UprightNet}, and normal flow~\cite{DiffPoseNet}, etc. Looking ahead to the future development of this community, we believe there is still great potential for designing more explicit and reasonable learning targets for calibration objectives.
\subsection{Pre-training}
Pre-training on ImageNet~\cite{ImageNet} has become a widely used strategy in deep learning. However, recent studies~\cite{STD} have shown that this approach provides less benefit for specific camera calibration tasks, such as wide-angle camera calibration. This is due to two main reasons: the data gap and the task gap. The ImageNet dataset only contains perspective images without distortions, making the initialized weights of networks irrelevant to distortion models. Furthermore, He et al.~\cite{he_rethinking} demonstrated that the task of ImageNet pre-training has limited benefits when the final task is more sensitive to localization. As a result, the performance of extrinsics estimation may be impacted by this task gap. Moreover, pre-training beyond a single image and a single modality, to our knowledge, has not been investigated in the related field. We suggest that designing a customized pre-training strategy for camera calibration is an interesting area of research.

\subsection{Implicit Unified Model}
Deep learning-based camera calibration methods use traditional parametric camera models, which lack the flexibility to fit complex situations. Non-parametric camera models relate each pixel to its corresponding 3D observation ray, overcoming parametric model limitations. However, they require strict calibration targets and are more complex for undistortion, projection, and unprojection. Deep learning methods show potential for calibration tasks, making non-parametric models worth revisiting and potentially replacing parametric models. Moreover, they allow for implicit and unified calibration, fitting all camera types through pixel-level regression and avoiding explicit feature extraction and geometry solving. Researchers combined the advantages of implicit and unified representation with the Neural Radiance Field (NeRF) for reconstructing 3D structures and synthesizing novel views. Self-calibration NeRF~\cite{jeong2021self} has been proposed for generic cameras with arbitrary non-linear distortions, and end-to-end pipelines have been explored to learn depth and ego-motion without calibration targets. We believe the implicit and unified camera models could be used to optimize learning-based algorithms or integrated into downstream 3D vision tasks.

\section{Conclusion}
In this paper, we present a comprehensive survey of the recent efforts in deep learning-based camera calibration. Our survey covers conventional camera models, classified learning paradigms and learning strategies, detailed reviews of the state-of-the-art approach, a public benchmark, and future research directions. To exhibit the development process and link the connections between existing works, we provide a fine-grained taxonomy that categorizes literature by jointly considering camera models and applications. Moreover, the relationships, strengths, distinctions, and limitations are thoroughly discussed in each category. An open-source repository will keep updating regularly with new works and datasets. We hope that this survey could promote future research in this field. 

\section*{Acknowledgment}
We thank Leidong Qin and Shangrong Yang at Beijing Jiaotong University for the partial dataset collection. We thank Jinlong Fan at the University of Sydney for the insightful discussion. We appreciate all reviewers for their insightful comments and invaluable suggestions.

	{
		\bibliographystyle{IEEEtran}
		\bibliography{bib}

\begin{thebibliography}{100}
\providecommand{\url}[1]{#1}
\csname url@samestyle\endcsname
\providecommand{\newblock}{\relax}
\providecommand{\bibinfo}[2]{#2}
\providecommand{\BIBentrySTDinterwordspacing}{\spaceskip=0pt\relax}
\providecommand{\BIBentryALTinterwordstretchfactor}{4}
\providecommand{\BIBentryALTinterwordspacing}{\spaceskip=\fontdimen2\font plus
\BIBentryALTinterwordstretchfactor\fontdimen3\font minus \fontdimen4\font\relax}
\providecommand{\BIBforeignlanguage}[2]{{%
\expandafter\ifx\csname l@#1\endcsname\relax
\typeout{** WARNING: IEEEtran.bst: No hyphenation pattern has been}%
\typeout{** loaded for the language `#1'. Using the pattern for}%
\typeout{** the default language instead.}%
\else
\language=\csname l@#1\endcsname
\fi
#2}}
\providecommand{\BIBdecl}{\relax}
\BIBdecl

\bibitem{duane1971close}
C.~B. Duane, ``Close-range camera calibration,'' \emph{Photogramm. Eng}, vol.~37, no.~8, pp. 855--866, 1971.

\bibitem{maybank1992theory}
S.~J. Maybank and O.~D. Faugeras, ``A theory of self-calibration of a moving camera,'' \emph{International Journal of Computer Vision}, vol.~8, no.~2, pp. 123--151, 1992.

\bibitem{weng1992camera}
J.~Weng, P.~Cohen, M.~Herniou \emph{et~al.}, ``Camera calibration with distortion models and accuracy evaluation,'' \emph{IEEE Transactions on Pattern Analysis and Machine Intelligence}, vol.~14, no.~10, pp. 965--980, 1992.

\bibitem{zhang2000flexible}
Z.~Zhang, ``A flexible new technique for camera calibration,'' \emph{IEEE Transactions on Pattern Analysis and Machine Intelligence}, vol.~22, no.~11, pp. 1330--1334, 2000.

\bibitem{qi2023minimal}
Z.~Qi, J.~Pang, Y.~Hao, R.~Hu, and Y.~Zhang, ``A minimal solution for sphere-based camera-projector pair calibration,'' \emph{IEEE Transactions on Circuits and Systems for Video Technology}, 2023.

\bibitem{brown1966decentering}
D.~C. Brown, ``Decentering distortion of lenses,'' \emph{Photogrammetric Engineering and Remote Sensing}, 1966.

\bibitem{zhang1999flexible}
Z.~Zhang, ``Flexible camera calibration by viewing a plane from unknown orientations,'' in \emph{International Conference on Computer Vision}, vol.~1, 1999, pp. 666--673.

\bibitem{gasparini2009plane}
S.~Gasparini, P.~Sturm, and J.~P. Barreto, ``Plane-based calibration of central catadioptric cameras,'' in \emph{International Conference on Computer Vision}, 2009, pp. 1195--1202.

\bibitem{shah1994simple}
S.~Shah and J.~Aggarwal, ``A simple calibration procedure for fish-eye (high distortion) lens camera,'' in \emph{Proceedings of IEEE International Conference on Robotics and Automation}, 1994, pp. 3422--3427.

\bibitem{barreto2005geometric}
J.~P. Barreto and H.~Araujo, ``Geometric properties of central catadioptric line images and their application in calibration,'' \emph{IEEE Transactions on Pattern Analysis and Machine Intelligence}, vol.~27, no.~8, pp. 1327--1333, 2005.

\bibitem{Carroll}
R.~Carroll, M.~Agrawal, and A.~Agarwala, ``Optimizing content-preserving projections for wide-angle images,'' in \emph{ACM Transactions on Graphics}, vol.~28, no.~3, 2009, p.~43.

\bibitem{Bukhari}
F.~Bukhari and M.~N. Dailey, ``Automatic radial distortion estimation from a single image,'' \emph{Journal of Mathematical Imaging and Vision}, vol.~45, no.~1, pp. 31--45, 2013.

\bibitem{Miguel}
M.~Alem{\'a}n-Flores, L.~Alvarez, L.~Gomez, and D.~Santana-Cedr{\'e}s, ``Automatic lens distortion correction using one-parameter division models,'' \emph{Image Processing On Line}, vol.~4, pp. 327--343, 2014.

\bibitem{faugeras1992camera}
O.~D. Faugeras, Q.-T. Luong, and S.~J. Maybank, ``Camera self-calibration: Theory and experiments,'' in \emph{European Conference on Computer Vision}, 1992, pp. 321--334.

\bibitem{fraser1997digital}
C.~S. Fraser, ``Digital camera self-calibration,'' \emph{ISPRS Journal of Photogrammetry and Remote sensing}, vol.~52, no.~4, pp. 149--159, 1997.

\bibitem{hartley1994self}
R.~I. Hartley, ``Self-calibration from multiple views with a rotating camera,'' in \emph{European Conference on Computer Vision}, 1994, pp. 471--478.

\bibitem{camposeco2015non}
F.~Camposeco, T.~Sattler, and M.~Pollefeys, ``Non-parametric structure-based calibration of radially symmetric cameras,'' in \emph{International Conference on Computer Vision}, 2015, pp. 2192--2200.

\bibitem{schops2020having}
T.~Schops, V.~Larsson, M.~Pollefeys, and T.~Sattler, ``Why having 10,000 parameters in your camera model is better than twelve,'' in \emph{Proceedings of the IEEE/CVF Conference on Computer Vision and Pattern Recognition}, 2020, pp. 2535--2544.

\bibitem{pan2022camera}
L.~Pan, M.~Pollefeys, and V.~Larsson, ``Camera pose estimation using implicit distortion models,'' in \emph{Proceedings of the IEEE/CVF Conference on Computer Vision and Pattern Recognition}, 2022, pp. 12\,819--12\,828.

\bibitem{opencv}
\BIBentryALTinterwordspacing
 [Online]. Available: \url{https://docs.opencv.org/4.x/dc/dbb/tutorial_py_calibration.html}
\BIBentrySTDinterwordspacing

\bibitem{matlab}
\BIBentryALTinterwordspacing
 [Online]. Available: \url{https://www.mathworks.com/help/vision/camera-calibration.html}
\BIBentrySTDinterwordspacing

\bibitem{salvi2002comparative}
J.~Salvi, X.~Armangu{\'e}, and J.~Batlle, ``A comparative review of camera calibrating methods with accuracy evaluation,'' \emph{Pattern Recognition}, vol.~35, no.~7, pp. 1617--1635, 2002.

\bibitem{hughes2008review}
C.~Hughes, M.~Glavin, E.~Jones, and P.~Denny, ``Review of geometric distortion compensation in fish-eye cameras,'' 2008.

\bibitem{fan2022wide}
J.~Fan, J.~Zhang, S.~J. Maybank, and D.~Tao, ``Wide-angle image rectification: a survey,'' \emph{International Journal of Computer Vision}, vol. 130, no.~3, pp. 747--776, 2022.

\bibitem{DeepFocal}
S.~Workman, C.~Greenwell, M.~Zhai, R.~Baltenberger, and N.~Jacobs, ``Deepfocal: A method for direct focal length estimation,'' in \emph{IEEE International Conference on Image Processing}, 2015, pp. 1369--1373.

\bibitem{PoseNet}
A.~Kendall, M.~Grimes, and R.~Cipolla, ``Posenet: A convolutional network for real-time 6-dof camera relocalization,'' in \emph{International Conference on Computer Vision}, 2015.

\bibitem{Rong}
J.~Rong, S.~Huang, Z.~Shang, and X.~Ying, ``Radial lens distortion correction using convolutional neural networks trained with synthesized images,'' in \emph{Asian Conference on Computer Vision}, 2016, pp. 35--49.

\bibitem{URS-CNN}
V.~Rengarajan, Y.~Balaji, and A.~Rajagopalan, ``Unrolling the shutter: Cnn to correct motion distortions,'' in \emph{Proceedings of the IEEE Conference on Computer Vision and Pattern Recognition}, 2017, pp. 2291--2299.

\bibitem{DHN}
D.~DeTone, T.~Malisiewicz, and A.~Rabinovich, ``Deep image homography estimation,'' \emph{arXiv preprint arXiv:1606.03798}, 2016.

\bibitem{Hold-Geoffroy}
Y.~Hold-Geoffroy, K.~Sunkavalli, J.~Eisenmann, M.~Fisher, E.~Gambaretto, S.~Hadap, and J.-F. Lalonde, ``A perceptual measure for deep single image camera calibration,'' in \emph{Proceedings of the IEEE Conference on Computer Vision and Pattern Recognition}, 2018.

\bibitem{schneider2017regnet}
N.~Schneider, F.~Piewak, C.~Stiller, and U.~Franke, ``Regnet: Multimodal sensor registration using deep neural networks,'' in \emph{IEEE intelligent vehicles symposium}, 2017, pp. 1803--1810.

\bibitem{pix2pix}
P.~Isola, J.-Y. Zhu, T.~Zhou, and A.~A. Efros, ``Image-to-image translation with conditional adversarial networks,'' in \emph{Proceedings of IEEE Conference on Computer Vision and Pattern Recognition}, 2017, pp. 1125--1134.

\bibitem{long2015fully}
J.~Long, E.~Shelhamer, and T.~Darrell, ``Fully convolutional networks for semantic segmentation,'' in \emph{Proceedings of the IEEE Conference on Computer Vision and Pattern Recognition}, 2015, pp. 3431--3440.

\bibitem{eigen2014depth}
D.~Eigen, C.~Puhrsch, and R.~Fergus, ``Depth map prediction from a single image using a multi-scale deep network,'' \emph{Advances in Neural Information Processing Systems}, vol.~27, 2014.

\bibitem{DR-GAN}
K.~Liao, C.~Lin, Y.~Zhao, and M.~Gabbouj, ``Dr-gan: Automatic radial distortion rectification using conditional gan in real-time,'' \emph{IEEE Transactions on Circuits and Systems for Video Technology}, vol.~30, no.~3, pp. 725--733, 2020.

\bibitem{DDM}
K.~Liao, C.~Lin, Y.~Zhao, and M.~Xu, ``Model-free distortion rectification framework bridged by distortion distribution map,'' \emph{IEEE Transactions on Image Processing}, vol.~29, pp. 3707--3718, 2020.

\bibitem{DaRecNet}
K.~Liao, C.~Lin, L.~Liao, Y.~Zhao, and W.~Lin, ``Multi-level curriculum for training a distortion-aware barrel distortion rectification model,'' in \emph{International Conference on Computer Vision}, 2021, pp. 4389--4398.

\bibitem{BlindCor}
X.~Li, B.~Zhang, P.~V. Sander, and J.~Liao, ``Blind geometric distortion correction on images through deep learning,'' in \emph{Proceedings of the IEEE/CVF Conference on Computer Vision and Pattern Recognition}, 2019.

\bibitem{ronneberger2015u}
O.~Ronneberger, P.~Fischer, and T.~Brox, ``U-net: Convolutional networks for biomedical image segmentation,'' in \emph{International Conference on Medical Image Computing and Computer-Assisted Intervention}, 2015, pp. 234--241.

\bibitem{CAR}
J.~Y. Zhang, A.~Lin, M.~Kumar, T.-H. Yang, D.~Ramanan, and S.~Tulsiani, ``Cameras as rays: Pose estimation via ray diffusion,'' in \emph{International Conference on Learning Representations}, 2024.

\bibitem{DiffCalib}
X.~He, G.~Xu, B.~Zhang, H.~Chen, Y.~Cui, and D.~Guo, ``Diffcalib: Reformulating monocular camera calibration as diffusion-based dense incident map generation,'' \emph{arXiv preprint arXiv:2405.15619}, 2024.

\bibitem{DM-Calib}
J.~Deng, W.~Yin, X.~Guo, Q.~Zhang, X.~Hu, W.~Ren, X.~Long, and P.~Tan, ``Boost 3d reconstruction using diffusion-based monocular camera calibration,'' \emph{arXiv preprint arXiv:2411.17240}, 2024.

\bibitem{RS-Diffusion}
Z.~Yang, H.~Li, M.~Hong, B.~Zeng, and S.~Liu, ``Single image rolling shutter removal with diffusion models,'' \emph{arXiv preprint arXiv:2407.02906}, 2024.

\bibitem{DeepVP}
M.~Zhai, S.~Workman, and N.~Jacobs, ``Detecting vanishing points using global image context in a non-manhattan world,'' in \emph{Proceedings of the IEEE Conference on Computer Vision and Pattern Recognition}, 2016.

\bibitem{DeepCalib}
O.~Bogdan, V.~Eckstein, F.~Rameau, and J.-C. Bazin, ``Deepcalib: a deep learning approach for automatic intrinsic calibration of wide field-of-view cameras,'' in \emph{Proceedings of the 15th ACM SIGGRAPH European Conference on Visual Media Production}, 2018.

\bibitem{DVPD}
Y.~Lin, R.~Wiersma, S.~L. Pintea, K.~Hildebrandt, E.~Eisemann, and J.~C. van Gemert, ``Deep vanishing point detection: Geometric priors make dataset variations vanish,'' in \emph{Proceedings of the IEEE/CVF Conference on Computer Vision and Pattern Recognition}, 2022, pp. 6103--6113.

\bibitem{EvUnroll}
X.~Zhou, P.~Duan, Y.~Ma, and B.~Shi, ``Evunroll: Neuromorphic events based rolling shutter image correction,'' in \emph{Proceedings of the IEEE/CVF Conference on Computer Vision and Pattern Recognition}, 2022, pp. 17\,775--17\,784.

\bibitem{FishFormer}
Y.~Shangrong, L.~Chunyu, L.~Kang, and Z.~Yao, ``Fishformer: Annulus slicing-based transformer for fisheye rectification with efficacy domain exploration,'' \emph{arXiv preprint arXiv:2207.01925}, 2022.

\bibitem{DAMG-Homo}
L.~Nie, C.~Lin, K.~Liao, S.~Liu, and Y.~Zhao, ``Depth-aware multi-grid deep homography estimation with contextual correlation,'' \emph{IEEE Transactions on Circuits and Systems for Video Technology}, vol.~32, no.~7, pp. 4460--4472, 2021.

\bibitem{SST-Calib}
K.~Akio, Z.~Yiyang, Z.~Pengwei, Z.~Wei, and T.~Masayoshi, ``Sst-calib: Simultaneous spatial-temporal parameter calibration between lidar and camera,'' in \emph{IEEE International Conference on Intelligent Transportation Systems}, 2022, pp. 2896--2902.

\bibitem{GeoCalib}
A.~Veicht, P.-E. Sarlin, P.~Lindenberger, and M.~Pollefeys, ``Geocalib: Learning single-image calibration with geometric optimization,'' in \emph{European Conference on Computer Vision}, 2024.

\bibitem{Zhao}
Y.~Zhao, Z.~Huang, T.~Li, W.~Chen, C.~LeGendre, X.~Ren, A.~Shapiro, and H.~Li, ``Learning perspective undistortion of portraits,'' in \emph{International Conference on Computer Vision}, 2019.

\bibitem{Tan}
J.~Tan, S.~Zhao, P.~Xiong, J.~Liu, H.~Fan, and S.~Liu, ``Practical wide-angle portraits correction with deep structured models,'' in \emph{Proceedings of the IEEE/CVF Conference on Computer Vision and Pattern Recognition}, 2021, pp. 3498--3506.

\bibitem{SPEC}
M.~Kocabas, C.-H.~P. Huang, J.~Tesch, L.~M\"uller, O.~Hilliges, and M.~J. Black, ``Spec: Seeing people in the wild with an estimated camera,'' in \emph{International Conference on Computer Vision}, 2021, pp. 11\,035--11\,045.

\bibitem{DeepUnrollNet}
P.~Liu, Z.~Cui, V.~Larsson, and M.~Pollefeys, ``Deep shutter unrolling network,'' in \emph{Proceedings of the IEEE/CVF Conference on Computer Vision and Pattern Recognition}, 2020, pp. 5941--5949.

\bibitem{SS-WPC}
F.~Zhu, S.~Zhao, P.~Wang, H.~Wang, H.~Yan, and S.~Liu, ``Semi-supervised wide-angle portraits correction by multi-scale transformer,'' in \emph{Proceedings of the IEEE/CVF Conference on Computer Vision and Pattern Recognition}, 2022, pp. 19\,689--19\,698.

\bibitem{Zhu}
R.~Zhu, X.~Yang, Y.~Hold-Geoffroy, F.~Perazzi, J.~Eisenmann, K.~Sunkavalli, and M.~Chandraker, ``Single view metrology in the wild,'' in \emph{European Conference on Computer Vision}, 2020, pp. 316--333.

\bibitem{UDHN}
T.~Nguyen, S.~W. Chen, S.~S. Shivakumar, C.~J. Taylor, and V.~Kumar, ``Unsupervised deep homography: A fast and robust homography estimation model,'' \emph{IEEE Robotics and Automation Letters}, vol.~3, no.~3, pp. 2346--2353, 2018.

\bibitem{CA-UDHN}
J.~Zhang, C.~Wang, S.~Liu, L.~Jia, N.~Ye, J.~Wang, J.~Zhou, and J.~Sun, ``Content-aware unsupervised deep homography estimation,'' in \emph{European Conference on Computer Vision}, 2020, pp. 653--669.

\bibitem{BasesHomo}
N.~Ye, C.~Wang, H.~Fan, and S.~Liu, ``Motion basis learning for unsupervised deep homography estimation with subspace projection,'' in \emph{International Conference on Computer Vision}, 2021, pp. 13\,117--13\,125.

\bibitem{HomoGAN}
M.~Hong, Y.~Lu, N.~Ye, C.~Lin, Q.~Zhao, and S.~Liu, ``Unsupervised homography estimation with coplanarity-aware gan,'' pp. 17\,663--17\,672, 2022.

\bibitem{Liu}
S.~Liu, N.~Ye, C.~Wang, J.~Zhang, L.~Jia, K.~Luo, J.~Wang, and J.~Sun, ``Content-aware unsupervised deep homography estimation and its extensions,'' \emph{IEEE Transactions on Pattern Analysis and Machine Intelligence}, vol.~45, no.~3, pp. 2849--2863, 2022.

\bibitem{UnFishCor}
S.~Yang, C.~Lin, K.~Liao, Y.~Zhao, and M.~Liu, ``Unsupervised fisheye image correction through bidirectional loss with geometric prior,'' \emph{Journal of Visual Communication and Image Representation}, vol.~66, p. 102692, 2020.

\bibitem{rani2023self}
V.~Rani, S.~T. Nabi, M.~Kumar, A.~Mittal, and K.~Kumar, ``Self-supervised learning: A succinct review,'' \emph{Archives of Computational Methods in Engineering}, vol.~30, no.~4, pp. 2761--2775, 2023.

\bibitem{SSR-Net}
C.~Wang, X.~Wang, X.~Bai, Y.~Liu, and J.~Zhou, ``Self-supervised deep homography estimation with invertibility constraints,'' \emph{Pattern Recognition Letters}, vol. 128, pp. 355--360, 2019.

\bibitem{SIR}
J.~Fan, J.~Zhang, and D.~Tao, ``Sir: Self-supervised image rectification via seeing the same scene from multiple different lenses,'' \emph{IEEE Transactions on Image Processing}, 2022.

\bibitem{Fang}
J.~Fang, I.~Vasiljevic, V.~Guizilini, R.~Ambrus, G.~Shakhnarovich, A.~Gaidon, and M.~R. Walter, ``Self-supervised camera self-calibration from video,'' pp. 8468--8475, 2022.

\bibitem{DQN-RecNet}
J.~Zhao, S.~Wei, L.~Liao, and Y.~Zhao, ``Dqn-based gradual fisheye image rectification,'' \emph{Pattern Recognition Letters}, vol. 152, pp. 129--134, 2021.

\bibitem{mnih2015human}
V.~Mnih, K.~Kavukcuoglu, D.~Silver, A.~A. Rusu, J.~Veness, M.~G. Bellemare, A.~Graves, M.~Riedmiller, A.~K. Fidjeland, G.~Ostrovski \emph{et~al.}, ``Human-level control through deep reinforcement learning,'' \emph{Nature}, vol. 518, no. 7540, pp. 529--533, 2015.

\bibitem{1DSfM}
K.~Wilson and N.~Snavely, ``Robust global translations with 1dsfm,'' in \emph{European Conference on Computer Vision}, 2014, pp. 61--75.

\bibitem{Cambridge_Landmarks}
\BIBentryALTinterwordspacing
 [Online]. Available: \url{https://www.repository.cam.ac.uk/handle/1810/251342;jsessionid=90AB1617B8707CD387CBF67437683F77}
\BIBentrySTDinterwordspacing

\bibitem{DeepHorizon}
S.~Workman, M.~Zhai, and N.~Jacobs, ``Horizon lines in the wild,'' \emph{British Machine Vision Conference}, 2016.

\bibitem{HLW}
\BIBentryALTinterwordspacing
 [Online]. Available: \url{https://mvrl.cse.wustl.edu/datasets/hlw/}
\BIBentrySTDinterwordspacing

\bibitem{YUD}
P.~Denis, J.~H. Elder, and F.~J. Estrada, ``Efficient edge-based methods for estimating manhattan frames in urban imagery,'' in \emph{European Conference on Computer Vision}, 2008, pp. 197--210.

\bibitem{ECD}
O.~Barinova, V.~Lempitsky, E.~Tretiak, and P.~Kohli, ``Geometric image parsing in man-made environments,'' in \emph{European Conference on Computer Vision}, 2010, pp. 57--70.

\bibitem{ImageNet}
J.~Deng, W.~Dong, R.~Socher, L.-J. Li, K.~Li, and L.~Fei-Fei, ``Imagenet: A large-scale hierarchical image database,'' in \emph{IEEE Conference on Computer Vision and Pattern Recognition}, 2009, pp. 248--255.

\bibitem{xiao2010sun}
J.~Xiao, J.~Hays, K.~A. Ehinger, A.~Oliva, and A.~Torralba, ``Sun database: Large-scale scene recognition from abbey to zoo,'' in \emph{IEEE Computer Society Conference on Computer Vision and Pattern Recognition}, 2010, pp. 3485--3492.

\bibitem{philbin2007object}
J.~Philbin, O.~Chum, M.~Isard, J.~Sivic, and A.~Zisserman, ``Object retrieval with large vocabularies and fast spatial matching,'' in \emph{2007 IEEE Conference on Computer Vision and Pattern Recognition}, 2007, pp. 1--8.

\bibitem{shao2003zubud}
H.~Shao, T.~Svoboda, and L.~Van~Gool, ``Zubud-zurich buildings database for image based recognition,'' \emph{Computer Vision Lab, Swiss Federal Institute of Technology}, vol. 260, no.~20, p.~6, 2003.

\bibitem{huang2008labeled}
G.~B. Huang, M.~Mattar, T.~Berg, and E.~Learned-Miller, ``Labeled faces in the wild: A database forstudying face recognition in unconstrained environments,'' in \emph{Workshop on Faces in'Real-Life'Images: Detection, Alignment, and Recognition}, 2008.

\bibitem{SUN360}
J.~Xiao, K.~A. Ehinger, A.~Oliva, and A.~Torralba, ``Recognizing scene viewpoint using panoramic place representation,'' in \emph{IEEE Conference on Computer Vision and Pattern Recognition}, 2012, pp. 2695--2702.

\bibitem{chang2018deepvp}
C.-K. Chang, J.~Zhao, and L.~Itti, ``Deepvp: Deep learning for vanishing point detection on 1 million street view images,'' in \emph{IEEE International Conference on Robotics and Automation}, 2018, pp. 4496--4503.

\bibitem{FishEyeRecNet}
X.~Yin, X.~Wang, J.~Yu, M.~Zhang, P.~Fua, and D.~Tao, ``Fisheyerecnet: A multi-context collaborative deep network for fisheye image rectification,'' in \emph{European Conference on Computer Vision}, 2018.

\bibitem{ADE20K}
B.~Zhou, H.~Zhao, X.~Puig, S.~Fidler, A.~Barriuso, and A.~Torralba, ``Scene parsing through ade20k dataset,'' in \emph{Proceedings of the IEEE Computer Vision and Pattern Recognition}, 2017, pp. 633--641.

\bibitem{Shi}
Y.~Shi, D.~Zhang, J.~Wen, X.~Tong, X.~Ying, and H.~Zha, ``Radial lens distortion correction by adding a weight layer with inverted foveal models to convolutional neural networks,'' in \emph{International Conference on Pattern Recognition}, 2018.

\bibitem{UprightNet}
W.~Xian, Z.~Li, M.~Fisher, J.~Eisenmann, E.~Shechtman, and N.~Snavely, ``Uprightnet: Geometry-aware camera orientation estimation from single images,'' in \emph{International Conference on Computer Vision}, 2019.

\bibitem{InteriorNet}
W.~Li, S.~Saeedi, J.~McCormac, R.~Clark, D.~Tzoumanikas, Q.~Ye, Y.~Huang, R.~Tang, and S.~Leutenegger, ``Interiornet: Mega-scale multi-sensor photo-realistic indoor scenes dataset,'' \emph{arXiv preprint arXiv:1809.00716}, 2018.

\bibitem{ScanNet}
A.~Dai, A.~X. Chang, M.~Savva, M.~Halber, T.~Funkhouser, and M.~Nie{\ss}ner, ``Scannet: Richly-annotated 3d reconstructions of indoor scenes,'' in \emph{Proceedings of the IEEE Conference on Computer Vision and Pattern Recognition}, 2017, pp. 5828--5839.

\bibitem{zhou2019neurvps}
Y.~Zhou, H.~Qi, J.~Huang, and Y.~Ma, ``Neurvps: Neural vanishing point scanning via conic convolution,'' \emph{Advances in Neural Information Processing Systems}, vol.~32, 2019.

\bibitem{SU3}
Y.~Zhou, H.~Qi, Y.~Zhai, Q.~Sun, Z.~Chen, L.-Y. Wei, and Y.~Ma, ``Learning to reconstruct 3d manhattan wireframes from a single image,'' in \emph{Proceedings of the IEEE/CVF International Conference on Computer Vision}, 2019, pp. 7698--7707.

\bibitem{Deep360Up}
R.~Jung, A.~S.~J. Lee, A.~Ashtari, and J.-C. Bazin, ``Deep360up: A deep learning-based approach for automatic vr image upright adjustment,'' in \emph{IEEE Conference on Virtual Reality and 3D User Interfaces}, 2019.

\bibitem{Log-cosh}
V.~Belagiannis, C.~Rupprecht, G.~Carneiro, and N.~Navab, ``Robust optimization for deep regression,'' in \emph{Proceedings of the IEEE International Conference on Computer Vision}, 2015, pp. 2830--2838.

\bibitem{Lopez}
M.~Lopez, R.~Mari, P.~Gargallo, Y.~Kuang, J.~Gonzalez-Jimenez, and G.~Haro, ``Deep single image camera calibration with radial distortion,'' in \emph{Proceedings of the IEEE/CVF Conference on Computer Vision and Pattern Recognition}, 2019.

\bibitem{Zhuang}
B.~Zhuang, Q.-H. Tran, G.~H. Lee, L.~F. Cheong, and M.~Chandraker, ``Degeneracy in self-calibration revisited and a deep learning solution for uncalibrated slam,'' in \emph{IEEE/RSJ International Conference on Intelligent Robots and Systems}, 2019, pp. 3766--3773.

\bibitem{KITTI}
A.~Geiger, P.~Lenz, and R.~Urtasun, ``Are we ready for autonomous driving? the kitti vision benchmark suite,'' in \emph{IEEE Conference on Computer Vision and Pattern Recognition}, 2012, pp. 3354--3361.

\bibitem{MS-COCO}
T.-Y. Lin, M.~Maire, S.~Belongie, J.~Hays, P.~Perona, D.~Ramanan, P.~Doll{\'a}r, and C.~L. Zitnick, ``Microsoft coco: Common objects in context,'' in \emph{European Conference on Computer Vision}, 2014, pp. 740--755.

\bibitem{STD}
K.~Liao, C.~Lin, Y.~Zhao, and M.~Gabbouj, ``Distortion rectification from static to dynamic: A distortion sequence construction perspective,'' \emph{IEEE Transactions on Circuits and Systems for Video Technology}, vol.~30, no.~11, pp. 3870--3882, 2020.

\bibitem{Places2}
B.~Zhou, A.~Lapedriza, A.~Khosla, A.~Oliva, and A.~Torralba, ``Places: A 10 million image database for scene recognition,'' \emph{IEEE Transactions on Pattern Analysis and Machine Intelligence}, vol.~40, no.~6, pp. 1452--1464, 2018.

\bibitem{RSC-Net}
B.~Zhuang, Q.-H. Tran, P.~Ji, L.-F. Cheong, and M.~Chandraker, ``Learning structure-and-motion-aware rolling shutter correction,'' in \emph{Proceedings of the IEEE/CVF Conference on Computer Vision and Pattern Recognition}, 2019.

\bibitem{Xue}
Z.~Xue, N.~Xue, G.-S. Xia, and W.~Shen, ``Learning to calibrate straight lines for fisheye image rectification,'' in \emph{Proceedings of the IEEE/CVF Conference on Computer Vision and Pattern Recognition}, 2019.

\bibitem{Wireframes}
K.~Huang, Y.~Wang, Z.~Zhou, T.~Ding, S.~Gao, and Y.~Ma, ``Learning to parse wireframes in images of man-made environments,'' in \emph{Proceedings of the IEEE Conference on Computer Vision and Pattern Recognition}, 2018, pp. 626--635.

\bibitem{SUNCG}
S.~Song, F.~Yu, A.~Zeng, A.~X. Chang, M.~Savva, and T.~Funkhouser, ``Semantic scene completion from a single depth image,'' in \emph{Proceedings of the IEEE Conference on Computer Vision and Pattern Recognition}, 2017, pp. 1746--1754.

\bibitem{BU-4DFE}
L.~Yin, X.~Sun, T.~Worm, and M.~Reale, ``A high-resolution 3d dynamic facial expression database, 2008,'' in \emph{IEEE International Conference on Automatic Face and Gesture Recognition, Amsterdam, The Netherlands}, vol. 126.

\bibitem{Lee}
J.~Lee, M.~Sung, H.~Lee, and J.~Kim, ``Neural geometric parser for single image camera calibration,'' in \emph{European Conference on Computer Vision}, 2020, pp. 541--557.

\bibitem{googleStreet}
\BIBentryALTinterwordspacing
 [Online]. Available: \url{https://developers.google.com/maps/}
\BIBentrySTDinterwordspacing

\bibitem{AUC}
O.~Barinova, V.~Lempitsky, E.~Tretiak, and P.~Kohli, ``Geometric image parsing in man-made environments,'' in \emph{European Conference on Computer Vision}, 2010, pp. 57--70.

\bibitem{Baradad}
M.~Baradad and A.~Torralba, ``Height and uprightness invariance for 3d prediction from a single view,'' in \emph{Proceedings of the IEEE/CVF Conference on Computer Vision and Pattern Recognition}, 2020.

\bibitem{NYU}
N.~Silberman, D.~Hoiem, P.~Kohli, and R.~Fergus, ``Indoor segmentation and support inference from rgbd images,'' in \emph{European Conference on Computer Vision}, 2012, pp. 746--760.

\bibitem{Zheng}
Q.~Zheng, J.~Chen, Z.~Lu, B.~Shi, X.~Jiang, K.-H. Yap, L.-Y. Duan, and A.~C. Kot, ``What does plate glass reveal about camera calibration?'' in \emph{Proceedings of the IEEE/CVF Conference on Computer Vision and Pattern Recognition}, 2020.

\bibitem{FocaLens}
\BIBentryALTinterwordspacing
 [Online]. Available: \url{https://figshare.com/articles/dataset/FocaLens/3399169/2}
\BIBentrySTDinterwordspacing

\bibitem{Davidson}
B.~Davidson, M.~S. Alvi, and J.~F. Henriques, ``360° camera alignment via segmentation,'' in \emph{European Conference on Computer Vision}, 2020, pp. 579--595.

\bibitem{DeepFEPE}
Y.-Y. Jau, R.~Zhu, H.~Su, and M.~Chandraker, ``Deep keypoint-based camera pose estimation with geometric constraints,'' in \emph{2020 IEEE/RSJ International Conference on Intelligent Robots and Systems (IROS)}, 2020, pp. 4950--4957.

\bibitem{Apolloscape}
X.~Huang, P.~Wang, X.~Cheng, D.~Zhou, Q.~Geng, and R.~Yang, ``The apolloscape open dataset for autonomous driving and its application,'' \emph{IEEE Transactions on Pattern Analysis and Machine Intelligence}, vol.~42, no.~10, pp. 2702--2719, 2019.

\bibitem{MisCaliDet}
A.~Cramariuc, A.~Petrov, R.~Suri, M.~Mittal, R.~Siegwart, and C.~Cadena, ``Learning camera miscalibration detection,'' in \emph{IEEE International Conference on Robotics and Automation}, 2020, pp. 4997--5003.

\bibitem{DeepPTZ}
C.~Zhang, F.~Rameau, J.~Kim, D.~M. Argaw, J.-C. Bazin, and I.~S. Kweon, ``Deepptz: Deep self-calibration for ptz cameras,'' in \emph{Proceedings of the IEEE/CVF Winter Conference on Applications of Computer Vision}, 2020.

\bibitem{Li}
Y.-H. Li, I.-C. Lo, and H.~H. Chen, ``Deep face rectification for 360° dual-fisheye cameras,'' \emph{IEEE Transactions on Image Processing}, vol.~30, pp. 264--276, 2021.

\bibitem{CelebA}
Y.~Guo, L.~Zhang, Y.~Hu, X.~He, and J.~Gao, ``Ms-celeb-1m: A dataset and benchmark for large-scale face recognition,'' in \emph{European Conference on Computer Vision}, 2016, pp. 87--102.

\bibitem{PSE-GAN}
Y.~Shi, X.~Tong, J.~Wen, H.~Zhao, X.~Ying, and H.~Zha, ``Position-aware and symmetry enhanced gan for radial distortion correction,'' in \emph{International Conference on Pattern Recognition}, 2021, pp. 1701--1708.

\bibitem{RDC-Net}
H.~Zhao, Y.~Shi, X.~Tong, X.~Ying, and H.~Zha, ``A simple yet effective pipeline for radial distortion correction,'' in \emph{IEEE International Conference on Image Processing}, 2020, pp. 878--882.

\bibitem{FE-GAN}
C.-H. Chao, P.-L. Hsu, H.-Y. Lee, and Y.-C.~F. Wang, ``Self-supervised deep learning for fisheye image rectification,'' in \emph{IEEE International Conference on Acoustics, Speech and Signal Processing}, 2020, pp. 2248--2252.

\bibitem{LSUN}
F.~Yu, A.~Seff, Y.~Zhang, S.~Song, T.~Funkhouser, and J.~Xiao, ``Lsun: Construction of a large-scale image dataset using deep learning with humans in the loop,'' \emph{arXiv preprint arXiv:1506.03365}, 2015.

\bibitem{RDCFace}
H.~Zhao, X.~Ying, Y.~Shi, X.~Tong, J.~Wen, and H.~Zha, ``Rdcface: Radial distortion correction for face recognition,'' in \emph{Proceedings of the IEEE/CVF Conference on Computer Vision and Pattern Recognition}, 2020.

\bibitem{IMDB-Face}
F.~Wang, L.~Chen, C.~Li, S.~Huang, Y.~Chen, C.~Qian, and C.~C. Loy, ``The devil of face recognition is in the noise,'' in \emph{European Conference on Computer Vision}, 2018, pp. 765--780.

\bibitem{LaRecNet}
Z.-C. Xue, N.~Xue, and G.-S. Xia, ``Fisheye distortion rectification from deep straight lines,'' \emph{arXiv preprint arXiv:2003.11386}, 2020.

\bibitem{StereoCaliNet}
Y.~Gil, S.~Elmalem, H.~Haim, E.~Marom, and R.~Giryes, ``Online training of stereo self-calibration using monocular depth estimation,'' \emph{IEEE Transactions on Computational Imaging}, vol.~7, pp. 812--823, 2021.

\bibitem{TAUAgent}
\BIBentryALTinterwordspacing
 [Online]. Available: \url{http://www.cs.toronto.edu/~harel/TAUAgent/download.html}
\BIBentrySTDinterwordspacing

\bibitem{CTRL-C}
J.~Lee, H.~Go, H.~Lee, S.~Cho, M.~Sung, and J.~Kim, ``Ctrl-c: Camera calibration transformer with line-classification,'' in \emph{International Conference on Computer Vision}, 2021, pp. 16\,228--16\,237.

\bibitem{SA-MobileNet}
S.~Garg, D.~P. Mohanty, S.~P. Thota, and S.~Moharana, ``A simple approach to image tilt correction with self-attention mobilenet for smartphones,'' \emph{British Machine Vision Conference}, 2021.

\bibitem{DirectionNet}
K.~Chen, N.~Snavely, and A.~Makadia, ``Wide-baseline relative camera pose estimation with directional learning,'' in \emph{Proceedings of the IEEE/CVF Conference on Computer Vision and Pattern Recognition}, 2021, pp. 3258--3268.

\bibitem{Matterport3D}
A.~Chang, A.~Dai, T.~Funkhouser, M.~Halber, M.~Niessner, M.~Savva, S.~Song, A.~Zeng, and Y.~Zhang, ``Matterport3d: Learning from rgb-d data in indoor environments,'' \emph{International Conference on 3D Vision}, 2017.

\bibitem{Wakai}
N.~Wakai and T.~Yamashita, ``Deep single fisheye image camera calibration for over 180-degree projection of field of view,'' in \emph{International Conference on Computer Vision Workshops}, 2021, pp. 1174--1183.

\bibitem{StreetLearn}
P.~Mirowski, A.~Banki-Horvath, K.~Anderson, D.~Teplyashin, K.~M. Hermann, M.~Malinowski, M.~K. Grimes, K.~Simonyan, K.~Kavukcuoglu, A.~Zisserman \emph{et~al.}, ``The streetlearn environment and dataset,'' \emph{arXiv preprint arXiv:1903.01292}, 2019.

\bibitem{OrdianlDistortion}
K.~Liao, C.~Lin, and Y.~Zhao, ``A deep ordinal distortion estimation approach for distortion rectification,'' \emph{IEEE Transactions on Image Processing}, vol.~30, pp. 3362--3375, 2021.

\bibitem{PolarRecNet}
K.~Zhao, C.~Lin, K.~Liao, S.~Yang, and Y.~Zhao, ``Revisiting radial distortion rectification in polar-coordinates: A new and efficient learning perspective,'' \emph{IEEE Transactions on Circuits and Systems for Video Technology}, vol.~32, no.~6, pp. 3552--3560, 2021.

\bibitem{LMS}
A.~Eichenseer and A.~Kaup, ``A data set providing synthetic and real-world fisheye video sequences,'' in \emph{IEEE International Conference on Acoustics, Speech and Signal Processing}, 2016, pp. 1541--1545.

\bibitem{PCN}
S.~Yang, C.~Lin, K.~Liao, C.~Zhang, and Y.~Zhao, ``Progressively complementary network for fisheye image rectification using appearance flow,'' in \emph{Proceedings of the IEEE/CVF Conference on Computer Vision and Pattern Recognition}, 2021, pp. 6348--6357.

\bibitem{JCD}
Z.~Zhong, Y.~Zheng, and I.~Sato, ``Towards rolling shutter correction and deblurring in dynamic scenes,'' in \emph{Proceedings of the IEEE/CVF Conference on Computer Vision and Pattern Recognition}, 2021, pp. 9219--9228.

\bibitem{Charbonnier}
W.-S. Lai, J.-B. Huang, N.~Ahuja, and M.-H. Yang, ``Fast and accurate image super-resolution with deep laplacian pyramid networks,'' \emph{IEEE Transactions on Pattern Analysis and Machine Intelligence}, vol.~41, no.~11, pp. 2599--2613, 2018.

\bibitem{fan2021inverting}
B.~Fan and Y.~Dai, ``Inverting a rolling shutter camera: bring rolling shutter images to high framerate global shutter video,'' in \emph{Proceedings of the IEEE/CVF International Conference on Computer Vision}, 2021, pp. 4228--4237.

\bibitem{SUNet}
B.~Fan, Y.~Dai, and M.~He, ``Sunet: symmetric undistortion network for rolling shutter correction,'' in \emph{Proceedings of the IEEE/CVF International Conference on Computer Vision}, 2021, pp. 4541--4550.

\bibitem{EuRoC}
M.~Burri, J.~Nikolic, P.~Gohl, T.~Schneider, J.~Rehder, S.~Omari, M.~W. Achtelik, and R.~Siegwart, ``The euroc micro aerial vehicle datasets,'' \emph{The International Journal of Robotics Research}, vol.~35, no.~10, pp. 1157--1163, 2016.

\bibitem{OmniCam}
M.~Sch{\"o}nbein, T.~Strau{\ss}, and A.~Geiger, ``Calibrating and centering quasi-central catadioptric cameras,'' in \emph{IEEE International Conference on Robotics and Automation}, 2014, pp. 4443--4450.

\bibitem{CPL}
T.~H. Butt and M.~Taj, ``Camera calibration through camera projection loss,'' in \emph{IEEE International Conference on Acoustics, Speech and Signal Processing}, 2022, pp. 2649--2653.

\bibitem{CARLA}
A.~Dosovitskiy, G.~Ros, F.~Codevilla, A.~Lopez, and V.~Koltun, ``Carla: An open urban driving simulator,'' in \emph{Conference on Robot Learning}, 2017, pp. 1--16.

\bibitem{CyclistDetection}
X.~Li, F.~Flohr, Y.~Yang, H.~Xiong, M.~Braun, S.~Pan, K.~Li, and D.~M. Gavrila, ``A new benchmark for vision-based cyclist detection,'' in \emph{IEEE Intelligent Vehicles Symposium}, 2016, pp. 1028--1033.

\bibitem{Do}
T.~Do, O.~Miksik, J.~DeGol, H.~S. Park, and S.~N. Sinha, ``Learning to detect scene landmarks for camera localization,'' in \emph{Proceedings of the IEEE/CVF Conference on Computer Vision and Pattern Recognition}, 2022, pp. 11\,132--11\,142.

\bibitem{RobustAngular}
T.~Do, K.~Vuong, S.~I. Roumeliotis, and H.~S. Park, ``Surface normal estimation of tilted images via spatial rectifier,'' in \emph{European Conference on Computer Vision}, 2020, pp. 265--280.

\bibitem{7-SCENES}
J.~Shotton, B.~Glocker, C.~Zach, S.~Izadi, A.~Criminisi, and A.~Fitzgibbon, ``Scene coordinate regression forests for camera relocalization in rgb-d images,'' in \emph{Proceedings of the IEEE Conference on Computer Vision and Pattern Recognition}, 2013, pp. 2930--2937.

\bibitem{DiffPoseNet}
C.~M. Parameshwara, G.~Hari, C.~Ferm{\"u}ller, N.~J. Sanket, and Y.~Aloimonos, ``Diffposenet: Direct differentiable camera pose estimation,'' in \emph{Proceedings of the IEEE/CVF Conference on Computer Vision and Pattern Recognition}, 2022, pp. 6845--6854.

\bibitem{TartanAir}
W.~Wang, D.~Zhu, X.~Wang, Y.~Hu, Y.~Qiu, C.~Wang, Y.~Hu, A.~Kapoor, and S.~Scherer, ``Tartanair: A dataset to push the limits of visual slam,'' in \emph{IEEE/RSJ International Conference on Intelligent Robots and Systems}, 2020, pp. 4909--4916.

\bibitem{TUM_RGB-D}
J.~Sturm, N.~Engelhard, F.~Endres, W.~Burgard, and D.~Cremers, ``A benchmark for the evaluation of rgb-d slam systems,'' in \emph{IEEE/RSJ International Conference on Intelligent Robots and Systems}, 2012, pp. 573--580.

\bibitem{AEE}
B.~J. Pijnacker~Hordijk, K.~Y. Scheper, and G.~C. De~Croon, ``Vertical landing for micro air vehicles using event-based optical flow,'' \emph{Journal of Field Robotics}, vol.~35, no.~1, pp. 69--90, 2018.

\bibitem{SceneSqueezer}
L.~Yang, R.~Shrestha, W.~Li, S.~Liu, G.~Zhang, Z.~Cui, and P.~Tan, ``Scenesqueezer: Learning to compress scene for camera relocalization,'' in \emph{Proceedings of the IEEE/CVF Conference on Computer Vision and Pattern Recognition}, 2022, pp. 8259--8268.

\bibitem{RobotCar}
W.~Maddern, G.~Pascoe, C.~Linegar, and P.~Newman, ``1 year, 1000 km: The oxford robotcar dataset,'' \emph{The International Journal of Robotics Research}, vol.~36, no.~1, pp. 3--15, 2017.

\bibitem{FocalPose}
G.~Ponimatkin, Y.~Labb{\'e}, B.~Russell, M.~Aubry, and J.~Sivic, ``Focal length and object pose estimation via render and compare,'' in \emph{Proceedings of the IEEE/CVF Conference on Computer Vision and Pattern Recognition}, 2022, pp. 3825--3834.

\bibitem{Pix3D}
X.~Sun, J.~Wu, X.~Zhang, Z.~Zhang, C.~Zhang, T.~Xue, J.~B. Tenenbaum, and W.~T. Freeman, ``Pix3d: Dataset and methods for single-image 3d shape modeling,'' in \emph{Proceedings of the IEEE Conference on Computer Vision and Pattern Recognition}, 2018, pp. 2974--2983.

\bibitem{StanfordCars}
Y.~Wang, X.~Tan, Y.~Yang, X.~Liu, E.~Ding, F.~Zhou, and L.~S. Davis, ``3d pose estimation for fine-grained object categories,'' in \emph{European Conference on Computer Vision Workshops}, 2018.

\bibitem{AW-RSC}
M.~Cao, Z.~Zhong, J.~Wang, Y.~Zheng, and Y.~Yang, ``Learning adaptive warping for real-world rolling shutter correction,'' in \emph{Proceedings of the IEEE/CVF Conference on Computer Vision and Pattern Recognition}, 2022, pp. 17\,785--17\,793.

\bibitem{zhang2022learning}
``Learning-based framework for camera calibration with distortion correction and high precision feature detection,'' \emph{IEEE Robotics and Automation Letters}, vol.~7, no.~4, pp. 10\,470--10\,477, 2022.

\bibitem{GenCaliNet}
N.~Wakai, S.~Sato, Y.~Ishii, and T.~Yamashita, ``Rethinking generic camera models for deep single image camera calibration to recover rotation and fisheye distortion,'' in \emph{{European Conference on Computer Vision}}, vol. 13678, 2022, pp. 679--698.

\bibitem{SP360}
S.-H. Chang, C.-Y. Chiu, C.-S. Chang, K.-W. Chen, C.-Y. Yao, R.-R. Lee, and H.-K. Chu, ``Generating 360 outdoor panorama dataset with reliable sun position estimation,'' in \emph{SIGGRAPH Asia}, 2018, pp. 1--2.

\bibitem{IFED}
Z.~Zhong, M.~Cao, X.~Sun, Z.~Wu, Z.~Zhou, Y.~Zheng, S.~Lin, and I.~Sato, ``Bringing rolling shutter images alive with dual reversed distortion,'' in \emph{European Conference on Computer Vision}, 2022, pp. 233--249.

\bibitem{PM-Calib}
Y.~Hold-Geoffroy, D.~Pich{\'e}-Meunier, K.~Sunkavalli, J.-C. Bazin, F.~Rameau, and J.-F. Lalonde, ``A perceptual measure for deep single image camera and lens calibration,'' \emph{IEEE Transactions on Pattern Analysis and Machine Intelligence}, vol.~45, no.~9, pp. 10\,603--10\,614, 2023.

\bibitem{360Cities}
\BIBentryALTinterwordspacing
 [Online]. Available: \url{https://www.360cities.net/}
\BIBentrySTDinterwordspacing

\bibitem{PerspectiveField}
L.~Jin, J.~Zhang, Y.~Hold-Geoffroy, O.~Wang, K.~Blackburn-Matzen, M.~Sticha, and D.~F. Fouhey, ``Perspective fields for single image camera calibration,'' in \emph{Proceedings of the IEEE/CVF Conference on Computer Vision and Pattern Recognition}, 2023, pp. 17\,307--17\,316.

\bibitem{Stanford2D3D}
I.~Armeni, ``Joint 2d-3d semantic data for indoor scene understanding,'' \emph{arXiv preprint arXiv:1702.01105}, 2017.

\bibitem{Orienternet}
P.-E. Sarlin, D.~DeTone, T.-Y. Yang, A.~Avetisyan, J.~Straub, T.~Malisiewicz, S.~R. Bulo, R.~Newcombe, P.~Kontschieder, and V.~Balntas, ``Orienternet: Visual localization in 2d public maps with neural matching,'' in \emph{Proceedings of the IEEE/CVF Conference on Computer Vision and Pattern Recognition}, 2023, pp. 21\,632--21\,642.

\bibitem{Neumap}
S.~Tang, S.~Tang, A.~Tagliasacchi, P.~Tan, and Y.~Furukawa, ``Neumap: Neural coordinate mapping by auto-transdecoder for camera localization,'' in \emph{Proceedings of the IEEE/CVF Conference on Computer Vision and Pattern Recognition}, 2023, pp. 929--939.

\bibitem{SESC}
T.~Kanai, I.~Vasiljevic, V.~Guizilini, A.~Gaidon, and R.~Ambrus, ``Robust self-supervised extrinsic self-calibration,'' in \emph{IEEE/RSJ International Conference on Intelligent Robots and Systems}, 2023, pp. 1932--1939.

\bibitem{DDAD}
V.~Guizilini, R.~Ambrus, S.~Pillai, A.~Raventos, and A.~Gaidon, ``3d packing for self-supervised monocular depth estimation,'' in \emph{Proceedings of the IEEE/CVF conference on computer vision and pattern recognition}, 2020, pp. 2485--2494.

\bibitem{CROSSFIRE}
A.~Moreau, N.~Piasco, M.~Bennehar, D.~Tsishkou, B.~Stanciulescu, and A.~de~La~Fortelle, ``Crossfire: Camera relocalization on self-supervised features from an implicit representation,'' in \emph{Proceedings of the IEEE/CVF International Conference on Computer Vision}, 2023, pp. 252--262.

\bibitem{WildCamera}
S.~Zhu, A.~Kumar, M.~Hu, and X.~Liu, ``Tame a wild camera: In-the-wild monocular camera calibration,'' \emph{Advances in Neural Information Processing Systems}, vol.~36, 2023.

\bibitem{MegaDepth}
Z.~Li and N.~Snavely, ``Megadepth: Learning single-view depth prediction from internet photos,'' in \emph{Proceedings of the IEEE conference on computer vision and pattern recognition}, 2018, pp. 2041--2050.

\bibitem{DroidCalib}
A.~Hagemann, M.~Knorr, and C.~Stiller, ``Deep geometry-aware camera self-calibration from video,'' in \emph{Proceedings of the IEEE/CVF International Conference on Computer Vision}, 2023, pp. 3438--3448.

\bibitem{NeuralLens}
W.~Xian, A.~Bo{\v{z}}i{\v{c}}, N.~Snavely, and C.~Lassner, ``Neural lens modeling,'' in \emph{Proceedings of the IEEE/CVF Conference on Computer Vision and Pattern Recognition}, 2023, pp. 8435--8445.

\bibitem{DDA}
S.~Yang, C.~Lin, K.~Liao, and Y.~Zhao, ``Innovating real fisheye image correction with dual diffusion architecture,'' in \emph{Proceedings of the IEEE/CVF International Conference on Computer Vision}, 2023, pp. 12\,699--12\,708.

\bibitem{yogamani2019woodscape}
S.~Yogamani, C.~Hughes, J.~Horgan, G.~Sistu, P.~Varley, D.~O'Dea, M.~Uric{\'a}r, S.~Milz, M.~Simon, K.~Amende \emph{et~al.}, ``Woodscape: A multi-task, multi-camera fisheye dataset for autonomous driving,'' in \emph{Proceedings of the IEEE/CVF International Conference on Computer Vision}, 2019, pp. 9308--9318.

\bibitem{CACM-Net}
Q.~Zhang, H.~Li, and Q.~Wang, ``Wide-angle rectification via content-aware conformal mapping,'' in \emph{Proceedings of the IEEE/CVF Conference on Computer Vision and Pattern Recognition}, 2023, pp. 17\,357--17\,365.

\bibitem{RDTR}
W.~Wang, H.~Feng, W.~Zhou, Z.~Liao, and H.~Li, ``Model-aware pre-training for radial distortion rectification,'' \emph{IEEE Transactions on Image Processing}, 2023.

\bibitem{DaFIR}
Z.~Liao, W.~Zhou, and H.~Li, ``Dafir: Distortion-aware representation learning for fisheye image rectification,'' \emph{IEEE Transactions on Circuits and Systems for Video Technology}, 2023.

\bibitem{SimFIR}
H.~Feng, W.~Wang, J.~Deng, W.~Zhou, L.~Li, and H.~Li, ``Simfir: A simple framework for fisheye image rectification with self-supervised representation learning,'' in \emph{Proceedings of the IEEE/CVF International Conference on Computer Vision}, 2023, pp. 12\,418--12\,427.

\bibitem{Recrecnet}
K.~Liao, L.~Nie, C.~Lin, Z.~Zheng, and Y.~Zhao, ``Recrecnet: Rectangling rectified wide-angle images by thin-plate spline model and dof-based curriculum learning,'' in \emph{Proceedings of the IEEE/CVF International Conference on Computer Vision}, 2023, pp. 10\,800--10\,809.

\bibitem{SDP-Net}
S.~Yang, C.~Lin, K.~Liao, and Y.~Zhao, ``Spatiotemporal deformation perception for fisheye video rectification,'' in \emph{Proceedings of the AAAI Conference on Artificial Intelligence}, vol.~37, no.~3, 2023, pp. 3181--3189.

\bibitem{perazzi2016benchmark}
F.~Perazzi, J.~Pont-Tuset, B.~McWilliams, L.~Van~Gool, M.~Gross, and A.~Sorkine-Hornung, ``A benchmark dataset and evaluation methodology for video object segmentation,'' in \emph{Proceedings of the IEEE conference on computer vision and pattern recognition}, 2016, pp. 724--732.

\bibitem{xu2018youtube}
N.~Xu, L.~Yang, Y.~Fan, D.~Yue, Y.~Liang, J.~Yang, and T.~Huang, ``Youtube-vos: A large-scale video object segmentation benchmark,'' \emph{arXiv preprint arXiv:1809.03327}, 2018.

\bibitem{Darswin}
A.~Athwale, A.~Afrasiyabi, J.~Lag{\"u}e, I.~Shili, O.~Ahmad, and J.-F. Lalonde, ``Darswin: Distortion aware radial swin transformer,'' in \emph{Proceedings of the IEEE/CVF International Conference on Computer Vision}, 2023, pp. 5929--5938.

\bibitem{REG-Net}
G.~Lin, J.~Han, M.~Cao, Z.~Zhong, and Y.~Zheng, ``Event-guided frame interpolation and dynamic range expansion of single rolling shutter image,'' in \emph{Proceedings of the 31st ACM International Conference on Multimedia}, 2023, pp. 3078--3088.

\bibitem{SelfDRSC}
W.~Shang, D.~Ren, C.~Feng, X.~Wang, L.~Lei, and W.~Zuo, ``Self-supervised learning to bring dual reversed rolling shutter images alive,'' in \emph{Proceedings of the IEEE/CVF International Conference on Computer Vision}, 2023, pp. 13\,086--13\,094.

\bibitem{SSL-RSC}
Y.~Lu, G.~Liang, and L.~Wang, ``Self-supervised learning of event-guided video frame interpolation for rolling shutter frames,'' \emph{arXiv preprint arXiv:2306.15507}, 2023.

\bibitem{EvShutter}
J.~Erbach, S.~Tulyakov, P.~Vitoria, A.~Bochicchio, and Y.~Li, ``Evshutter: Transforming events for unconstrained rolling shutter correction,'' in \emph{Proceedings of the IEEE/CVF Conference on Computer Vision and Pattern Recognition}, 2023, pp. 13\,904--13\,913.

\bibitem{SelfUnroll}
Y.~Wang, X.~Zhang, M.~Lin, L.~Yu, B.~Shi, W.~Yang, and G.-S. Xia, ``Self-supervised scene dynamic recovery from rolling shutter images and events,'' \emph{arXiv preprint arXiv:2304.06930}, 2023.

\bibitem{PatchNet}
A.~Gupta, S.~K. Singh, and A.~K. Roy-Chowdhury, ``Joint video rolling shutter correction and super-resolution,'' in \emph{Proceedings of the IEEE/CVF Winter Conference on Applications of Computer Vision}, 2023, pp. 4946--4955.

\bibitem{JAMNet}
B.~Fan, Y.~Mao, Y.~Dai, Z.~Wan, and Q.~Liu, ``Joint appearance and motion learning for efficient rolling shutter correction,'' in \emph{Proceedings of the IEEE/CVF Conference on Computer Vision and Pattern Recognition}, 2023, pp. 5671--5681.

\bibitem{QRSC}
D.~Qu, Y.~Lao, Z.~Wang, D.~Wang, B.~Zhao, and X.~Li, ``Towards nonlinear-motion-aware and occlusion-robust rolling shutter correction,'' in \emph{Proceedings of the IEEE/CVF International Conference on Computer Vision}, 2023, pp. 10\,680--10\,688.

\bibitem{Deep_HM}
W.~Yan, R.~T. Tan, B.~Zeng, and S.~Liu, ``Deep homography mixture for single image rolling shutter correction,'' in \emph{Proceedings of the IEEE/CVF International Conference on Computer Vision}, 2023, pp. 9868--9877.

\bibitem{caesar2020nuscenes}
H.~Caesar, V.~Bankiti, A.~H. Lang, S.~Vora, V.~E. Liong, Q.~Xu, A.~Krishnan, Y.~Pan, G.~Baldan, and O.~Beijbom, ``nuscenes: A multimodal dataset for autonomous driving,'' in \emph{Proceedings of the IEEE/CVF conference on computer vision and pattern recognition}, 2020, pp. 11\,621--11\,631.

\bibitem{SUN3D}
J.~Xiao, A.~Owens, and A.~Torralba, ``Sun3d: A database of big spaces reconstructed using sfm and object labels,'' in \emph{Proceedings of the IEEE International Conference on Computer Vision}, 2013, pp. 1625--1632.

\bibitem{CityScapes}
M.~Cordts, M.~Omran, S.~Ramos, T.~Rehfeld, M.~Enzweiler, R.~Benenson, U.~Franke, S.~Roth, and B.~Schiele, ``The cityscapes dataset for semantic urban scene understanding,'' in \emph{Proceedings of the IEEE Conference on Computer Vision and Pattern Recognition}, 2016, pp. 3213--3223.

\bibitem{ExtremeRotation}
H.~Bezalel, D.~Ankri, R.~Cai, and H.~Averbuch-Elor, ``Extreme rotation estimation in the wild,'' \emph{arXiv preprint arXiv:2411.07096}, 2024.

\bibitem{GAT-Calib}
G.~D'Amicantonio, E.~Bondarev \emph{et~al.}, ``Automated camera calibration via homography estimation with gnns,'' in \emph{Proceedings of the IEEE/CVF Winter Conference on Applications of Computer Vision}, 2024, pp. 5876--5883.

\bibitem{homayounfar2017sports}
N.~Homayounfar, S.~Fidler, and R.~Urtasun, ``Sports field localization via deep structured models,'' in \emph{Proceedings of the IEEE Conference on Computer Vision and Pattern Recognition}, 2017, pp. 5212--5220.

\bibitem{SOFI}
S.~Janampa and M.~Pattichis, ``{SOFI: Multi-Scale Deformable Transformer for Camera Calibration with Enhanced Line Queries},'' in \emph{British Machine Vision Conference}, 2024.

\bibitem{PWT-Calib}
K.~Vuong, R.~Tamburo, and S.~G. Narasimhan, ``Toward planet-wide traffic camera calibration,'' in \emph{Proceedings of the IEEE/CVF Winter Conference on Applications of Computer Vision}, 2024, pp. 8553--8562.

\bibitem{NeFeS}
S.~Chen, Y.~Bhalgat, X.~Li, J.-W. Bian, K.~Li, Z.~Wang, and V.~A. Prisacariu, ``Neural refinement for absolute pose regression with feature synthesis,'' in \emph{Proceedings of the IEEE/CVF Conference on Computer Vision and Pattern Recognition}, 2024, pp. 20\,987--20\,996.

\bibitem{U-ARE-ME}
A.~Patwardhan, C.~Rhodes, G.~Bae, and A.~J. Davison, ``U-are-me: Uncertainty-aware rotation estimation in manhattan environments,'' \emph{arXiv preprint arXiv:2403.15583}, 2024.

\bibitem{ICL-NUIM}
A.~Handa, T.~Whelan, J.~McDonald, and A.~J. Davison, ``A benchmark for rgb-d visual odometry, 3d reconstruction and slam,'' in \emph{IEEE International Conference on Robotics and Automation}, 2014, pp. 1524--1531.

\bibitem{FlowMap}
C.~Smith, D.~Charatan, A.~Tewari, and V.~Sitzmann, ``Flowmap: High-quality camera poses, intrinsics, and depth via gradient descent,'' \emph{arXiv preprint arXiv:2404.15259}, 2024.

\bibitem{Barron_2021_ICCV}
J.~T. Barron, B.~Mildenhall, M.~Tancik, P.~Hedman, R.~Martin-Brualla, and P.~P. Srinivasan, ``Mip-nerf: A multiscale representation for anti-aliasing neural radiance fields,'' in \emph{IEEE/CVF International Conference on Computer Vision}, 2021, pp. 5855--5864.

\bibitem{TAT}
A.~Knapitsch, J.~Park, Q.-Y. Zhou, and V.~Koltun, ``Tanks and temples: Benchmarking large-scale scene reconstruction,'' \emph{ACM Transactions on Graphics}, vol.~36, no.~4, pp. 1--13, 2017.

\bibitem{LLFF}
B.~Mildenhall, P.~P. Srinivasan, R.~Ortiz-Cayon, N.~K. Kalantari, R.~Ramamoorthi, R.~Ng, and A.~Kar, ``Local light field fusion: Practical view synthesis with prescriptive sampling guidelines,'' \emph{ACM Transactions on Graphics}, vol.~38, no.~4, pp. 1--14, 2019.

\bibitem{CO3D}
J.~Reizenstein, R.~Shapovalov, P.~Henzler, L.~Sbordone, P.~Labatut, and D.~Novotny, ``Common objects in 3d: Large-scale learning and evaluation of real-life 3d category reconstruction,'' in \emph{Proceedings of the IEEE/CVF International Conference on Computer Vision}, 2021, pp. 10\,901--10\,911.

\bibitem{MSCC}
X.~Song, H.~Kang, A.~Moteki, G.~Suzuki, Y.~Kobayashi, and Z.~Tan, ``Mscc: Multi-scale transformers for camera calibration,'' in \emph{Proceedings of the IEEE/CVF Winter Conference on Applications of Computer Vision}, 2024, pp. 3262--3271.

\bibitem{HC-Net}
X.~Wang, R.~Xu, Z.~Cui, Z.~Wan, and Y.~Zhang, ``Fine-grained cross-view geo-localization using a correlation-aware homography estimator,'' \emph{Advances in Neural Information Processing Systems}, vol.~36, 2024.

\bibitem{VIGOR}
S.~Zhu, T.~Yang, and C.~Chen, ``Vigor: Cross-view image geo-localization beyond one-to-one retrieval,'' in \emph{Proceedings of the IEEE/CVF Conference on Computer Vision and Pattern Recognition}, 2021, pp. 3640--3649.

\bibitem{Hypersim}
M.~Roberts, J.~Ramapuram, A.~Ranjan, A.~Kumar, M.~A. Bautista, N.~Paczan, R.~Webb, and J.~M. Susskind, ``Hypersim: A photorealistic synthetic dataset for holistic indoor scene understanding,'' in \emph{Proceedings of the IEEE/CVF International Conference on Computer Vision}, 2021, pp. 10\,912--10\,922.

\bibitem{ADPs}
N.~Wakai, S.~Sato, Y.~Ishii, and T.~Yamashita, ``Deep single image camera calibration by heatmap regression to recover fisheye images under manhattan world assumption,'' in \emph{Proceedings of the IEEE/CVF Conference on Computer Vision and Pattern Recognition}, 2024, pp. 11\,884--11\,894.

\bibitem{CDM}
J.~Zhao, S.~Wei, Y.~Chang, T.~Ruan, and Y.~Zhao, ``Model-free rectification via cascaded distortion model and enhanced backward flow network,'' \emph{IEEE Transactions on Circuits and Systems for Video Technology}, 2024.

\bibitem{QueryCDR}
P.~Guo, C.~Liu, X.~Hou, and X.~Qian, ``Querycdr: Query-based controllable distortion rectification network for fisheye images,'' 2024.

\bibitem{VACR}
A.~P. Dal~Cin, F.~Azzoni, G.~Boracchi, and L.~Magri, ``Revisiting calibration of wide-angle radially symmetric cameras,'' in \emph{European Conference on Computer Vision}, 2024, pp. 214--230.

\bibitem{liao2022kitti}
Y.~Liao, J.~Xie, and A.~Geiger, ``Kitti-360: A novel dataset and benchmarks for urban scene understanding in 2d and 3d,'' \emph{IEEE Transactions on Pattern Analysis and Machine Intelligence}, 2022.

\bibitem{DualPriorsCorrection}
L.~Yao, C.~Chen, X.~Li, Z.~Yan, and W.~Zuo, ``Combining generative and geometry priors for wide-angle portrait correction,'' in \emph{European Conference on Computer Vision}, 2024, pp. 395--411.

\bibitem{Disco}
Z.~Wang, Y.-L. Liu, J.-B. Huang, S.~Satoh, S.~Ma, G.~Krishnan, and J.~Wang, ``Disco: Portrait distortion correction with perspective-aware 3d gans,'' \emph{International Journal of Computer Vision}, pp. 1--18, 2024.

\bibitem{fried2016perspective}
O.~Fried, E.~Shechtman, D.~B. Goldman, and A.~Finkelstein, ``Perspective-aware manipulation of portrait photos,'' \emph{ACM Transactions on Graphics}, vol.~35, no.~4, pp. 1--10, 2016.

\bibitem{MOWA}
K.~Liao, Z.~Yue, Z.~Wu, and C.~C. Loy, ``Mowa: Multiple-in-one image warping model,'' \emph{arXiv preprint arXiv:2404.10716}, 2024.

\bibitem{StitchRect}
L.~Nie, C.~Lin, K.~Liao, S.~Liu, and Y.~Zhao, ``Deep rectangling for image stitching: A learning baseline,'' in \emph{Proceedings of the IEEE/CVF Conference on Computer Vision and Pattern Recognition}, 2022, pp. 5740--5748.

\bibitem{RotationCorr}
L.~\vspace{0mm}Nie, C.~Lin, K.~Liao, S.~Liu, and Y.~Zhao, ``Deep rotation correction without angle prior,'' \emph{IEEE Transactions on Image Processing}, vol.~32, pp. 2879--2888, 2023.

\bibitem{LBCNet}
B.~Fan, Y.~Dai, and H.~Li, ``Learning bilateral cost volume for rolling shutter temporal super-resolution,'' \emph{IEEE Transactions on Pattern Analysis and Machine Intelligence}, 2024.

\bibitem{TACA-Net}
H.~Huang, X.~Jia, X.~Zhang, S.~Li, and H.~Lu, ``Event-guided rolling shutter correction with time-aware cross-attentions,'' in \emph{Proceedings of the ACM International Conference on Multimedia}, 2024, pp. 5230--5239.

\bibitem{UniINR}
Y.~Lu, G.~Liang, Y.~Wang, L.~Wang, and H.~Xiong, ``Uniinr: Event-guided unified rolling shutter correction, deblurring, and interpolation,'' in \emph{European Conference on Computer Vision}, 2024, pp. 1--20.

\bibitem{DFRSC}
M.~Cao, S.~Yang, Y.~Yang, and Y.~Zheng, ``Rolling shutter correction with intermediate distortion flow estimation,'' in \emph{Proceedings of the IEEE/CVF Conference on Computer Vision and Pattern Recognition}, 2024, pp. 25\,338--25\,347.

\bibitem{CLKN}
C.-H. Chang, C.-N. Chou, and E.~Y. Chang, ``Clkn: Cascaded lucas-kanade networks for image alignment,'' in \emph{Proceedings of the IEEE Conference on Computer Vision and Pattern Recognition}, 2017.

\bibitem{HierarchicalNet}
F.~Erlik~Nowruzi, R.~Laganiere, and N.~Japkowicz, ``Homography estimation from image pairs with hierarchical convolutional networks,'' in \emph{International Conference on Computer Vision Workshops}, 2017.

\bibitem{DeepFM}
R.~Ranftl and V.~Koltun, ``Deep fundamental matrix estimation,'' in \emph{European Conference on Computer Vision}, 2018.

\bibitem{TT}
A.~Knapitsch, J.~Park, Q.-Y. Zhou, and V.~Koltun, ``Tanks and temples: Benchmarking large-scale scene reconstruction,'' \emph{ACM Transactions on Graphics}, vol.~36, no.~4, pp. 1--13, 2017.

\bibitem{Poursaeed}
O.~Poursaeed, G.~Yang, A.~Prakash, Q.~Fang, H.~Jiang, B.~Hariharan, and S.~Belongie, ``Deep fundamental matrix estimation without correspondences,'' in \emph{European Conference on Computer Vision Workshops}, 2018.

\bibitem{PFNet}
R.~Zeng, S.~Denman, S.~Sridharan, and C.~Fookes, ``Rethinking planar homography estimation using perspective fields,'' in \emph{Asian Conference on Computer Vision}, 2018, pp. 571--586.

\bibitem{iyer2018calibnet}
G.~Iyer, R.~K. Ram, J.~K. Murthy, and K.~M. Krishna, ``Calibnet: Geometrically supervised extrinsic calibration using 3d spatial transformer networks,'' in \emph{IEEE/RSJ International Conference on Intelligent Robots and Systems}, 2018, pp. 1110--1117.

\bibitem{Abbas}
S.~Ammar~Abbas and A.~Zisserman, ``A geometric approach to obtain a bird's eye view from an image,'' in \emph{International Conference on Computer Vision Workshops}, 2019.

\bibitem{Sha}
L.~Sha, J.~Hobbs, P.~Felsen, X.~Wei, P.~Lucey, and S.~Ganguly, ``End-to-end camera calibration for broadcast videos,'' in \emph{Proceedings of the IEEE/CVF Conference on Computer Vision and Pattern Recognition}, 2020.

\bibitem{MHN}
H.~Le, F.~Liu, S.~Zhang, and A.~Agarwala, ``Deep homography estimation for dynamic scenes,'' in \emph{Proceedings of the IEEE/CVF Conference on Computer Vision and Pattern Recognition}, 2020.

\bibitem{SRHEN}
Y.~Li, W.~Pei, and Z.~He, ``Srhen: stepwise-refining homography estimation network via parsing geometric correspondences in deep latent space,'' in \emph{Proceedings of the ACM International Conference on Multimedia}, 2020, pp. 3063--3071.

\bibitem{yuan2020rggnet}
K.~Yuan, Z.~Guo, and Z.~J. Wang, ``Rggnet: Tolerance aware lidar-camera online calibration with geometric deep learning and generative model,'' \emph{IEEE Robotics and Automation Letters}, vol.~5, no.~4, pp. 6956--6963, 2020.

\bibitem{shi2020calibrcnn}
J.~Shi, Z.~Zhu, J.~Zhang, R.~Liu, Z.~Wang, S.~Chen, and H.~Liu, ``Calibrcnn: Calibrating camera and lidar by recurrent convolutional neural network and geometric constraints,'' in \emph{IEEE/RSJ International Conference on Intelligent Robots and Systems}, 2020, pp. 10\,197--10\,202.

\bibitem{zhu2020online}
Y.~Zhu, C.~Li, and Y.~Zhang, ``Online camera-lidar calibration with sensor semantic information,'' in \emph{IEEE International Conference on Robotics and Automation}, 2020, pp. 4970--4976.

\bibitem{pascal-voc-2012}
M.~Everingham, L.~Van~Gool, C.~K.~I. Williams, J.~Winn, and A.~Zisserman, ``The {PASCAL} {V}isual {O}bject {C}lasses {C}hallenge 2012 {(VOC2012)} {R}esults,'' http://www.pascal-network.org/challenges/VOC/voc2012/workshop/index.html.

\bibitem{wang2020soic}
W.~Wang, S.~Nobuhara, R.~Nakamura, and K.~Sakurada, ``Soic: Semantic online initialization and calibration for lidar and camera,'' \emph{arXiv preprint arXiv:2003.04260}, 2020.

\bibitem{wu2021netcalib}
S.~Wu, A.~Hadachi, D.~Vivet, and Y.~Prabhakar, ``Netcalib: A novel approach for lidar-camera auto-calibration based on deep learning,'' in \emph{International Conference on Pattern Recognition}, 2021, pp. 6648--6655.

\bibitem{DLKFM}
Y.~Zhao, X.~Huang, and Z.~Zhang, ``Deep lucas-kanade homography for multimodal image alignment,'' in \emph{Proceedings of the IEEE/CVF Conference on Computer Vision and Pattern Recognition}, 2021, pp. 15\,950--15\,959.

\bibitem{LocalTrans}
R.~Shao, G.~Wu, Y.~Zhou, Y.~Fu, L.~Fang, and Y.~Liu, ``Localtrans: A multiscale local transformer network for cross-resolution homography estimation,'' in \emph{International Conference on Computer Vision}, 2021, pp. 14\,890--14\,899.

\bibitem{ShuffleHomoNet}
Y.~Chen, G.~Wang, P.~An, Z.~You, and X.~Huang, ``Fast and accurate homography estimation using extendable compression network,'' in \emph{International Conference on Image Processing}, 2021, pp. 1024--1028.

\bibitem{UDIS}
L.~Nie, C.~Lin, K.~Liao, S.~Liu, and Y.~Zhao, ``Unsupervised deep image stitching: Reconstructing stitched features to images,'' \emph{IEEE Transactions on Image Processing}, vol.~30, pp. 6184--6197, 2021.

\bibitem{lv2021lccnet}
X.~Lv, B.~Wang, Z.~Dou, D.~Ye, and S.~Wang, ``Lccnet: Lidar and camera self-calibration using cost volume network,'' in \emph{Proceedings of the IEEE/CVF Conference on Computer Vision and Pattern Recognition}, 2021, pp. 2894--2901.

\bibitem{lv2021cfnet}
X.~Lv, S.~Wang, and D.~Ye, ``Cfnet: Lidar-camera registration using calibration flow network,'' \emph{Sensors}, vol.~21, no.~23, p. 8112, 2021.

\bibitem{liu2021semalign}
Z.~Liu, H.~Tang, S.~Zhu, and S.~Han, ``Semalign: Annotation-free camera-lidar calibration with semantic alignment loss,'' in \emph{IEEE/RSJ International Conference on Intelligent Robots and Systems}, 2021, pp. 8845--8851.

\bibitem{IHN}
S.-Y. Cao, J.~Hu, Z.~Sheng, and H.-L. Shen, ``Iterative deep homography estimation,'' pp. 1879--1888, 2022.

\bibitem{jing2022dxq}
X.~Jing, X.~Ding, R.~Xiong, H.~Deng, and Y.~Wang, ``Dxq-net: Differentiable lidar-camera extrinsic calibration using quality-aware flow,'' in \emph{IEEE/RSJ International Conference on Intelligent Robots and Systems}, 2022, pp. 6235--6241.

\bibitem{ATOP}
Y.~Sun, J.~Li, Y.~Wang, X.~Xu, X.~Yang, and Z.~Sun, ``Atop: An attention-to-optimization approach for automatic lidar-camera calibration via cross-modal object matching,'' \emph{IEEE Transactions on Intelligent Vehicles}, 2022.

\bibitem{wang2022fusionnet}
G.~Wang, J.~Qiu, Y.~Guo, and H.~Wang, ``Fusionnet: Coarse-to-fine extrinsic calibration network of lidar and camera with hierarchical point-pixel fusion,'' in \emph{International Conference on Robotics and Automation}, 2022, pp. 8964--8970.

\bibitem{RGKCNet}
C.~Ye, H.~Pan, and H.~Gao, ``Keypoint-based lidar-camera online calibration with robust geometric network,'' \emph{IEEE Transactions on Instrumentation and Measurement}, vol.~71, pp. 1--11, 2021.

\bibitem{BinoStereo}
J.~G. Song and J.~W. Lee, ``A cnn-based online self-calibration of binocular stereo cameras for pose change,'' \emph{IEEE Transactions on Intelligent Vehicles}, vol.~9, no.~1, pp. 2542--2552, 2023.

\bibitem{DPO-Net}
B.~Roessle and M.~Nie{\ss}ner, ``End2end multi-view feature matching with differentiable pose optimization,'' in \emph{Proceedings of the IEEE/CVF International Conference on Computer Vision}, 2023.

\bibitem{RHWF}
S.-Y. Cao, R.~Zhang, L.~Luo, B.~Yu, Z.~Sheng, J.~Li, and H.-L. Shen, ``Recurrent homography estimation using homography-guided image warping and focus transformer,'' in \emph{Proceedings of the IEEE/CVF Conference on Computer Vision and Pattern Recognition}, 2023, pp. 9833--9842.

\bibitem{EAC-Homo}
X.~Feng, Q.~Jia, Z.~Zhao, Y.~Liu, X.~Xue, and X.~Fan, ``Edge-aware correlation learning for unsupervised progressive homography estimation,'' \emph{IEEE Transactions on Circuits and Systems for Video Technology}, 2023.

\bibitem{PLS-Homo}
Z.~Zhou, Q.~Zhu, M.~Feng, Y.~Wang, J.~Luo, Z.~Miao, L.~Chen, and Y.~Mo, ``Unsupervised homography estimation with pixel-level svdd,'' \emph{IEEE Transactions on Circuits and Systems for Video Technology}, 2024.

\bibitem{LBHomo}
H.~Jiang, H.~Li, Y.~Lu, S.~Han, and S.~Liu, ``Semi-supervised deep large-baseline homography estimation with progressive equivalence constraint,'' in \emph{Proceedings of the AAAI Conference on Artificial Intelligence}, vol.~37, no.~1, 2023, pp. 1024--1032.

\bibitem{RealSH}
H.~Jiang, H.~Li, S.~Han, H.~Fan, B.~Zeng, and S.~Liu, ``Supervised homography learning with realistic dataset generation,'' in \emph{Proceedings of the IEEE/CVF International Conference on Computer Vision}, 2023, pp. 9806--9815.

\bibitem{Gyroflow+}
H.~Li, K.~Luo, B.~Zeng, and S.~Liu, ``Gyroflow+: Gyroscope-guided unsupervised deep homography and optical flow learning,'' \emph{International Journal of Computer Vision}, vol. 132, no.~6, pp. 2331--2349, 2024.

\bibitem{SE-Calib}
Y.~Liao, J.~Li, S.~Kang, Q.~Li, G.~Zhu, S.~Yuan, Z.~Dong, and B.~Yang, ``Se-calib: Semantic edge-based lidar--camera boresight online calibration in urban scenes,'' \emph{IEEE Transactions on Geoscience and Remote Sensing}, vol.~61, pp. 1--13, 2023.

\bibitem{CM-GNN}
J.~Zhu, X.~Li, Q.~Xu, and Z.~Sun, ``Robust online calibration of lidar and camera based on cross-modal graph neural network,'' \emph{IEEE Transactions on Instrumentation and Measurement}, 2023.

\bibitem{Calibdepth}
J.~Zhu, J.~Xue, and P.~Zhang, ``Calibdepth: Unifying depth map representation for iterative lidar-camera online calibration,'' in \emph{IEEE International Conference on Robotics and Automation}, 2023, pp. 726--733.

\bibitem{DEdgeNet}
Y.~Hu, H.~Ma, L.~Jie, and H.~Zhang, ``Dedgenet: Extrinsic calibration of camera and lidar with depth-discontinuous edges,'' in \emph{IEEE International Conference on Robotics and Automation}, 2023, pp. 11\,439--11\,445.

\bibitem{MOISST}
Q.~Herau, N.~Piasco, M.~Bennehar, L.~Rold{\~a}o, D.~Tsishkou, C.~Migniot, P.~Vasseur, and C.~Demonceaux, ``Moisst: Multimodal optimization of implicit scene for spatiotemporal calibration,'' in \emph{IEEE/RSJ International Conference on Intelligent Robots and Systems}, 2023, pp. 1810--1817.

\bibitem{ELR-Calib}
Z.~Zhang, Z.~Yu, S.~You, R.~Rao, S.~Agarwal, and F.~Ren, ``Enhanced low-resolution lidar-camera calibration via depth interpolation and supervised contrastive learning,'' in \emph{IEEE International Conference on Acoustics, Speech and Signal Processing}, 2023, pp. 1--5.

\bibitem{RELLIS-3D}
P.~Jiang, P.~Osteen, M.~Wigness, and S.~Saripalli, ``Rellis-3d dataset: Data, benchmarks and analysis,'' in \emph{IEEE International Conference on Robotics and Automation}, 2021, pp. 1110--1116.

\bibitem{P2O-Calib}
S.~Wang, S.~Zhang, and X.~Qiu, ``P2o-calib: Camera-lidar calibration using point-pair spatial occlusion relationship,'' in \emph{IEEE/RSJ International Conference on Intelligent Robots and Systems}, 2023, pp. 1840--1847.

\bibitem{SCNet}
Z.~Duan, X.~Hu, J.~Ding, P.~An, X.~Huang, and J.~Ma, ``A robust lidar-camera self-calibration via rotation-based alignment and multi-level cost volume,'' \emph{IEEE Robotics and Automation Letters}, vol.~9, no.~1, pp. 627--634, 2023.

\bibitem{RobustCalib}
S.~Xu, S.~Zhou, Z.~Tian, J.~Ma, Q.~Nie, and X.~Chu, ``Robustcalib: Robust lidar-camera extrinsic calibration with consistency learning,'' \emph{arXiv preprint arXiv:2312.01085}, 2023.

\bibitem{PseudoCal}
M.~Cocheteux, J.~Moreau, and F.~Davoine, ``Pseudocal: Towards initialisation-free deep learning-based camera-lidar self-calibration,'' \emph{British Machine Vision Conference}, 2023.

\bibitem{BatchCalib}
L.~F.~T. Fu and M.~Fallon, ``Batch differentiable pose refinement for in-the-wild camera/lidar extrinsic calibration,'' in \emph{Conference on Robot Learning}, 2023, pp. 1362--1377.

\bibitem{NeuralRecalibration}
B.~Mehmandar, R.~Talakoob, and C.~Poullis, ``Neural real-time recalibration for infrared multi-camera systems,'' \emph{arXiv preprint arXiv:2410.14505}, 2024.

\bibitem{ArcGeo}
M.~Shugaev, I.~Semenov, K.~Ashley, M.~Klaczynski, N.~Cuntoor, M.~W. Lee, and N.~Jacobs, ``Arcgeo: Localizing limited field-of-view images using cross-view matching,'' in \emph{Proceedings of the IEEE/CVF Winter Conference on Applications of Computer Vision}, 2024, pp. 209--218.

\bibitem{CVUSA}
S.~Workman, R.~Souvenir, and N.~Jacobs, ``Wide-area image geolocalization with aerial reference imagery,'' in \emph{Proceedings of the IEEE International Conference on Computer Vision}, 2015, pp. 3961--3969.

\bibitem{CVACT}
L.~Liu and H.~Li, ``Lending orientation to neural networks for cross-view geo-localization,'' in \emph{Proceedings of the IEEE/CVF Conference on Computer Vision and Pattern Recognition}, 2019, pp. 5624--5633.

\bibitem{CalibRBEV}
W.~Liao, S.~Qiang, X.~Li, X.~Chen, H.~Wang, Y.~Liang, J.~Yan, T.~He, and P.~Peng, ``Calibrbev: Multi-camera calibration via reversed bird's-eye-view representations for autonomous driving,'' in \emph{Proceedings of the ACM International Conference on Multimedia}, 2024, pp. 9145--9154.

\bibitem{Waymo}
P.~Sun, H.~Kretzschmar, X.~Dotiwalla, A.~Chouard \emph{et~al.}, ``Scalability in perception for autonomous driving: Waymo open dataset,'' in \emph{Proceedings of the IEEE/CVF Conference on Computer Vision and Pattern Recognition}, 2020, pp. 2446--2454.

\bibitem{FG-Rect}
A.~Kumar, F.~Mannan, O.~H. Jafari, S.~Li, and F.~Heide, ``Flow-guided online stereo rectification for wide baseline stereo,'' in \emph{Proceedings of the IEEE/CVF Conference on Computer Vision and Pattern Recognition}, 2024, pp. 15\,375--15\,385.

\bibitem{Mask-Homo}
Y.~Wang, H.~Liu, C.~Zhang, L.~Xu, and Q.~Wang, ``Mask-homo: Pseudo plane mask-guided unsupervised multi-homography estimation,'' in \emph{Proceedings of the AAAI Conference on Artificial Intelligence}, vol.~38, no.~6, 2024, pp. 5678--5685.

\bibitem{DSM-DHN}
S.~Cheng, Z.~Chen, L.~Guo, and W.~Tao, ``Deep homography estimation via dense scene matching,'' \emph{IEEE Robotics and Automation Letters}, 2024.

\bibitem{DMHomo}
H.~Li, H.~Jiang, A.~Luo, P.~Tan, H.~Fan, B.~Zeng, and S.~Liu, ``Dmhomo: Learning homography with diffusion models,'' \emph{ACM Transactions on Graphics}, vol.~43, no.~3, pp. 1--16, 2024.

\bibitem{DPP-Homo}
X.~Feng, Q.~Jia, Y.~Liu, X.~Fan, and L.~J. Latecki, ``Depth-guided dominant plane perception for unsupervised homography estimation,'' in \emph{IEEE International Conference on Acoustics, Speech and Signal Processing}, 2024, pp. 3480--3484.

\bibitem{DHE-VPR}
F.~Lu, S.~Dong, L.~Zhang, B.~Liu, X.~Lan, D.~Jiang, and C.~Yuan, ``Deep homography estimation for visual place recognition,'' in \emph{Proceedings of the AAAI Conference on Artificial Intelligence}, vol.~38, no.~9, 2024, pp. 10\,341--10\,349.

\bibitem{Pitts30k}
A.~Torii, J.~Sivic, T.~Pajdla, and M.~Okutomi, ``Visual place recognition with repetitive structures,'' in \emph{Proceedings of the IEEE Conference on Computer Vision and Pattern Recognition}, 2013, pp. 883--890.

\bibitem{SRMatcher}
Y.~Liu, Q.~Huang, S.~Hui, J.~Fu, S.~Zhou, K.~Wu, P.~Li, and J.~Wang, ``Semantic-aware representation learning for homography estimation,'' in \emph{Proceedings of the ACM International Conference on Multimedia}, 2024, pp. 2506--2514.

\bibitem{Oxford5K}
J.~Philbin, O.~Chum, M.~Isard, J.~Sivic, and A.~Zisserman, ``Object retrieval with large vocabularies and fast spatial matching,'' in \emph{IEEE Conference on Computer Vision and Pattern Recognition}, 2007.

\bibitem{Paris6K}
J.~\vspace{0mm}Philbin, O.~Chum, M.~Isard, J.~Sivic, and A.~Zisserman, ``Lost in quantization: Improving particular object retrieval in large scale image databases,'' in \emph{Proceedings of the IEEE Conference on Computer Vision and Pattern Recognition}, 2008.

\bibitem{AbHE}
M.~Huo, Z.~Zhang, X.~Ren, X.~Yang, and C.~Ye, ``Abhe: All attention-based homography estimation,'' \emph{IEEE Transactions on Instrumentation and Measurement}, 2024.

\bibitem{AGNet}
Z.~Li, F.~Fang, T.~Wang, and G.~Zhang, ``Homography estimation with adaptive query transformer and gated interaction module,'' \emph{IEEE Transactions on Circuits and Systems for Video Technology}, 2024.

\bibitem{CrossHomo}
X.~Deng, E.~Liu, C.~Gao, S.~Li, S.~Gu, and M.~Xu, ``Crosshomo: Cross-modality and cross-resolution homography estimation,'' \emph{IEEE Transactions on Pattern Analysis and Machine Intelligence}, 2024.

\bibitem{DPDN}
G.~Riegler, D.~Ferstl, M.~R{\"u}ther, and H.~Bischof, ``A deep primal-dual network for guided depth super-resolution,'' \emph{arXiv preprint arXiv:1607.08569}, 2016.

\bibitem{AltO}
S.~Song, J.~Lew, H.~Jang, and S.~Yoon, ``Unsupervised homography estimation on multimodal image pair via alternating optimization,'' \emph{Advances in Neural Information Processing Systems}, 2024.

\bibitem{DeepNIR}
I.~Sa, J.~Y. Lim, H.~S. Ahn, and B.~MacDonald, ``deepnir: Datasets for generating synthetic nir images and improved fruit detection system using deep learning techniques,'' \emph{Sensors}, vol.~22, no.~13, p. 4721, 2022.

\bibitem{Gyroflow}
H.~Li, K.~Luo, and S.~Liu, ``Gyroflow: Gyroscope-guided unsupervised optical flow learning,'' in \emph{Proceedings of the IEEE/CVF International Conference on Computer Vision}, 2021, pp. 12\,869--12\,878.

\bibitem{HEN}
M.~Shao, T.~Tasdizen, and S.~Joshi, ``Analyzing the domain shift immunity of deep homography estimation,'' in \emph{Proceedings of the IEEE/CVF Winter Conference on Applications of Computer Vision}, 2024, pp. 4800--4808.

\bibitem{JEDL-Homo}
Q.~Jia, Z.~Zhao, X.~Feng, J.~Liu, Y.~Liu, and X.~Xue, ``Joint edge detection learning for recurrent homography estimation,'' in \emph{IEEE International Conference on Multimedia and Expo}, 2024, pp. 1--6.

\bibitem{InterNet}
J.~Yu, S.-Y. Cao, R.~Zhang, C.~Zhang, J.~Hu, Z.~Yu, B.~Yu, and H.-l. Shen, ``Internet: Unsupervised cross-modal homography estimation based on interleaved modality transfer and self-supervised homography prediction,'' \emph{arXiv preprint arXiv:2409.17993}, 2024.

\bibitem{RGB/NIR}
M.~Brown and S.~S{\"u}sstrunk, ``Multi-spectral sift for scene category recognition,'' in \emph{Proceedings of the IEEE/CVF Conference on Computer Vision and Pattern Recognition}, 2011, pp. 177--184.

\bibitem{STHN}
J.~Xiao, N.~Zhang, D.~Tortei, and G.~Loianno, ``Sthn: Deep homography estimation for uav thermal geo-localization with satellite imagery,'' \emph{IEEE Robotics and Automation Letters}, vol.~9, no.~10, pp. 8754--8761, 2024.

\bibitem{Boson-nighttime}
J.~Xiao, D.~Tortei, E.~Roura, and G.~Loianno, ``Long-range uav thermal geo-localization with satellite imagery,'' in \emph{IEEE/RSJ International Conference on Intelligent Robots and Systems}, 2023, pp. 5820--5827.

\bibitem{SCPNet}
R.~Zhang, J.~Ma, S.-Y. Cao, L.~Luo, B.~Yu, S.-J. Chen, J.~Li, and H.-L. Shen, ``Scpnet: Unsupervised cross-modal homography estimation via intra-modal self-supervised learning,'' in \emph{European Conference on Computer Vision}, 2024, pp. 460--477.

\bibitem{MCNet}
H.~Zhu, S.-Y. Cao, J.~Hu, S.~Zuo, B.~Yu, J.~Ying, J.~Li, and H.-L. Shen, ``Mcnet: Rethinking the core ingredients for accurate and efficient homography estimation,'' in \emph{Proceedings of the IEEE/CVF Conference on Computer Vision and Pattern Recognition}, 2024, pp. 25\,932--25\,941.

\bibitem{CodingHomo}
Y.~Liu, H.~Li, S.~Liu, and B.~Zeng, ``Codinghomo: Bootstrapping deep homography with video coding,'' \emph{IEEE Transactions on Circuits and Systems for Video Technology}, 2024.

\bibitem{SOAC}
Q.~Herau, N.~Piasco, M.~Bennehar, L.~Roldao, D.~Tsishkou, C.~Migniot, P.~Vasseur, and C.~Demonceaux, ``Soac: Spatio-temporal overlap-aware multi-sensor calibration using neural radiance fields,'' in \emph{Proceedings of the IEEE/CVF Conference on Computer Vision and Pattern Recognition}, 2024, pp. 15\,131--15\,140.

\bibitem{PandaSet}
P.~Xiao, Z.~Shao, S.~Hao, Z.~Zhang, X.~Chai, J.~Jiao, Z.~Li, J.~Wu, K.~Sun, K.~Jiang \emph{et~al.}, ``Pandaset: Advanced sensor suite dataset for autonomous driving,'' in \emph{IEEE International Intelligent Transportation Systems Conference}, 2021, pp. 3095--3101.

\bibitem{UniCal}
Z.~Yang, G.~Chen, H.~Zhang, K.~Ta, I.~A. B{\^a}rsan, D.~Murphy, S.~Manivasagam, and R.~Urtasun, ``Unical: Unified neural sensor calibration,'' in \emph{European Conference on Computer Vision}, 2024, pp. 327--345.

\bibitem{L2C-Calib}
F.~Liu, Y.~Cao, X.~Cheng, and X.~Wu, ``Transformer-based local-to-global lidar-camera targetless calibration with multiple constraints,'' \emph{IEEE Transactions on Instrumentation and Measurement}, 2024.

\bibitem{CalibFormer}
Y.~Xiao, Y.~Li, C.~Meng, X.~Li, J.~Ji, and Y.~Zhang, ``Calibformer: A transformer-based automatic lidar-camera calibration network,'' in \emph{IEEE International Conference on Robotics and Automation}, 2024, pp. 16\,714--16\,720.

\bibitem{LCCRAFT}
Y.-C. Lee and K.-W. Chen, ``Lccraft: Lidar and camera calibration using recurrent all-pairs field transforms without precise initial guess,'' in \emph{IEEE International Conference on Robotics and Automation}, 2024, pp. 16\,669--16\,675.

\bibitem{SGCalib}
Z.~Lin, Z.~Gao, X.~Liu, J.~Wang, W.~Song, B.~M. Chen, C.~Li, Y.~Huang, and Y.~Zhu, ``Sgcalib: A two-stage camera-lidar calibration method using semantic information and geometric features,'' in \emph{IEEE International Conference on Robotics and Automation}, 2024, pp. 14\,527--14\,533.

\bibitem{LCANet}
A.~Zhu, Y.~Xiao, C.~Liu, M.~Tan, and Z.~Cao, ``Lightweight lidar-camera alignment with homogeneous local-global aware representation,'' \emph{IEEE Transactions on Intelligent Transportation Systems}, 2024.

\bibitem{SAM-Calib}
Z.~Luo, G.~Yan, X.~Cai, and B.~Shi, ``Zero-training lidar-camera extrinsic calibration method using segment anything model,'' in \emph{IEEE International Conference on Robotics and Automation}, 2024, pp. 14\,472--14\,478.

\bibitem{HIFMNet}
X.~Hu, Z.~Duan, J.~Ding, Z.~Zhang, X.~Huang, and J.~Ma, ``Lidar-camera extrinsic calibration with hierachical and iterative feature matching,'' in \emph{IEEE International Conference on Robotics and Automation}, 2024, pp. 16\,691--16\,697.

\bibitem{SensorX2Vehicle}
G.~Yan, Z.~Luo, Z.~Liu, Y.~Li, B.~Shi, and K.~Zhang, ``Sensorx2vehicle: Online sensors-to-vehicle rotation calibration methods in road scenarios,'' \emph{IEEE Robotics and Automation Letters}, 2024.

\bibitem{Edgecalib}
X.~Li, Y.~Duan, B.~Wang, H.~Ren, G.~You, Y.~Sheng, J.~Ji, and Y.~Zhang, ``Edgecalib: Multi-frame weighted edge features for automatic targetless lidar-camera calibration,'' \emph{IEEE Robotics and Automation Letters}, 2024.

\bibitem{wu2013towards}
C.~Wu, ``Towards linear-time incremental structure from motion,'' in \emph{International Conference on 3D Vision}, 2013, pp. 127--134.

\bibitem{detone2018superpoint}
D.~DeTone, T.~Malisiewicz, and A.~Rabinovich, ``Superpoint: Self-supervised interest point detection and description,'' in \emph{Proceedings of the IEEE/CVF Conference on Computer Vision and Pattern Recognition Workshop}, 2018, pp. 224--236.

\bibitem{sun2021loftr}
J.~Sun, Z.~Shen, Y.~Wang, H.~Bao, and X.~Zhou, ``Loftr: Detector-free local feature matching with transformers,'' in \emph{Proceedings of the IEEE/CVF Conference on Computer Vision and Pattern Recognition}, 2021, pp. 8922--8931.

\bibitem{sarlin2020superglue}
P.-E. Sarlin, D.~DeTone, T.~Malisiewicz, and A.~Rabinovich, ``Superglue: Learning feature matching with graph neural networks,'' in \emph{Proceedings of the IEEE/CVF Conference on Computer Vision and Pattern Recognition}, 2020, pp. 4938--4947.

\bibitem{lindenberger2023lightglue}
P.~Lindenberger, P.-E. Sarlin, and M.~Pollefeys, ``Lightglue: Local feature matching at light speed,'' in \emph{Proceedings of the IEEE/CVF International Conference on Computer Vision}, 2023, pp. 17\,627--17\,638.

\bibitem{brachmann2017dsac}
E.~Brachmann, A.~Krull, S.~Nowozin, J.~Shotton, F.~Michel, S.~Gumhold, and C.~Rother, ``Dsac-differentiable ransac for camera localization,'' in \emph{Proceedings of the IEEE/CVF Conference on Computer Vision and Pattern Recognition}, 2017, pp. 6684--6692.

\bibitem{sarlin2021back}
P.-E. Sarlin, A.~Unagar, M.~Larsson, H.~Germain, C.~Toft, V.~Larsson, M.~Pollefeys, V.~Lepetit, L.~Hammarstrand, F.~Kahl \emph{et~al.}, ``Back to the feature: Learning robust camera localization from pixels to pose,'' in \emph{Proceedings of the IEEE/CVF Conference on Computer Vision and Pattern Recognition}, 2021, pp. 3247--3257.

\bibitem{wang2017deepvo}
S.~Wang, R.~Clark, H.~Wen, and N.~Trigoni, ``Deepvo: Towards end-to-end visual odometry with deep recurrent convolutional neural networks,'' in \emph{IEEE International Conference on Robotics and Automation}, 2017, pp. 2043--2050.

\bibitem{li2018undeepvo}
R.~Li, S.~Wang, Z.~Long, and D.~Gu, ``Undeepvo: Monocular visual odometry through unsupervised deep learning,'' in \emph{IEEE International Conference on Robotics and Automation}, 2018, pp. 7286--7291.

\bibitem{bian2019unsupervised}
J.~Bian, Z.~Li, N.~Wang, H.~Zhan, C.~Shen, M.-M. Cheng, and I.~Reid, ``Unsupervised scale-consistent depth and ego-motion learning from monocular video,'' \emph{Advances in Neural Information Processing Systems}, vol.~32, 2019.

\bibitem{yang2020d3vo}
N.~Yang, L.~v. Stumberg, R.~Wang, and D.~Cremers, ``D3vo: Deep depth, deep pose and deep uncertainty for monocular visual odometry,'' in \emph{Proceedings of the IEEE/CVF Conference on Computer Vision and Pattern Recognition}, 2020, pp. 1281--1292.

\bibitem{zhu2024revisit}
S.~Zhu and X.~Liu, ``Revisit self-supervised depth estimation with local structure-from-motion,'' in \emph{European Conference on Computer Vision}.\hskip 1em plus 0.5em minus 0.4em\relax Springer, 2024, pp. 38--56.

\bibitem{min2020voldor}
Z.~Min, Y.~Yang, and E.~Dunn, ``Voldor: Visual odometry from log-logistic dense optical flow residuals,'' in \emph{Proceedings of the IEEE/CVF Conference on Computer Vision and Pattern Recognition}, 2020, pp. 4898--4909.

\bibitem{teed2021droid}
Z.~Teed and J.~Deng, ``Droid-slam: Deep visual slam for monocular, stereo, and rgb-d cameras,'' \emph{Advances in Neural Information Processing Systems}, vol.~34, pp. 16\,558--16\,569, 2021.

\bibitem{wang2021tartanvo}
W.~Wang, Y.~Hu, and S.~Scherer, ``Tartanvo: A generalizable learning-based vo,'' in \emph{Conference on Robot Learning}, 2021, pp. 1761--1772.

\bibitem{yu2018ds}
C.~Yu, Z.~Liu, X.-J. Liu, F.~Xie, Y.~Yang, Q.~Wei, and Q.~Fei, ``Ds-slam: A semantic visual slam towards dynamic environments,'' in \emph{IEEE/RSJ International Conference on Intelligent Robots and Systems}, 2018, pp. 1168--1174.

\bibitem{wu2020eao}
Y.~Wu, Y.~Zhang, D.~Zhu, Y.~Feng, S.~Coleman, and D.~Kerr, ``Eao-slam: Monocular semi-dense object slam based on ensemble data association,'' in \emph{IEEE/RSJ International Conference on Intelligent Robots and Systems}, 2020, pp. 4966--4973.

\bibitem{chen2022accurate}
K.~Chen, J.~Liu, Q.~Chen, Z.~Wang, and J.~Zhang, ``Accurate object association and pose updating for semantic slam,'' \emph{IEEE Transactions on Intelligent Transportation Systems}, vol.~23, no.~12, pp. 25\,169--25\,179, 2022.

\bibitem{sucar2021imap}
E.~Sucar, S.~Liu, J.~Ortiz, and A.~J. Davison, ``imap: Implicit mapping and positioning in real-time,'' in \emph{Proceedings of the IEEE/CVF International Conference on Computer Vision}, 2021, pp. 6229--6238.

\bibitem{zhu2022nice}
Z.~Zhu, S.~Peng, V.~Larsson, W.~Xu, H.~Bao, Z.~Cui, M.~R. Oswald, and M.~Pollefeys, ``Nice-slam: Neural implicit scalable encoding for slam,'' in \emph{Proceedings of the IEEE/CVF Conference on Computer Vision and Pattern Recognition}, 2022, pp. 12\,786--12\,796.

\bibitem{rosinol2022nerf}
A.~Rosinol, J.~J. Leonard, and L.~Carlone, ``Nerf-slam: Real-time dense monocular slam with neural radiance fields,'' \emph{arXiv preprint arXiv:2210.13641}, 2022.

\bibitem{ortiz2022isdf}
J.~Ortiz, A.~Clegg, J.~Dong, E.~Sucar, D.~Novotny, M.~Zollhoefer, and M.~Mukadam, ``isdf: Real-time neural signed distance fields for robot perception,'' \emph{arXiv preprint arXiv:2204.02296}, 2022.

\bibitem{johari2023eslam}
M.~M. Johari, C.~Carta, and F.~Fleuret, ``Eslam: Efficient dense slam system based on hybrid representation of signed distance fields,'' in \emph{Proceedings of the IEEE/CVF Conference on Computer Vision and Pattern Recognition}, 2023, pp. 17\,408--17\,419.

\bibitem{sandstrom2023point}
E.~Sandstr{\"o}m, Y.~Li, L.~Van~Gool, and M.~R. Oswald, ``Point-slam: Dense neural point cloud-based slam,'' in \emph{Proceedings of the IEEE/CVF International Conference on Computer Vision}, 2023, pp. 18\,433--18\,444.

\bibitem{hu2024cp}
J.~Hu, M.~Mao, H.~Bao, G.~Zhang, and Z.~Cui, ``Cp-slam: Collaborative neural point-based slam system,'' \emph{Advances in Neural Information Processing Systems}, vol.~36, 2024.

\bibitem{yan2024gs}
C.~Yan, D.~Qu, D.~Xu, B.~Zhao, Z.~Wang, D.~Wang, and X.~Li, ``Gs-slam: Dense visual slam with 3d gaussian splatting,'' in \emph{Proceedings of the IEEE/CVF Conference on Computer Vision and Pattern Recognition}, 2024, pp. 19\,595--19\,604.

\bibitem{truong2023sparf}
P.~Truong, M.-J. Rakotosaona, F.~Manhardt, and F.~Tombari, ``Sparf: Neural radiance fields from sparse and noisy poses,'' in \emph{Proceedings of the IEEE/CVF Conference on Computer Vision and Pattern Recognition}, 2023, pp. 4190--4200.

\bibitem{wang2023altnerf}
K.~Wang, Z.~Yan, H.~Tian, Z.~Zhang, X.~Li, J.~Li, and J.~Yang, ``Altnerf: Learning robust neural radiance field via alternating depth-pose optimization,'' vol.~38, no.~6, pp. 5508--5516, 2024.

\bibitem{jeong2021self}
Y.~Jeong, S.~Ahn, C.~Choy, A.~Anandkumar, M.~Cho, and J.~Park, ``Self-calibrating neural radiance fields,'' in \emph{Proceedings of the IEEE/CVF International Conference on Computer Vision}, 2021, pp. 5846--5854.

\bibitem{bian2023nope}
W.~Bian, Z.~Wang, K.~Li, J.-W. Bian, and V.~A. Prisacariu, ``Nope-nerf: Optimising neural radiance field with no pose prior,'' in \emph{Proceedings of the IEEE/CVF Conference on Computer Vision and Pattern Recognition}, 2023, pp. 4160--4169.

\bibitem{meng2021gnerf}
Q.~Meng, A.~Chen, H.~Luo, M.~Wu, H.~Su, L.~Xu, X.~He, and J.~Yu, ``Gnerf: Gan-based neural radiance field without posed camera,'' in \emph{Proceedings of the IEEE/CVF International Conference on Computer Vision}, 2021, pp. 6351--6361.

\bibitem{howard2019searching}
A.~Howard, M.~Sandler, G.~Chu, L.-C. Chen, B.~Chen, M.~Tan, W.~Wang, Y.~Zhu, R.~Pang, V.~Vasudevan \emph{et~al.}, ``Searching for mobilenetv3,'' in \emph{Proceedings of the IEEE/CVF International Conference on Computer Vision}, 2019, pp. 1314--1324.

\bibitem{wang2021nerf}
Z.~Wang, S.~Wu, W.~Xie, M.~Chen, and V.~A. Prisacariu, ``Nerf--: Neural radiance fields without known camera parameters,'' \emph{arXiv preprint arXiv:2102.07064}, 2021.

\bibitem{xia2022sinerf}
Y.~Xia, H.~Tang, R.~Timofte, and L.~Van~Gool, ``Sinerf: Sinusoidal neural radiance fields for joint pose estimation and scene reconstruction,'' \emph{British Machine Vision Conference}, 2022.

\bibitem{park2023camp}
K.~Park, P.~Henzler, B.~Mildenhall, J.~T. Barron, and R.~Martin-Brualla, ``Camp: Camera preconditioning for neural radiance fields,'' \emph{arXiv preprint arXiv:2308.10902}, 2023.

\bibitem{muller2022instant}
T.~M{\"u}ller, A.~Evans, C.~Schied, and A.~Keller, ``Instant neural graphics primitives with a multiresolution hash encoding,'' \emph{ACM Transactions on Graphics}, vol.~41, no.~4, pp. 1--15, 2022.

\bibitem{tancik2023nerfstudio}
M.~Tancik, E.~Weber, E.~Ng, R.~Li, B.~Yi, T.~Wang, A.~Kristoffersen, J.~Austin, K.~Salahi, A.~Ahuja \emph{et~al.}, ``Nerfstudio: A modular framework for neural radiance field development,'' in \emph{ACM SIGGRAPH Conference Proceedings}, 2023, pp. 1--12.

\bibitem{qi2017pointnet}
C.~R. Qi, H.~Su, K.~Mo, and L.~J. Guibas, ``Pointnet: Deep learning on point sets for 3d classification and segmentation,'' in \emph{Proceedings of the IEEE Conference on Computer Vision and Pattern Recognition}, 2017, pp. 652--660.

\bibitem{li2018pointcnn}
Y.~Li, R.~Bu, M.~Sun, W.~Wu, X.~Di, and B.~Chen, ``Pointcnn: Convolution on x-transformed points,'' \emph{Advances in Neural Information Processing Systems}, vol.~31, 2018.

\bibitem{lin2023learning}
E.~Y. Lin, Z.~Wang, R.~Lin, D.~Miau, F.~Kainz, J.~Chen, X.~C. Zhang, D.~B. Lindell, and K.~N. Kutulakos, ``Learning lens blur fields,'' \emph{arXiv preprint arXiv:2310.11535}, 2023.

\bibitem{DarSwin-Unet}
A.~Athwale, I.~Shili, {\'E}.~Bergeron, O.~Ahmad, and J.-F. Lalonde, ``Darswin-unet: Distortion aware encoder-decoder architecture,'' \emph{arXiv preprint arXiv:2407.17328}, 2024.

\bibitem{Sun_2018_CVPR}
D.~Sun, X.~Yang, M.-Y. Liu, and J.~Kautz, ``Pwc-net: Cnns for optical flow using pyramid, warping, and cost volume,'' in \emph{Proceedings of the IEEE Conference on Computer Vision and Pattern Recognition}, 2018.

\bibitem{hartley2003multiple}
R.~Hartley and A.~Zisserman, \emph{Multiple view geometry in computer vision}.\hskip 1em plus 0.5em minus 0.4em\relax Cambridge University Press, 2003.

\bibitem{lucas1981iterative}
B.~D. Lucas, T.~Kanade \emph{et~al.}, \emph{An iterative image registration technique with an application to stereo vision}.\hskip 1em plus 0.5em minus 0.4em\relax Vancouver, 1981, vol.~81.

\bibitem{ma2018shufflenet}
N.~Ma, X.~Zhang, H.-T. Zheng, and J.~Sun, ``Shufflenet v2: Practical guidelines for efficient cnn architecture design,'' in \emph{European conference on computer vision}, 2018, pp. 116--131.

\bibitem{fischler1981random}
M.~A. Fischler and R.~C. Bolles, ``Random sample consensus: a paradigm for model fitting with applications to image analysis and automated cartography,'' \emph{Communications of the ACM}, vol.~24, no.~6, pp. 381--395, 1981.

\bibitem{nie2022learning}
L.~Nie, C.~Lin, K.~Liao, and Y.~Zhao, ``Learning edge-preserved image stitching from multi-scale deep homography,'' \emph{Neurocomputing}, vol. 491, pp. 533--543, 2022.

\bibitem{baker2004lucas}
S.~Baker and I.~Matthews, ``Lucas-kanade 20 years on: A unifying framework,'' \emph{International Journal of Computer Vision}, vol.~56, no.~3, pp. 221--255, 2004.

\bibitem{nocedal1999numerical}
J.~Nocedal and S.~J. Wright, \emph{Numerical optimization}.\hskip 1em plus 0.5em minus 0.4em\relax Springer, 1999.

\bibitem{teed2020raft}
Z.~Teed and J.~Deng, ``Raft: Recurrent all-pairs field transforms for optical flow,'' in \emph{European Conference on Computer Vision}, 2020, pp. 402--419.

\bibitem{li2022ssorn}
Y.~Li, W.~Pei, and Z.~He, ``Ssorn: Self-supervised outlier removal network for robust homography estimation,'' \emph{arXiv preprint arXiv:2208.14093}, 2022.

\bibitem{vaswani2017attention}
A.~Vaswani, N.~Shazeer, N.~Parmar, J.~Uszkoreit, L.~Jones, A.~N. Gomez, {\L}.~Kaiser, and I.~Polosukhin, ``Attention is all you need,'' \emph{Advances in Neural Information Processing Systems}, vol.~30, 2017.

\bibitem{handa2016gvnn}
A.~Handa, M.~Bloesch, V.~P{\u{a}}tr{\u{a}}ucean, S.~Stent, J.~McCormac, and A.~Davison, ``gvnn: Neural network library for geometric computer vision,'' in \emph{European Conference on Computer Vision}.\hskip 1em plus 0.5em minus 0.4em\relax Springer, 2016, pp. 67--82.

\bibitem{qi2017pointnet++}
C.~R. Qi, L.~Yi, H.~Su, and L.~J. Guibas, ``Pointnet++: Deep hierarchical feature learning on point sets in a metric space,'' \emph{Advances in Neural Information Processing Systems}, vol.~30, 2017.

\bibitem{bochkovskiy2020yolov4}
A.~Bochkovskiy, C.-Y. Wang, and H.-Y.~M. Liao, ``Yolov4: Optimal speed and accuracy of object detection,'' \emph{arXiv preprint arXiv:2004.10934}, 2020.

\bibitem{lang2019pointpillars}
A.~H. Lang, S.~Vora, H.~Caesar, L.~Zhou, J.~Yang, and O.~Beijbom, ``Pointpillars: Fast encoders for object detection from point clouds,'' in \emph{Proceedings of the IEEE/CVF Conference on Computer Vision and Pattern Recognition}, 2019, pp. 12\,697--12\,705.

\bibitem{poli2007particle}
R.~Poli, J.~Kennedy, and T.~Blackwell, ``Particle swarm optimization,'' \emph{Swarm Intelligence}, vol.~1, no.~1, pp. 33--57, 2007.

\bibitem{gould2021deep}
S.~Gould, R.~Hartley, and D.~Campbell, ``Deep declarative networks,'' \emph{IEEE Transactions on Pattern Analysis and Machine Intelligence}, vol.~44, no.~8, pp. 3988--4004, 2021.

\bibitem{mildenhall2020nerf}
B.~Mildenhall, P.~P. Srinivasan, M.~Tancik, J.~T. Barron, R.~Ramamoorthi, and R.~Ng, ``Nerf: Representing scenes as neural radiance fields for view synthesis,'' in \emph{European Conference on Computer Vision}, 2020, pp. 405--421.

\bibitem{huang2019apolloscape}
X.~Huang, P.~Wang, X.~Cheng, D.~Zhou, Q.~Geng, and R.~Yang, ``The apolloscape open dataset for autonomous driving and its application,'' \emph{IEEE Transactions on Pattern Analysis and Machine Intelligence}, vol.~42, no.~10, pp. 2702--2719, 2019.

\bibitem{yu2022dair}
H.~Yu, Y.~Luo, M.~Shu, Y.~Huo, Z.~Yang, Y.~Shi, Z.~Guo, H.~Li, X.~Hu, J.~Yuan \emph{et~al.}, ``Dair-v2x: A large-scale dataset for vehicle-infrastructure cooperative 3d object detection,'' in \emph{Proceedings of the IEEE/CVF Conference on Computer Vision and Pattern Recognition}, 2022, pp. 21\,361--21\,370.

\bibitem{kang-2020-jfr}
J.~Kang and N.~L. Doh, ``Automatic targetless camera–{LIDAR} calibration by aligning edge with {Gaussian} mixture model,'' \emph{Journal of Field Robotics}, vol.~37, no.~1, pp. 158--179, 2020.

\bibitem{mao2021one}
J.~Mao, M.~Niu, C.~Jiang, X.~Liang, Y.~Li, C.~Ye, W.~Zhang, Z.~Li, J.~Yu, C.~Xu \emph{et~al.}, ``One million scenes for autonomous driving: Once dataset,'' 2021.

\bibitem{tang2018ba}
C.~Tang and P.~Tan, ``Ba-net: Dense bundle adjustment network,'' \emph{International Conference on Learning Representation}, 2019.

\bibitem{teed2018deepv2d}
Z.~Teed and J.~Deng, ``Deepv2d: Video to depth with differentiable structure from motion,'' \emph{International Conference on Learning Representation}, 2020.

\bibitem{wei2020deepsfm}
X.~Wei, Y.~Zhang, Z.~Li, Y.~Fu, and X.~Xue, ``Deepsfm: Structure from motion via deep bundle adjustment,'' in \emph{European Conference on Computer Vision}, 2020, pp. 230--247.

\bibitem{gu2023dro}
X.~Gu, W.~Yuan, Z.~Dai, S.~Zhu, C.~Tang, Z.~Dong, and P.~Tan, ``Dro: Deep recurrent optimizer for video to depth,'' \emph{IEEE Robotics and Automation Letters}, vol.~8, no.~5, pp. 2844--2851, 2023.

\bibitem{teed2021tangent}
Z.~\vspace{0mm}Teed and J.~Deng, ``Tangent space backpropagation for 3d transformation groups,'' in \emph{Proceedings of the IEEE/CVF Conference on Computer Vision and Pattern Recognition}, 2021.

\bibitem{he_rethinking}
K.~He, R.~Girshick, and P.~Doll{\'a}r, ``Rethinking imagenet pre-training,'' in \emph{International Conference on Computer Vision}, 2019, pp. 4918--4927.

\end{thebibliography}

              

\section*{Supplementary material}

# Deep Learning for Camera Calibration and Beyond: A Survey
## -Supplementary Material-

Kang Liao, Lang Nie, Shujuan Huang, Chunyu Lin, Jing Zhang, Yao Zhao, *Fellow, IEEE*, Moncef Gabbouj, *Fellow, IEEE*, Dacheng Tao, *Fellow, IEEE*

## I. DEVELOPMENT RECAP

### A. Milestone

A concrete milestone from 2015 to 2024 of deep learning-based camera calibration is shown in Figure 1, spanning the main deep learning era. We classify all literature based on the uncalibrated camera model and its extended applications: standard model, distortion model, cross-view model, and cross-sensor model.

### B. Statistic Analysis

As we can observe in Figure 2, the number of learning-based camera calibrations has grown since 2015 and boomed since 2019. And the learning targets are extended from the simple and pure parameters to complicated and hybrid parameters, driven by larger datasets, more reasonable learning strategies, more explicit learning representations, and more solid network architectures, etc.

The data analysis of different learning strategies used in learning-based camera calibration is also shown in Figure 2. From the statistics, six strategies have been investigated, in which supervised learning accounts for the largest majority (more than 90%). Considering the expensive labeling works, some recent research explores liberating the training demand for camera parameters using semi-supervised learning, weakly-supervised learning, unsupervised learning, and self-supervised learning. Reinforcement learning also has been exploited to dynamically address the camera calibration problem.

## II. CAMERA MODEL

Researchers utilize mathematical formulas to establish camera models that describe the imaging process from a point in 3D world coordinates to its projection on a 2D image plane. Different cameras and systems correspond to different types of parametric models. In this section, we first provide a detailed formulation of the basic pinhole camera model. Then, we review more complex and useful camera models, as well as extended models studied in recent literature, to meet the advanced development of cameras and academic/industrial demands.

### A. Pinhole Camera Model

The most popular and commonly applied camera model in computer vision is the pinhole camera model. It can be regarded as a geometrically accurate first-order approximation of the traditional camera. A pinhole camera has one single effective perspective because the pinhole aperture is thought to be an infinitesimal point through which all projection lines must pass.

Using a mathematical formulation, the camera model depicts the imaging process from a point in the 3D world coordinate to its projection on the 2D image plane. Assuming the homogeneous coordinates $\mathbf{P}_w = [X, Y, Z, 1]^\mathsf{T} \in \mathbb{R}^{4 \times 1}$ and $\mathbf{P}_i = [u, v, 1]^\mathsf{T} \in \mathbb{R}^{3 \times 1}$ denote a point in the 3D world coordinate and its corresponding point on a 2D image plane, respectively. Then, a camera model can be described by a projection mapping $M \in \mathbb{R}^{3 \times 4}$ between $\mathbf{P}_w$ and $\mathbf{P}_i$:

$$\mathbf{P}_i = M\mathbf{P}_w, \tag{1}$$

where the projection can be further formed by:

$$\mathbf{P}_c = [\mathbf{R} \,|\, \mathbf{t}]\mathbf{P}_w, \tag{2}$$

where $\mathbf{P}_c = [x_c, y_c, z_c]^\mathsf{T} \in \mathbb{R}^{3 \times 1}$ denotes a transformed point in the camera coordinate using a $3 \times 3$ rotation $\mathbf{R}$ and a 3-dimension translation $\mathbf{t}$. The $3 \times 4$ matrix $[\mathbf{R} \,|\, \mathbf{t}]$ is generally named as the extrinsic camera matrix, in which the camera rotation can be further parameterized by three angles: yaw $\varphi$, pitch $\theta$, and roll $\psi$. Subsequently, the point $\mathbf{P}_c$ is projected onto a surface. This surface is represented by the pinhole camera model as a plane $z = 1$, and the normalized coordinate of the point in camera coordinate is expressed by $[x_n, y_n]^\mathsf{T} = [\frac{x_c}{z_c}, \frac{y_c}{z_c}]^\mathsf{T}$.

Finally, the point on the normalized plane is projected onto the image plane, obtaining a pixel $\mathbf{P}_i$ by:

$$\mathbf{P}_i = K[x_n, y_n, 1]^\mathsf{T}, \tag{3}$$

where $K \in \mathbb{R}^{3 \times 3}$ is an intrinsic camera matrix, which consists of various camera intrinsic parameters such as the focal length, skew coefficient, and image center:

$$K = \begin{bmatrix} f_x m_u & s & c_u \\ 0 & f_y m_v & c_v \\ 0 & 0 & 1 \end{bmatrix}, \tag{4}$$

where $f_x$ and $f_y$ are the focal lengths at X-axis and Y-axis of the camera, respectively. Generally, for most cameras, $f_x = f_y$, and they are unified to $f$. $m_u$ and $m_v$ are the number of pixels per unit distance, in which $m_u = m_v$, if the image has square pixels. $s$ is the skew coefficient. A CCD sensor's pixels might not be precisely square, which would cause a

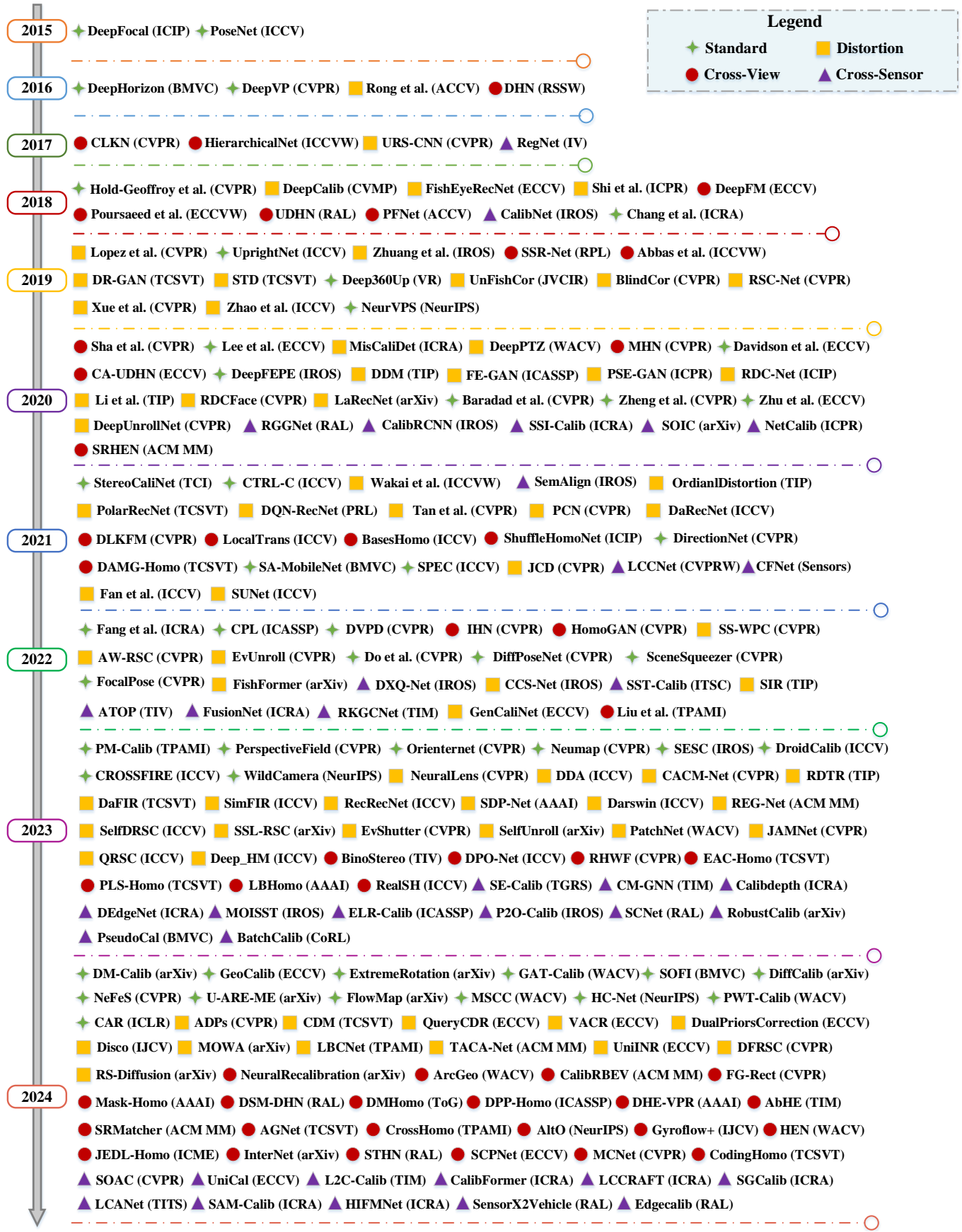

Fig. 1. A concise milestone of deep learning-based camera calibration methods. We classify all methods based on the uncalibrated camera model and its extended applications: standard model, distortion model, cross-view model, and cross-sensor model.

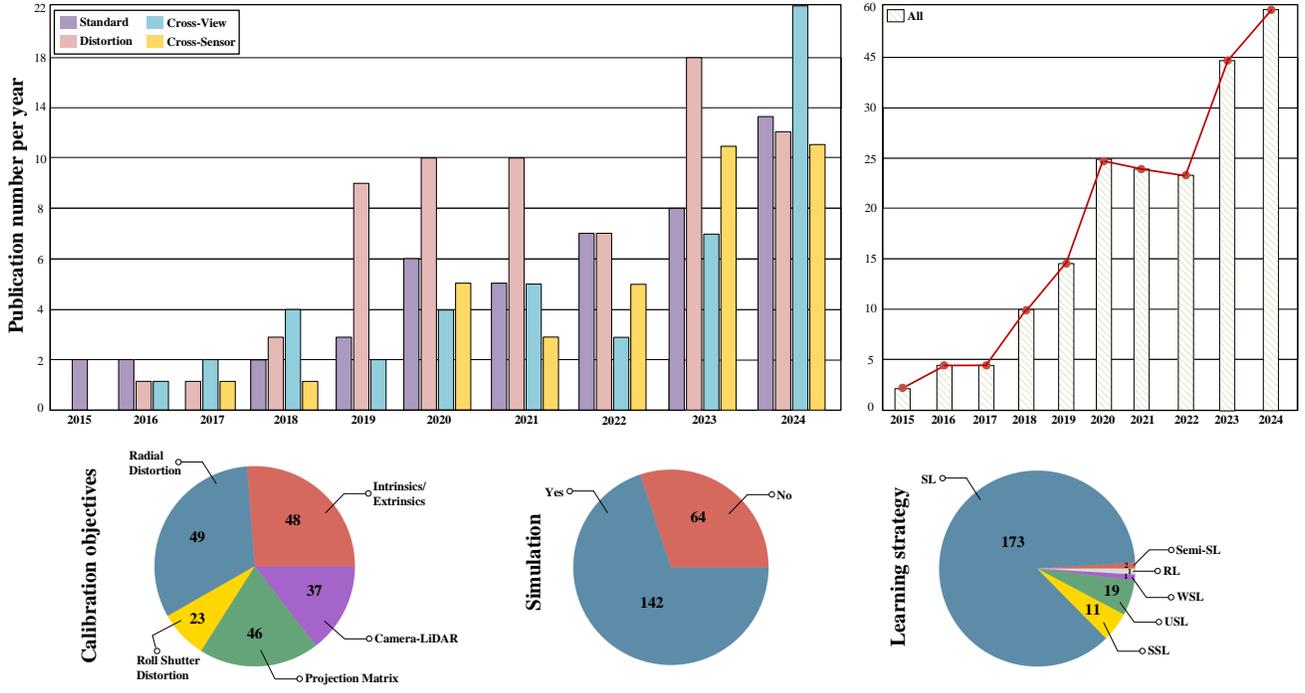

Fig. 2. A statistic analysis of deep learning-based camera calibration methods. Specifically, we summarize all literature based on the number of publications per year, calibration objectives, simulation of the dataset, and learning strategy.

slight distortion in the X or Y axes. The number of pixels on the CCD sensor per unit length in each direction is known as the skew coefficient. It would become 0 when X-axis and Y-axis are perpendicular to each other. $[c_u, c_v]^\mathsf{T}$ is the coordinate of the image center. According to previous works and factory design, the intrinsic parameters can be refined by $s = 0, m_u = m_v$ and focal length in the pixel unit, then Eq. (3) can be reformulated as:

$$\mathbf{P}_i = \begin{bmatrix} f_x & 0 & c_u \\ 0 & f_y & c_v \\ 0 & 0 & 1 \end{bmatrix} [x_n, y_n, 1]^\mathsf{T}. \tag{5}$$

In addition to numerical camera parameters, some geometric representations can provide useful clues for camera calibration, such as vanishing points and horizon lines. These representations establish clear relationships between image features and calibration objectives, which can alleviate the difficulty of learning conventional and implicit camera parameters.

Lines and points are both represented as three-dimensional vectors in homogeneous coordinates. The definitions for computing the line $\mathbf{l}$ that connects two points and the point $\mathbf{p}$ at the intersection of two lines can be given by:

$$\mathbf{l} = \frac{\mathbf{p}_1 \times \mathbf{p}_2}{||\mathbf{p}_1 \times \mathbf{p}_2||} \qquad \mathbf{p} = \frac{\mathbf{l}_1 \times \mathbf{l}_2}{||\mathbf{l}_1 \times \mathbf{l}_2||} \tag{6}$$

There are two parameterizations of the horizon line: slope-offset $(\theta, \rho)$ and left-right $(l, r)$. Assuming that the viewing orientation is down the negative $z$-axis, with the positive $x$-direction to the right, and the positive $y$-direction to the up. As a result, the world viewing direction of the camera can be described by $R_c^\mathsf{T}[0, 0, -1]^\mathsf{T}$. For the world vector $[0, 1, 0]^\mathsf{T}$

points in the zenith direction, a set of points $p$ can represent the horizon line:

$$p^\mathsf{T} K^{-T} R[0, 1, 0]^\mathsf{T} = 0. \tag{7}$$

As mentioned in Barnard [1], the normalized line direction vector $\mathbf{d}$ can be formulated for the Gaussian sphere representation of a vanishing point $\mathbf{v}$. In particular, supposed a 3D ray is described by $\mathbf{o} + \lambda\mathbf{d}$, where $\mathbf{o}$ and $\mathbf{d}$ are its origin and unit direction vector, respectively. Then, the vanishing point can be represented by $\lambda \to \infty$, in which the image coordinate is formed by $\mathbf{v} = [v_x, v_y]^\mathsf{T} := \lim_{\lambda \to \infty}[p_x, p_y]^\mathsf{T} \in \mathbb{R}^2$. Thus, the 3D direction of a line based on its vanishing point can be calculated by:

$$\mathbf{d} = \begin{bmatrix} v_x - c_x & v_y - c_y & f \end{bmatrix}^\mathsf{T} \in \mathbb{R}^3. \tag{8}$$

By using $\mathbf{d}$ rather than $\mathbf{v}$, the degraded situations where $\mathbf{d}$ is parallel to the image plane are eliminated. Additionally, it provides a natural measurement for determining the separation between two vanishing points.

### B. Wide-angle Camera Model

The perspective projection model, given a typical pinhole camera with focal length $f$, can be expressed as:

$$r = f \tan \theta, \tag{9}$$

where $r$ denotes the projection distance between the principal point and the points in the image. $\theta$ denotes the angle between the incident ray and the optical axis of the camera. It is straightforward to determine that $\theta$ should be less than $90°$. Without a projection point on the image plane, the incoming ray will not cross with the image plane and the pinhole camera

will not be able to view anything behind. Because of their restricted field of view (FoV), most cameras cannot see all of the points in the 3D environment at the same time.

Due to the wide FoV, wide-angle cameras are increasingly widely used in computer vision and robotics tasks such as navigation, localization, and tracking. Specifically, an extra wide-angle lens called a fisheye camera is used to create a broad, hemispherical, or panoramic image. Fisheye lenses employ a specific mapping to produce convex and non-rectilinear images as opposed to images with straight lines of perspective (rectilinear images). However, the wide-angle camera violates the pinhole camera assumption and the captured image suffers from geometric distortions.

Geometric distortion induced by wide-angle cameras can generally be classified into radial distortion and tangential distortion (de-centering distortion). Radial distortion is the primary distortion in central single-view camera systems, exhibiting circular symmetry with respect to the distortion center. This distortion results in points on the image plane being moved away from their ideal location under the perspective camera model along the radial axis from the distortion center. Radial distortion models can be formulated as nonlinear functions of the radial distance [2]. On the other hand, tangential distortion occurs when the lens and image plane are not parallel. Tangential distortion, also known as de-centering distortion, is primarily caused by the lens assembly not being centered over and parallel to the image plane. Unlike radial distortion, tangential distortion has a geometric impact that is not solely along the radial axis, and can also cause rotation and skewing of the image plane with respect to the distance from the image center. The camera model with radial distortion and tangential distortion can be parameterized by:

$$
\begin{cases}
x_r &= x_d + \bar{x}(k_1 r_d^2 + k_2 r_d^4 + k_3 r_d^6 + \cdots) \\
&\quad + (p_1(r_d^2 + 2\bar{x}^2) + 2p_2\bar{x}\bar{y})(1 + p_3 r_d^2 + \cdots) \\
y_r &= y_d + \bar{y}(k_1 r_d^2 + k_2 r_d^4 + k_3 r_d^6 + \cdots) \\
&\quad + (p_2(r_d^2 + 2\bar{y}^2) + 2p_1\bar{x}\bar{y})(1 + p_3 r_d^2 + \cdots)
\end{cases}, \quad (10)
$$

where $\bar{x} = x_d - c_x$ and $\bar{y} = y_d - c_y$. $K = (k_1, k_2, k_3, \dots)$ and $P = (p_1, p_2, p_3, \dots)$ are the radial distortion parameters and decentering distortion parameters, respectively. $r_d$ describes the radial distance from an image point to the distortion center $(c_x, c_y)$. Such an equation represents the mapping from a point $[x_d, y_d]^\mathsf{T}$ in the image captured by the wide-angle camera to that in the rectified image without the geometric distortion $[x_r, y_r]^\mathsf{T}$.

Previous works demonstrate that tangential distortion is basically insignificant and can be neglected. Moreover, as we surveyed, all learning-based camera calibration methods only consider the radial distortion for calibrating the wide-angle camera. To this end, Eq.10 can be simplified by a Taylor expansion:

$$
\begin{cases}
x_r = x_d(k_1 r_d^2 + k_2 r_d^4 + k_3 r_d^6 + \cdots) \\
y_r = y_d(k_1 r_d^2 + k_2 r_d^4 + k_3 r_d^6 + \cdots)
\end{cases}. \quad (11)
$$

This equation is known as the even-order polynomial model, which can also be expressed as an odd-order polynomial model by shifting the power. However, according to Wang et al. [3], while the polynomial model is suitable for small distortions, it

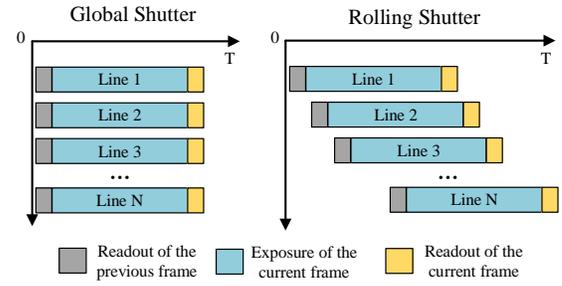

Fig. 3. Comparison of the mechanism of global shutter camera and rolling shutter camera.

requires an unreasonably high number of non-zero distortion parameters for severe distortions. As an alternative, Fitzgibbon et al. [4] proposed a division model that more accurately approximates the genuine undistortion function of a common camera. For significant distortion, the division model is preferred over the polynomial model because it requires fewer terms:

$$
\begin{cases}
x_r = \frac{x_d}{k_1 r_d^2 + k_2 r_d^4 + k_3 r_d^6 + \cdots} \\
y_r = \frac{y_d}{k_1 r_d^2 + k_2 r_d^4 + k_3 r_d^6 + \cdots}
\end{cases}. \quad (12)
$$

Some classical works demonstrate the single-parameter division model (only with distortion parameter $k_1$ in Eq.12) seems to be sufficient for most wide-angle cameras, which has been widely applied in learning-based wide-angle camera calibration [5]–[8].

### C. Rolling Shutter Camera Model

Due to the compact design, low price, and high frame rate, numerous consumer cameras, including webcams and mobile phones, employ CMOS (complementary metal–oxide–semiconductor) sensors. However, they are restricted to using rolling shutter (RS) devices. With a consistent time delay between each row, RS exposes the sensor array row by row from top to bottom, as opposed to global shutter (GS) based on CCD sensors, which simultaneously read out all rows of the sensor array. If the RS camera is moving while capturing the image, various distortions, such as skew, smear, or wobble, will break the reality of the original scene, which deviates from the pinhole camera paradigm. The unknown camera movements during the capturing process induce the so-called RS effects (also known as the jelly effect). In other words, an RS image is a row-by-row combination of GS images taken by a virtual moving GS camera throughout the camera readout time. The comparison of the RS camera and GS camera is shown in Figure 3.

The RS camera can be regarded as a high-frequency sensor that produces sparse spatial information with rich temporal coverage conveyed by distortions [9]. Modeling the RS camera faces a common challenge of estimating the transformation between RS and GS images. Assume a 3D latent space-time volume captures the desired scene across the desired time period $[0, t_0]$ and creates a virtual GS image $\mathbf{I}^{\text{GS}}$. We suppose

the readout direction is from top to bottom, and then the row-by-row readout RS imaging $\mathbf{I}^{\text{RS}}$ can be expressed by:

$$\mathbf{I}^{\text{RS}} = \sum_{y=1}^{H} M(\mathbf{I}^{\text{GS}}_{t_r}, y), \tag{13}$$

where $H$ is the height of the RS image (total number of rows) and $y$ indicates the vertical coordinate. $M(\cdot, \cdot)$ masks a specific row in the GS image, in which $t_r$ represents the readout (offset) time for each row of RS.

On the other hand, by warping the RS features backward with an estimated displacement field, the GS image can be formulated by:

$$\mathbf{I}^{\text{GS}}(\mathbf{x}) = \mathbf{I}^{\text{RS}}(\mathbf{x} + \mathbf{F}_{GS \to RS}(\mathbf{x})), \tag{14}$$

where $\mathbf{F}_{GS \to RS} \in \mathbb{R}^2$ denotes the displacement field of the pixel $\mathbf{x}$ from the GS image to the RS image.

The above formulations describe the rolling shutter camera model under a short exposure scenario. When the exposure time of the camera increases, the motion blur effects occur in the captured image, jointly with the RS distortion:

$$\mathbf{I}^{\text{RS}'}_{(t)}[i] = \frac{1}{T} \int_{t-t_h+it_r-T/2}^{t-t_h+it_r+T/2} \mathbf{I}^{\text{GS}}_{(t-t_h+it_r)}[i] dt, \tag{15}$$

where $\mathbf{I}^{\text{RS}'}_{(t)}[i]$ denotes the $i^{th}$ row of the RS distortion image $\mathbf{I}^{\text{RS}}$ with the middle moment of exposure at time $t$. $T$ indicates the exposure time of camera and $t_h = (H/2)t_r$.

### D. Cross-View Camera Model

The cross-view camera model is a type of multi-view camera system used in computer vision. It involves placing two or more cameras at opposite sides of a scene to capture multiple views of the same scene. This setup enables the creation of 3D reconstructions of the scene by triangulating corresponding points from multiple camera views. The cross-view camera model is commonly used in surveillance, robotics, and augmented reality applications, and provides a more accurate and complete representation of the scene than what can be achieved with a single camera. Alternatively, a camera with stable movement can also be regarded as a cross-view camera model.

In a cross-view camera model, the captured images can be used to calculate the fundamental matrix and homography matrix, which are essential tools for 3D reconstruction, image rectification, and camera calibration.

**Fundamental Matrix** Geometric relationships between the 3D points and their projections onto the 2D plane impose constraints on the image points when two cameras capture the same 3D scene from different perspectives. This intrinsic projective geometry can be embodied by a fundamental matrix $\mathbf{F}$.

$$\mathbf{F} = \mathbf{K_2}^{-T}[\mathbf{t}]_{\times} \mathbf{R} \mathbf{K_1}^{-1}. \tag{16}$$

Such an equation describes the epipolar geometry, where $\mathbf{K_1}$ and $\mathbf{K_2}$ indicate the intrinsic parameters of two cameras, and $\mathbf{R}$ and $[\mathbf{t}]_{\times}$ are the relative camera rotation and translation, respectively.

The fundamental matrix can be calculated from the correspondences of projected scene points by $q^T \mathbf{F} p = 0$, in which $q$ and $p$ are the matching points derived from two views. Specifically, the eight-point algorithm [10] uses 8 point correspondences and enforces the rank-2 constraint using Singular Value Decomposition (SVD), computing a matrix with the minimum Frobenius distance.

**Homography Matrix** Estimating a 2D homography matrix (or projection transformation) is an elemental geometric task for a pair of images that are captured from the same planar surface in a 3D scene with different perspectives. An invertible mapping from one image plane to another with eight degrees of freedom: two each for translation, rotation, scale, and lines at infinity, is known as a homography. Supposed that the homogeneous coordinates $\mathbf{x} = [u, v, 1]^T \in \mathbb{R}^{3 \times 1}$ and $\mathbf{x}' = [u', v', 1]^T \in \mathbb{R}^{3 \times 1}$ are points from two images but indicating the same point in the 3D scene, a non-singular $3 \times 3$ matrix can represent a linear transformation that maps $\mathbf{x} \Leftrightarrow \mathbf{x}'$ as a planar projective transformation or homography $\mathbf{H}$:

$$\begin{bmatrix} u' \\ v' \\ 1 \end{bmatrix} \sim \begin{bmatrix} h_{11} & h_{12} & h_{13} \\ h_{21} & h_{22} & h_{23} \\ h_{31} & h_{32} & h_{33} \end{bmatrix} \begin{bmatrix} u \\ v \\ 1 \end{bmatrix}, \tag{17}$$

where the transformation can be simplified as $\mathbf{x}' \sim \mathbf{H}\mathbf{x}$. This transformation can be rewritten by two following equations:

$$u' = \frac{h_{11}u + h_{12}v + h_{13}}{h_{31}u + h_{32}v + h_{33}}; v' = \frac{h_{21}u + h_{22}v + h_{23}}{h_{31}u + h_{32}v + h_{33}}. \tag{18}$$

Previous methods [11], [12] point out that the above conventional $3 \times 3$ parameterization $\mathbf{H}$ is not desirable for training neural networks. Concretely, it is challenging to guarantee the non-singularity of $\mathbf{H}$ due to the significant variance in the size of the members of the $3 \times 3$ homography matrix. Moreover, the rotation, translation, scale, and shear components of the homography transformation are mixed in $\mathbf{H}$. For instance, the submatrix $[h_{11} \quad h_{12}; h_{21} \quad h_{22}]$ describes the homography's rotational term and the vector $[h_{13}, h_{23}]^T$ denotes the translation transformation. Considering the rotation and shear components typically have smaller magnitudes than the translation component, it will have a negligible impact on the loss function of the component elements, leading to an imbalance training problem with a neural network. Instead, a 4-point parameterization [13] has been demonstrated to be more learning-friendly for learning-based homography estimation than the $3 \times 3$ parameterization. Supposed that the offsets of the image's vertex are $\Delta u_i = u'_i - u_i$ and $\Delta v_i = v'_i - v_i$, then the 4-point parameterization $\widetilde{\mathbf{H}}$ can describe a homography by:

$$\widetilde{\mathbf{H}} = \begin{pmatrix} \Delta u_1 & \Delta v_1 \\ \Delta u_2 & \Delta v_2 \\ \Delta u_3 & \Delta v_3 \\ \Delta u_4 & \Delta v_4 \end{pmatrix}. \tag{19}$$

The 4-point parameterization owns eight variables, which are equivalent to the matrix formulation of the homography. It is straightforward to transform from $\widetilde{\mathbf{H}}$ to $\mathbf{H}$ using the normalized Direct Linear Transform (DLT) [14] if the four corners' displacement is known.

### E. Cross-Sensor Model

Modern robots are often equipped with various sensors to provide a comprehensive understanding of the environment. These sensors capture scenes using different types of representations. For autonomous cars and robotics, cameras and Light Detection and Ranging sensors (LiDAR) are commonly used for vision tasks. The 3D LiDAR records long-range spatial data as sparse point clouds, while the camera captures texturally dense 2D color RGB images. Combining these sensors can facilitate 3D reconstruction and provide precise and robust perception for the robots, overcoming the limitations of individual sensors.

However, collision and vibration problems can occur when using different sensors in a robot or system. Additionally, the 3D point clouds cannot be effectively projected onto a 2D image without accurate extrinsic parameters, making it difficult to reliably correlate pixels in an image with depth information. Therefore, it is crucial to precisely calibrate the 2D-3D matching correspondences between pairs of temporally synchronized camera and LiDAR data.

The appropriate extrinsic calibration of the transformation (*i.e.*, rotation and translation) between the camera and LiDAR in 6-DoF is a key condition for data fusion. To be more specific, 3D LiDAR point cloud $PC = [X, Y, Z] \in \mathbb{R}^3$ can be projected onto the image plane by transforming it into the camera coordinate using the extrinsic matrix $T$ between the camera and LiDAR as well as camera intrinsic $K$. The inverse depth and the projected 2D coordinates can be represented as $d = 1/Z$ and $p = [u, v] \in \mathbb{R}^2$, respectively. Then, the camera-LiDAR model can be described by:

$$\begin{bmatrix} u \\ v \\ d \end{bmatrix} = \begin{bmatrix} f_x(\hat{X}/\hat{Z}) + c_x \\ f_y(\hat{Y}/\hat{Z}) + c_y \\ 1/\hat{Z} \end{bmatrix}, \tag{20}$$

where $(f_x, f_y)$ and $(c_x, c_y)$ indicate the focal lengths and the image center as listed in Eq. 4. $[\hat{X}, \hat{Y}, \hat{Z}]$ is the transformed point cloud $\hat{PC}$ using the estimated extrinsic matrix:

$$[\hat{X}, \hat{Y}, \hat{Z}, 1]^\mathsf{T} = T[X, Y, Z, 1]^\mathsf{T}. \tag{21}$$

Most deep learning works exploit the Lie algebra to parameterize the calibration camera-LiDAR extrinsic parameters. In particular, the output of the calibration network is a $1 \times 6$ vector $\xi = (v, \omega) \in se(3)$ in which $v$ is the translation vector, and $\omega$ is the rotation vector. To recover the original objectives, the rotation vector in $so(3)$ should be transformed to its corresponding rotation matrix. Supposed that $\omega = (\omega_1, \omega_2, \omega_3)^T$, an element $\omega \in so(3)$ can be transformed to $SO(3)$ using the exponential map by:

$$exp : so(3) \to SO(3); \; \hat{\omega} \mapsto e^{\hat{\omega}}, \tag{22}$$

where $\hat{\omega}$ and $e^{\hat{\omega}}$ denote the skew-symmetric matrix from $\omega$ and Taylor series expansion for the matrix exponential function, respectively. Then, the rotation matrix can be formed in $SO(3)$, and its Rodrigues formula is derived from the above equation by:

$$R = e^{\hat{\omega}} = I + \frac{\hat{\omega}}{\|\omega\|} \sin \|\omega\| + \frac{\hat{\omega}^2}{\|\omega\|^2}(1 - \cos(\|\omega\|)). \tag{23}$$

Thus, the 3D rigid body transformation $T \in SE(3)$ between camera and LiDAR can be represented by:

$$T = \begin{pmatrix} R & t \\ 0 & 1 \end{pmatrix} \text{ where } R \in SO(3), t \triangleq v \in \mathbb{R}^3. \tag{24}$$

## III. Benchmark Evaluation

To intuitively demonstrate the effectiveness of the existing learning-based camera calibration methods, we provide quantitative and qualitative evaluations of representative methods using our benchmark. Considering the application gap and comparison integrity, we also collect and summarize the reported metrics from previous works of certain camera models.

As listed in Tab. I, the quantitative evaluations on standard model calibration methods are provided. They are summarized from previous works [19] [20] [18] and evaluated with the Google Street View dataset. The angular errors of predicted camera Up direction, pitch, roll, and FoV are calculated in terms of the mean error (*Mean*) and median error (*Med*). AUC denotes the area under the curve of the cumulative distribution of the horizon line error.

Tab. II and Fig. 4 show the quantitative evaluation and qualitative evaluation of representative works for calibrating the distortion camera model, especially in radial distortion. In particular, we split the benchmark into two test sets (Test1 and Test2) based on the square and circle of the wide-angle image. Two types of metrics, namely, image-level metrics (PSNR, SSIM, MS-SSIM) and perception-level metrics (FID, LPIPS-Alex, LPIPS-Vgg) are used to calculate the errors between the corrected wide-angle image and the ground truth. Since FishFormer [21] is customized for the fisheye image with a circular shape, we omit its quantitative metric on Test1 and use the original input as its placeholder in Fig. 4.

We evaluate the performance of representative methods for the cross-view camera model in Tab. III. The mean average corner error (MACE) is provided for each method. For IHN [29], we test it from two scales following the same setting reported in the paper. Moreover, the point matching errors (PME) of representative methods are listed in Tab. IV, which are collected and summarized from previous works [33], [44] and evaluated on the GHOF [44] test set. Five categories are split to exhibit the performance of different scenes, including regular (RE), foggy (FOG), low light (LL), rainy (RAIN), and snowy (SNOW) scenes.

We also evaluate the performance of several open source methods on cross-sensor camera model in Tab. V. Experimental results are reported according to the metrics of rotation and translation. For the rotation part, we calculate quaternion angle distance and absolute angle error of Euler angles including Roll, Pitch, and Yaw. For the translation part, we compute the Euclidean distance between the predicted translation vector and ground truth with absolute error in $X$, $Y$, $Z$ directions. Since there has not been a common benchmark for camera-LiDAR calibration yet, all selected methods were trained on the KITTI [35] dataset with different specific settings. In the evaluation, we devote ourselves to aligning the proposed benchmark to each method's original setting. Due to the difference in image size, image content, LiDAR beam number, and

TABLE I
QUANTITATIVE EVALUATION ON THE REPRESENTATIVE CAMERA CALIBRATION METHODS FOR THE STANDARD MODEL.

| Method | Up Direction (°) ↓ | | Pitch (°) ↓ | | Roll (°) ↓ | | FoV (°) ↓ | | AUC (%) ↑ |
|---|---|---|---|---|---|---|---|---|---|
| | Mean | Med. | Mean | Med. | Mean | Med. | Mean | Med. | |
| DeepHorizon [15] | 3.58 | 3.01 | 2.76 | 2.12 | 1.78 | 1.67 | - | - | 80.29 |
| Hold-Geoffroy et al. [16] | 2.73 | 2.13 | 2.39 | 1.78 | 0.96 | 0.66 | 4.61 | 3.89 | 80.40 |
| UprightNet [17] | 28.20 | 26.10 | 26.56 | 24.56 | 6.22 | 4.33 | - | - | - |
| Lee et al. [18] | 2.12 | 1.61 | 1.92 | 1.38 | 0.75 | 0.47 | 6.01 | 3.72 | 83.12 |
| CTRL-C [19] | 1.80 | 1.52 | 1.58 | 1.31 | 0.66 | 0.53 | 3.59 | 2.72 | 87.29 |
| PerspectiveField [20] | - | - | 1.36 | 1.18 | 0.66 | 0.52 | 3.07 | 2.33 | - |

TABLE II
QUANTITATIVE EVALUATION ON THE REPRESENTATIVE CAMERA CALIBRATION METHODS FOR DISTORTION MODEL.

| Comparison on Test1 | | Metrics | | | | | |
|---|---|---|---|---|---|---|---|
| Methods | Type | PSNR ↑ | SSIM ↑ | MS-SSIM ↑ | FID ↓ | LPIPS-Alex ↓ | LPIPS-Vgg ↓ |
| DCCNN [5] | Regression | 9.18 | 0.1583 | 0.1617 | 360.90 | 0.4081 | 0.5258 |
| DeepCalib [7] | Regression | 10.71 | 0.2289 | 0.2534 | 161.31 | 0.3330 | 0.4482 |
| FishFormer [21] | Regression | - | - | - | - | - | - |
| BlindCor [6] | Reconstruction | 8.02 | 0.1356 | 0.1392 | 400.77 | 0.4956 | 0.5536 |
| DR-GAN [22] | Reconstruction | 16.81 | 0.5058 | 0.6794 | 148.50 | 0.1782 | 0.3268 |
| DDM [23] | Reconstruction | 17.43 | 0.5659 | 0.7191 | 125.34 | 0.1455 | 0.2333 |
| PCN [24] | Reconstruction | 24.59 | 0.8726 | 0.9594 | 42.78 | 0.0388 | 0.0671 |

| Comparison on Test2 | | Metrics | | | | | |
|---|---|---|---|---|---|---|---|
| Methods | Type | PSNR ↑ | SSIM ↑ | MS-SSIM ↑ | FID ↓ | LPIPS-Alex ↓ | LPIPS-Vgg ↓ |
| DCCNN [5] | Regression | 15.23 | 0.4205 | 0.3609 | 125.99 | 0.1586 | 0.2110 |
| DeepCalib [7] | Regression | 11.47 | 0.2715 | 0.3655 | 139.26 | 0.2853 | 0.3886 |
| FishFormer [21] | Regression | 21.52 | 0.7540 | 0.8971 | 66.37 | 0.1133 | 0.1107 |
| BlindCor [6] | Reconstruction | 12.01 | 0.3075 | 0.2531 | 133.16 | 0.3595 | 0.3907 |
| DR-GAN [22] | Reconstruction | 17.31 | 0.5133 | 0.6901 | 116.78 | 0.1218 | 0.2056 |
| DDM [23] | Reconstruction | 18.84 | 0.6247 | 0.7614 | 78.38 | 0.1055 | 0.1894 |
| PCN [24] | Reconstruction | 21.28 | 0.7027 | 0.8595 | 54.64 | 0.0812 | 0.1088 |

TABLE III
QUANTITATIVE EVALUATION ON THE REPRESENTATIVE METHODS FOR CROSS-VIEW CAMERA MODEL.

| Method | MSCOCO [25] | GoogleEarth [26] | GoogleMap [26] | CA-Homo [27] |
|---|---|---|---|---|
| MHN [28] | 1.1512 | 10.3078 | 13.1610 | - |
| MHN+DLKFM [26] | 0.7687 | 3.9629 | 5.2664 | - |
| IHN-scale1 [29] | 0.2652 | 1.5135 | 0.9610 | - |
| IHN-scale2 [29] | 0.1234 | 1.2110 | 0.6751 | - |
| CA-UDHN [27] | - | - | - | 0.8605 |
| BasesHomo [30] | - | - | - | 0.6808 |
| HomoGAN [31] | - | - | - | 0.3651 |

procedure of generating input depth map, the input color image and depth map to different trained models are inconsistent. From Tab. V, Fig. 5, and Fig. 6, we can observe there are noticeable performance degradations of comparison methods under the cross-domain evaluations.

## IV. MORE FUTURE DIRECTIONS

### A. Dataset

One of the main challenges of learning-based camera calibrations is the difficulty in constructing datasets with high accuracy. This requires laborious manual intervention to obtain real-world data with labels. As we summarized, approximately 70% of the works rely on synthesized datasets. However, the significant differences between synthesized and real-world datasets cannot be ignored, leading to domain gaps in the

learned models. Therefore, the construction of a standardized, large-scale calibration dataset would significantly benefit this community. Recent works have demonstrated that well-designed learning strategies, such as semi-supervised learning [45], self-supervised learning [46], [47], and unsupervised learning [12], [27], can help address the demand for annotations in learning-based camera calibrations. These strategies also have the potential to discover additional calibration priors within the data itself.

### B. Transfer Learning

The advancements in deep learning have led to the development of transfer learning techniques, which could facilitate the transfer of knowledge learned from one camera to another. This approach can significantly speed up and streamline the

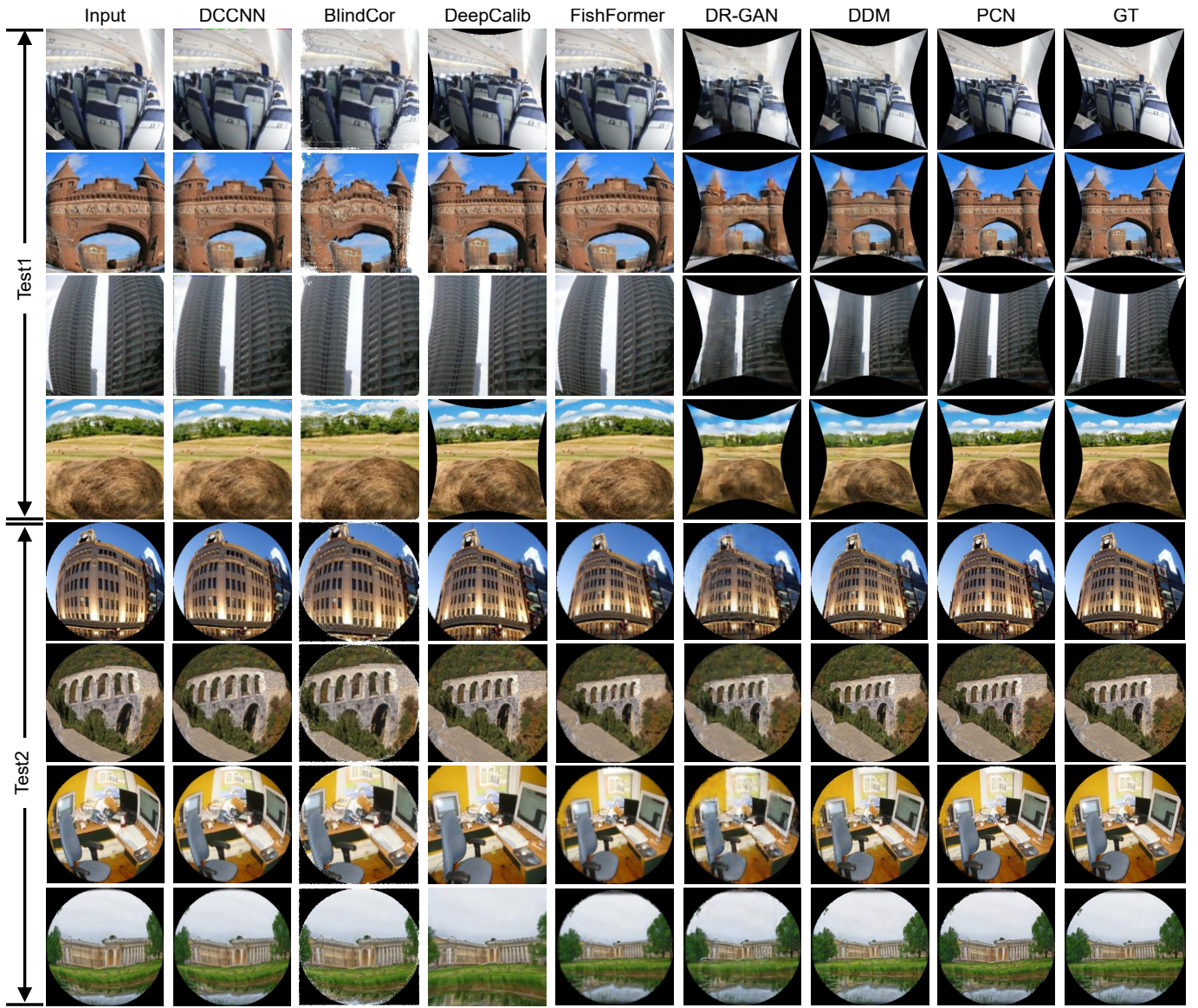

Fig. 4. Quantitative evaluation on the representative camera calibration methods for distortion model.

TABLE IV
THE POINT MATCHING ERRORS (PME) OF REPRESENTATIVE METHODS
FOR CROSS-VIEW CAMERA MODEL.

| Methods | AVG | RE | FOG | LL | RAIN | SNOW |
|---|---|---|---|---|---|---|
| $\mathcal{I}_{3\times3}$ | 6.33 | 4.94 | 7.24 | 8.09 | 5.48 | 5.89 |
| CA-UDHN [27] | 3.87 | 4.10 | 3.84 | 6.99 | 1.27 | 3.17 |
| BasesHomo [30] | 2.28 | 2.02 | 1.43 | 4.90 | 0.78 | 2.29 |
| HomoGAN [31] | 1.95 | 1.73 | 0.60 | 3.95 | 0.47 | 3.02 |
| DHN [11] | 6.61 | 6.04 | 6.02 | 7.68 | 6.99 | 6.32 |
| LocalTrans [32] | 5.72 | 4.06 | 6.49 | 5.95 | 5.78 | 6.34 |
| IHN [29] | 8.17 | 7.10 | 8.71 | 9.34 | 6.57 | 9.13 |
| DHN* [11] | 3.01 | 1.92 | 3.94 | 4.54 | 1.98 | 2.66 |
| LocalTrans* [32] | 2.89 | 1.78 | 4.27 | 4.59 | 1.37 | 2.43 |
| IHN* [29] | 2.59 | 2.21 | 3.05 | 4.70 | 0.98 | 2.03 |
| RealSH [33] | 1.72 | 1.60 | 0.88 | 4.42 | 0.43 | 1.28 |

calibration process, making it more efficient and cost-effective.

Transfer learning can be especially useful in applications that involve multiple cameras or mobile devices. For example, in a multi-camera system, transfer learning can be used to calibrate all the cameras using the data collected from a single camera, reducing the time and effort required for calibration. Similarly, in mobile devices, transfer learning can enable faster and more accurate calibration of the camera, resulting in improved image quality and performance.

### C. Robustness to Noise and Outliers

Another promising application of deep learning in camera calibration is improving the robustness of calibration to noise and outliers in the data. This approach can help ensure accurate calibration even in challenging environments, with low-quality data or noisy sensor readings. Conventionally, camera calibration algorithms are sensitive to noise and outliers in the data, which can lead to significant errors in the estimated camera

TABLE V
QUANTITATIVE EVALUATION ON THE REPRESENTATIVE CALIBRATION METHODS FOR CROSS-SENSOR MODEL.

| Methods | | Translation (cm) | | | | Rotation (°) | | | |
|---|---|---|---|---|---|---|---|---|---|
| | | $E_t$ | X | Y | Z | $E_R$ | Roll | Pitch | Yaw |
| CMRNet [34] 10°/2.0 m | KITTI [35] | 126.2820 | 72.8394 | 55.3118 | 77.1015 | 12.9863 | 5.7665 | 5.3867 | 3.9606 |
| | Apollo [36], [37] | 136.7067 | 44.6239 | 109.4236 | 82.5008 | 12.2843 | 8.2973 | 6.6979 | 5.7054 |
| | NuScenes [38] | 124.7204 | 43.6849 | 69.5890 | 64.1339 | 11.4439 | 9.9125 | 9.3100 | 5.3005 |
| | ONCE [39] | 145.6419 | 83.0297 | 26.1744 | 70.9436 | 11.3011 | 7.2812 | 7.4463 | 3.2014 |
| | DAIR-V2X [40] | 158.3873 | 51.8842 | 85.4628 | 79.1833 | 10.3119 | 3.1635 | 3.7430 | 5.5995 |
| CalibNet [41] 10°/0.2 m | KITTI [35] | 72.3976 | 37.4153 | 46.3160 | 30.4167 | 10.5577 | 4.8652 | 7.3470 | 5.4789 |
| | Apollo [36], [37] | 71.6006 | 18.7960 | 50.2996 | 16.5990 | 11.0372 | 6.0806 | 5.4223 | 9.4484 |
| | NuScenes [38] | 73.4821 | 52.6383 | 45.7675 | 27.4490 | 11.5573 | 8.0735 | 8.4622 | 5.6564 |
| | ONCE [39] | 74.0186 | 46.3395 | 49.3340 | 31.7829 | 12.4912 | 5.3240 | 5.3686 | 6.7609 |
| | DAIR-V2X [40] | 77.3034 | 41.9963 | 54.7920 | 51.8221 | 11.6145 | 7.4061 | 1.7497 | 6.4258 |
| LCCNet [42] 20°/1.5 m | KITTI [35] | 45.0945 | 9.2671 | 7.2877 | 11.6077 | 1.8611 | 0.6823 | 0.5427 | 0.2704 |
| | Apollo [36], [37] | 159.9080 | 45.5809 | 67.7147 | 106.1134 | 15.1098 | 2.5412 | 3.8089 | 2.8912 |
| | NuScenes [38] | 232.1684 | 73.4198 | 120.1943 | 98.3628 | 12.3146 | 2.5942 | 3.7928 | 9.0684 |
| | ONCE [39] | 185.0779 | 74.6479 | 72.9388 | 74.6713 | 11.5796 | 8.6251 | 6.2554 | 4.7137 |
| | DAIR-V2X [40] | 228.1306 | 113.5558 | 86.6683 | 98.3726 | 7.3732 | 2.0581 | 3.7936 | 4.7334 |
| RGGNet [43] 20°/0.3 m | KITTI [35] | 77.7839 | 12.5584 | 14.4128 | 68.0927 | 9.8177 | 2.1106 | 2.6808 | 6.7095 |
| | Apollo [36], [37] | 542.7074 | 215.6324 | 254.2824 | 218.8926 | 35.4403 | 17.8385 | 17.7508 | 18.2130 |
| | NuScenes [38] | 92.3324 | 44.9838 | 51.1874 | 51.5143 | 35.5249 | 22.6364 | 25.5031 | 18.2290 |
| | ONCE [39] | 73.8569 | 55.6384 | 49.2977 | 33.9010 | 26.0675 | 6.7498 | 6.7093 | 20.2362 |
| | DAIR-V2X [40] | 397.6241 | 260.4813 | 208.1069 | 187.8416 | 34.8505 | 22.0118 | 23.7622 | 18.6647 |

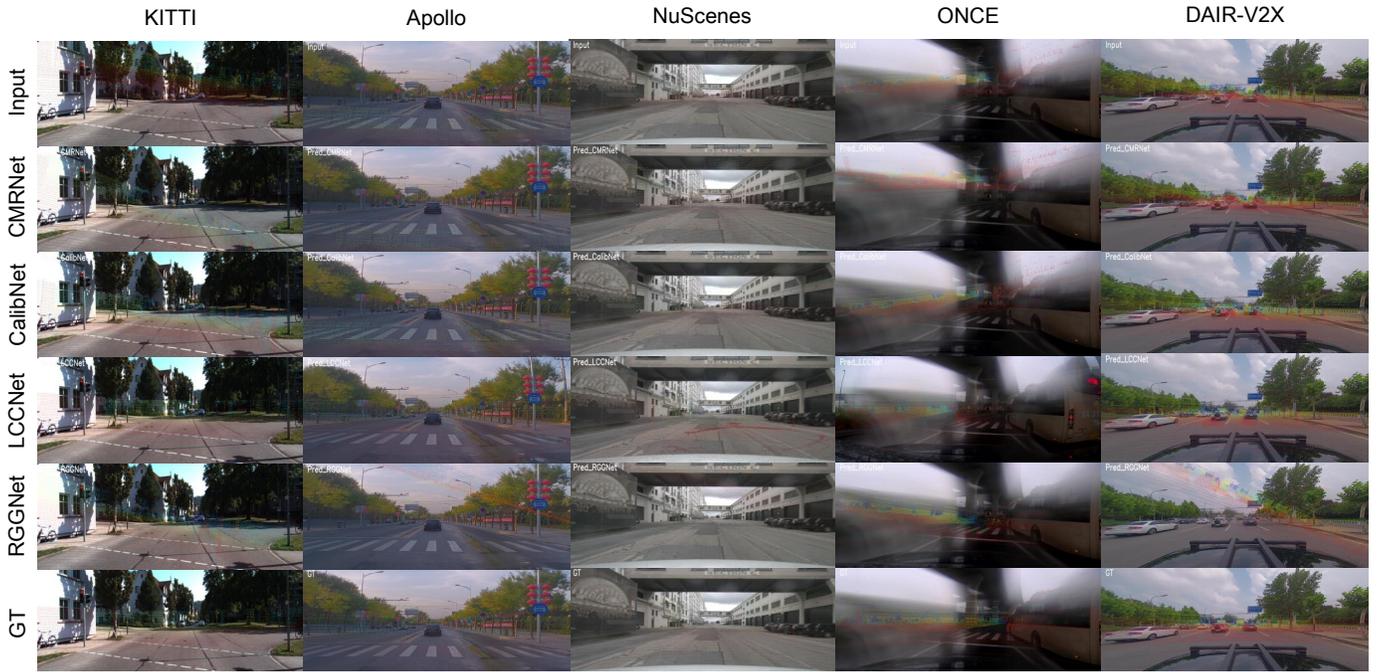

Fig. 5. Qualitative evaluation on the representative calibration methods for the cross-sensor model.

parameters. However, with the application of deep learning, it is possible to learn more robust and accurate models that can better handle noise and outliers in the data. For instance, regularization techniques can be used to impose constraints on the learned parameters, preventing overfitting and enhancing the generalization ability of the model. Moreover, outlier detection techniques can be used to identify and exclude data points that are likely to be outliers, reducing their impact on the calibration process. This can be achieved using various statistical and machine-learning methods, such as clustering, classification, and regression.

## D. Online Calibration

With the rapid development of deep learning, online camera calibration is becoming more efficient and practical. This technique involves updating the calibration parameters in real time, allowing for better performance as the camera moves or as the environment changes. This can be achieved using deep learning algorithms that can learn the complex relationships between the camera parameters and the image data. Learning-based camera calibration has the potential to revolutionize various industries, such as robotics and augmented reality. In robotics, online calibration can improve the accuracy of

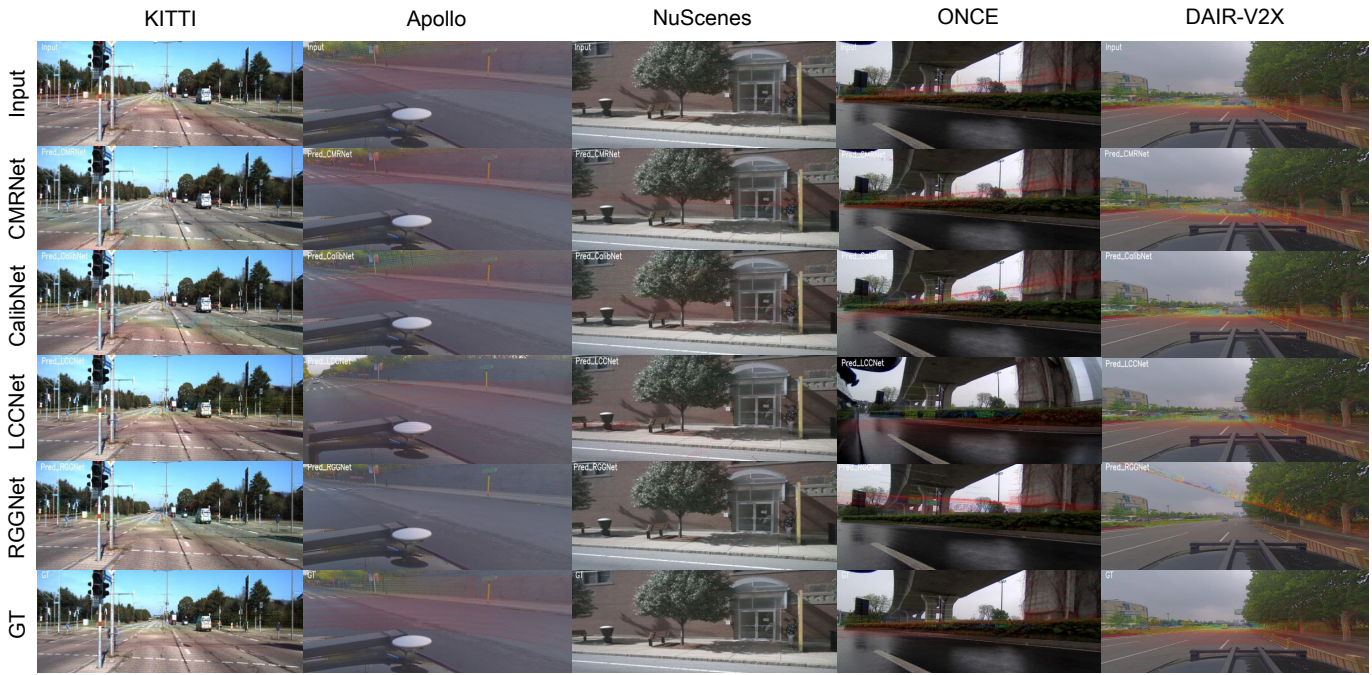

Fig. 6. Qualitative evaluation on the representative calibration methods for the cross-sensor model.

robot vision, which is crucial for tasks such as object detection and manipulation. Similarly, in augmented reality, online calibration can enhance the user experience by ensuring that virtual objects are correctly aligned with the real world. This can help create more realistic and immersive AR applications, which have numerous practical applications in fields such as entertainment, education, and training.

### E. Multimodal Calibration

The potential of deep learning techniques in camera calibration goes beyond traditional photography and computer vision applications. It could also be applied to calibrate cameras with other sensors, such as remote sensing, infrared sensors, or radar. This advancement could lead to more precise and robust perception in various applications, including but not limited to autonomous driving, where multiple sensors are used. Incorporating deep learning-based calibration methods with multiple sensors could enhance the accuracy of the fusion of data from different sources. It could facilitate more accurate perception in challenging environments such as low-light conditions, occlusions, and adverse weather conditions. Furthermore, the ability to calibrate multiple sensors with deep learning methods could provide more reliable and consistent results compared to traditional calibration techniques.

These are a few potential directions for future research in camera calibration with deep learning. As the field continues to evolve, there may be many other exciting avenues for exploration and innovation. In addition, it is also thrilling to see how this technology will continue to impact various industries in the future.

### REFERENCES

[1] S. T. Barnard, "Interpreting perspective images," *Artificial Intelligence*, vol. 21, no. 4, pp. 435–462, 1983.

[2] J. Fan, J. Zhang, S. J. Maybank, and D. Tao, "Wide-angle image rectification: a survey," *International Journal of Computer Vision*, vol. 130, no. 3, pp. 747–776, 2022.

[3] A. Wang, T. Qiu, and L. Shao, "A simple method of radial distortion correction with centre of distortion estimation," *Journal of Mathematical Imaging and Vision*, vol. 35, no. 3, pp. 165–172, 2009.

[4] A. W. Fitzgibbon, "Simultaneous linear estimation of multiple view geometry and lens distortion," in *Proceedings of IEEE Computer Society Conference on Computer Vision and Pattern Recognition*, 2001.

[5] J. Rong, S. Huang, Z. Shang, and X. Ying, "Radial lens distortion correction using convolutional neural networks trained with synthesized images," in *Asian Conference on Computer Vision*, 2016, pp. 35–49.

[6] X. Li, B. Zhang, P. V. Sander, and J. Liao, "Blind geometric distortion correction on images through deep learning," in *Proceedings of the IEEE/CVF Conference on Computer Vision and Pattern Recognition*, 2019.

[7] O. Bogdan, V. Eckstein, F. Rameau, and J.-C. Bazin, "Deepcalib: a deep learning approach for automatic intrinsic calibration of wide field-of-view cameras," in *Proceedings of the 15th ACM SIGGRAPH European Conference on Visual Media Production*, 2018.

[8] J. Zhao, S. Wei, L. Liao, and Y. Zhao, "Dqn-based gradual fisheye image rectification," *Pattern Recognition Letters*, vol. 152, pp. 129–134, 2021.

[9] O. Ait-Aider, N. Andreff, J. M. Lavest, and P. Martinet, "Simultaneous object pose and velocity computation using a single view from a rolling shutter camera," in *European Conference on Computer Vision*, 2006, pp. 56–68.

[10] H. C. Longuet-Higgins, "A computer algorithm for reconstructing a scene from two projections," *Nature*, vol. 293, no. 5828, pp. 133–135, 1981.

[11] D. DeTone, T. Malisiewicz, and A. Rabinovich, "Deep image homography estimation," *arXiv preprint arXiv:1606.03798*, 2016.

[12] T. Nguyen, S. W. Chen, S. S. Shivakumar, C. J. Taylor, and V. Kumar, "Unsupervised deep homography: A fast and robust homography estimation model," *IEEE Robotics and Automation Letters*, vol. 3, no. 3, pp. 2346–2353, 2018.

[13] S. Baker, A. Datta, and T. Kanade, "Parameterizing homographies," *Robotics Institute, Pittsburgh, PA, Tech. Rep. CMU-RI-TR-06-11*, 2006.

[14] B. K. Horn, "The direct linear transformation from comparator coordinates into object-space coordinates in close-range photogrammetry," *ISPRS Journal of Photogrammetry and Remote Sensing*, vol. 42, no. 3, pp. 125–133, 1987.

[15] S. Workman, M. Zhai, and N. Jacobs, "Horizon lines in the wild," *British Machine Vision Conference*, 2016.

[16] Y. Hold-Geoffroy, K. Sunkavalli, J. Eisenmann, M. Fisher, E. Gambaretto, S. Hadap, and J.-F. Lalonde, "A perceptual measure for deep single image camera calibration," in *Proceedings of the IEEE Conference on Computer Vision and Pattern Recognition*, 2018.

[17] W. Xian, Z. Li, M. Fisher, J. Eisenmann, E. Shechtman, and N. Snavely, "Uprightnet: Geometry-aware camera orientation estimation from single images," in *International Conference on Computer Vision*, 2019.

[18] J. Lee, M. Sung, H. Lee, and J. Kim, "Neural geometric parser for single image camera calibration," in *European Conference on Computer Vision*, 2020, pp. 541–557.

[19] J. Lee, H. Go, H. Lee, S. Cho, M. Sung, and J. Kim, "Ctrl-c: Camera calibration transformer with line-classification," in *International Conference on Computer Vision*, 2021, pp. 16 228–16 237.

[20] L. Jin, J. Zhang, Y. Hold-Geoffroy, O. Wang, K. Blackburn-Matzen, M. Sticha, and D. F. Fouhey, "Perspective fields for single image camera calibration," in *Proceedings of the IEEE/CVF Conference on Computer Vision and Pattern Recognition*, 2023, pp. 17 307–17 316.

[21] Y. Shangrong, L. Chunyu, L. Kang, and Z. Yao, "Fishformer: Annulus slicing-based transformer for fisheye rectification with efficacy domain exploration," *arXiv preprint arXiv:2207.01925*, 2022.

[22] K. Liao, C. Lin, Y. Zhao, and M. Gabbouj, "Dr-gan: Automatic radial distortion rectification using conditional gan in real-time," *IEEE Transactions on Circuits and Systems for Video Technology*, vol. 30, no. 3, pp. 725–733, 2020.

[23] K. Liao, C. Lin, Y. Zhao, and M. Xu, "Model-free distortion rectification framework bridged by distortion distribution map," *IEEE Transactions on Image Processing*, vol. 29, pp. 3707–3718, 2020.

[24] S. Yang, C. Lin, K. Liao, C. Zhang, and Y. Zhao, "Progressively complementary network for fisheye image rectification using appearance flow," in *Proceedings of the IEEE/CVF Conference on Computer Vision and Pattern Recognition*, 2021, pp. 6348–6357.

[25] T.-Y. Lin, M. Maire, S. Belongie, J. Hays, P. Perona, D. Ramanan, P. Dollár, and C. L. Zitnick, "Microsoft coco: Common objects in context," in *European Conference on Computer Vision*, 2014, pp. 740–755.

[26] Y. Zhao, X. Huang, and Z. Zhang, "Deep lucas-kanade homography for multimodal image alignment," in *Proceedings of the IEEE/CVF Conference on Computer Vision and Pattern Recognition*, 2021, pp. 15 950–15 959.

[27] J. Zhang, C. Wang, S. Liu, L. Jia, N. Ye, J. Wang, J. Zhou, and J. Sun, "Content-aware unsupervised deep homography estimation," in *European Conference on Computer Vision*, 2020, pp. 653–669.

[28] H. Le, F. Liu, S. Zhang, and A. Agarwala, "Deep homography estimation for dynamic scenes," in *Proceedings of the IEEE/CVF Conference on Computer Vision and Pattern Recognition*, 2020.

[29] S.-Y. Cao, J. Hu, Z. Sheng, and H.-L. Shen, "Iterative deep homography estimation," pp. 1879–1888, 2022.

[30] N. Ye, C. Wang, H. Fan, and S. Liu, "Motion basis learning for unsupervised deep homography estimation with subspace projection," in *International Conference on Computer Vision*, 2021, pp. 13 117–13 125.

[31] M. Hong, Y. Lu, N. Ye, C. Lin, Q. Zhao, and S. Liu, "Unsupervised homography estimation with coplanarity-aware gan," pp. 17 663–17 672, 2022.

[32] R. Shao, G. Wu, Y. Zhou, Y. Fu, L. Fang, and Y. Liu, "Localtrans: A multiscale local transformer network for cross-resolution homography estimation," in *International Conference on Computer Vision*, 2021, pp. 14 890–14 899.

[33] H. Jiang, H. Li, S. Han, H. Fan, B. Zeng, and S. Liu, "Supervised homography learning with realistic dataset generation," in *Proceedings of the IEEE/CVF International Conference on Computer Vision*, 2023, pp. 9806–9815.

[34] D. Cattaneo, M. Vaghi, A. L. Ballardini, S. Fontana, D. G. Sorrenti, and W. Burgard, "Cmrnet: Camera to lidar-map registration," in *IEEE Intelligent Transportation Systems Conference*, 2019, pp. 1283–1289.

[35] A. Geiger, P. Lenz, and R. Urtasun, "Are we ready for autonomous driving? the kitti vision benchmark suite," in *IEEE Conference on Computer Vision and Pattern Recognition*, 2012, pp. 3354–3361.

[36] X. Huang, P. Wang, X. Cheng, D. Zhou, Q. Geng, and R. Yang, "The apolloscape open dataset for autonomous driving and its application," *IEEE Transactions on Pattern Analysis and Machine Intelligence*, vol. 42, no. 10, pp. 2702–2719, 2019.

[37] W. Lu, Y. Zhou, G. Wan, S. Hou, and S. Song, "L3-net: Towards learning based lidar localization for autonomous driving," in *Proceedings of the IEEE Conference on Computer Vision and Pattern Recognition*, 2019, pp. 6389–6398.

[38] H. Caesar, V. Bankiti, A. H. Lang, S. Vora, V. E. Liong, Q. Xu, A. Krishnan, Y. Pan, G. Baldan, and O. Beijbom, "nuscenes: A multimodal dataset for autonomous driving," in *Proceedings of the IEEE/CVF conference on computer vision and pattern recognition*, 2020, pp. 11 621–11 631.

[39] J. Mao, M. Niu, C. Jiang, X. Liang, Y. Li, C. Ye, W. Zhang, Z. Li, J. Yu, C. Xu *et al.*, "One million scenes for autonomous driving: Once dataset," 2021.

[40] H. Yu, Y. Luo, M. Shu, Y. Huo, Z. Yang, Y. Shi, Z. Guo, H. Li, X. Hu, J. Yuan *et al.*, "Dair-v2x: A large-scale dataset for vehicle-infrastructure cooperative 3d object detection," in *Proceedings of the IEEE/CVF Conference on Computer Vision and Pattern Recognition*, 2022, pp. 21 361–21 370.

[41] G. Iyer, R. K. Ram, J. K. Murthy, and K. M. Krishna, "Calibnet: Geometrically supervised extrinsic calibration using 3d spatial transformer networks," in *IEEE/RSJ International Conference on Intelligent Robots and Systems*, 2018, pp. 1110–1117.

[42] X. Lv, B. Wang, Z. Dou, D. Ye, and S. Wang, "Lccnet: Lidar and camera self-calibration using cost volume network," in *Proceedings of the IEEE/CVF Conference on Computer Vision and Pattern Recognition*, 2021, pp. 2894–2901.

[43] K. Yuan, Z. Guo, and Z. J. Wang, "Rggnet: Tolerance aware lidar-camera online calibration with geometric deep learning and generative model," *IEEE Robotics and Automation Letters*, vol. 5, no. 4, pp. 6956–6963, 2020.

[44] H. Li, K. Luo, B. Zeng, and S. Liu, "Gyroflow+: Gyroscope-guided unsupervised deep homography and optical flow learning," *International Journal of Computer Vision*, vol. 132, no. 6, pp. 2331–2349, 2024.

[45] F. Zhu, S. Zhao, P. Wang, H. Wang, H. Yan, and S. Liu, "Semi-supervised wide-angle portraits correction by multi-scale transformer," in *Proceedings of the IEEE/CVF Conference on Computer Vision and Pattern Recognition*, 2022, pp. 19 689–19 698.

[46] J. Fang, I. Vasiljevic, V. Guizilini, R. Ambrus, G. Shakhnarovich, A. Gaidon, and M. R. Walter, "Self-supervised camera self-calibration from video," 2022.

[47] C. Wang, X. Wang, X. Bai, Y. Liu, and J. Zhou, "Self-supervised deep homography estimation with invertibility constraints," *Pattern Recognition Letters*, vol. 128, pp. 355–360, 2019.


\end{document}